\documentclass[10pt,journal]{IEEEtran}  
\usepackage[T1]{fontenc}
\ifCLASSOPTIONcompsoc
\usepackage[nocompress]{cite}
\else
\usepackage{cite}
\fi

\ifCLASSINFOpdf
\else
\fi

\usepackage{xcolor}
\usepackage{color}
\usepackage{booktabs}
\usepackage{multirow}
\usepackage{colortbl}
\usepackage{graphicx}
\usepackage{caption}
\usepackage{array}
\usepackage{dblfloatfix}
\usepackage{balance}
\usepackage{enumitem}
\usepackage{url}
\setlist[enumerate]{itemsep=0mm}
\usepackage{flushend}
\newcolumntype{C}[1]{>{\centering\arraybackslash}m{#1}}

\usepackage[framemethod=tikz]{mdframed}
\definecolor{mycolor}{rgb}{0.9, 0.0, 0.0}
\newmdenv[innerlinewidth=0.5pt, roundcorner=4pt,linecolor=mycolor,innerleftmargin=6pt,
innerrightmargin=6pt,innertopmargin=6pt,innerbottommargin=6pt]{rolebox}
\definecolor{mycolor1}{rgb}{0.0, 0.0, 0.9}
\newmdenv[innerlinewidth=0.5pt, roundcorner=4pt,linecolor=mycolor1,innerleftmargin=6pt,
innerrightmargin=6pt,innertopmargin=6pt,innerbottommargin=6pt]{commentbox}

\usepackage{bigstrut}
\usepackage{multirow}
\usepackage{amsmath}
\usepackage[caption=false]{subfig}
\usepackage[skip=0pt]{caption}
\usepackage{comment}

\makeatletter
\long\def\@IEEEtitleabstractindextextbox#1{\parbox{0.922\textwidth}{#1}}
\makeatother

\usepackage[rawfloats=true]{floatrow} 
\restylefloat{figure} 
\restylefloat{table} 

\newcommand{\Rev}[1]{\textcolor{black}{#1}}

\setlist[itemize]{leftmargin=*}

\usepackage{pgfplots}
\usepackage{tikz}
\pgfplotsset{compat=1.16,
	tick label style = {font = {\fontsize{40pt}{10pt}\selectfont}},
	label style = {font = {\fontsize{40pt}{10pt}\selectfont}},
	legend style = {font = {\fontsize{40pt}{10pt}\selectfont}},
	title style = {font = {\fontsize{40pt}{10pt}\selectfont}},
}
\usepackage{amsmath,pgfplots,gincltex}
\usepackage{amsmath, pgfplots, graphicx, siunitx, tikzscale}
\usepackage[version=3]{mhchem}

\usepackage[normalem]{ulem}
\usepackage{xcolor}
\newcommand\redsout{\bgroup\markoverwith{\textcolor{red}{\rule[0.5ex]{2pt}{0.4pt}}}\ULon}

\DeclareOldFontCommand{\bf}{\normalfont\bfseries}{\mathbf} 
\providecommand{\DIFadd}[1]{{\color{black} #1}} 
\providecommand{\DIFdel}[1]{} 
\providecommand{\DIFaddbegin}{} 
\providecommand{\DIFaddend}{} 
\providecommand{\DIFdelbegin}{} 
\providecommand{\DIFdelend}{} 
\providecommand{\DIFaddbeginFL}{} 
\providecommand{\DIFaddendFL}{} 
\providecommand{\DIFdelbeginFL}{} 
\providecommand{\DIFdelendFL}{} 
\newcommand{\DIFscaledelfig}{0.5}
\RequirePackage{settobox} 
\RequirePackage{letltxmacro} 
\newsavebox{\DIFdelgraphicsbox} 
\newlength{\DIFdelgraphicswidth} 
\newlength{\DIFdelgraphicsheight} 
\LetLtxMacro{\DIFOincludegraphics}{\includegraphics} 
\newcommand{\DIFaddincludegraphics}[2][]{{\color{blue}\fbox{\DIFOincludegraphics[#1]{#2}}}} 
\newcommand{\DIFdelincludegraphics}[2][]{
	\sbox{\DIFdelgraphicsbox}{\DIFOincludegraphics[#1]{#2}}
	\settoboxwidth{\DIFdelgraphicswidth}{\DIFdelgraphicsbox} 
	\settoboxtotalheight{\DIFdelgraphicsheight}{\DIFdelgraphicsbox} 
	\scalebox{\DIFscaledelfig}{
		\parbox[b]{\DIFdelgraphicswidth}{\usebox{\DIFdelgraphicsbox}\\[-\baselineskip] \rule{\DIFdelgraphicswidth}{0em}}\llap{\resizebox{\DIFdelgraphicswidth}{\DIFdelgraphicsheight}{
				\setlength{\unitlength}{\DIFdelgraphicswidth}
				\begin{picture}(1,1)
					\thicklines\linethickness{2pt} 
					{\color[rgb]{1,0,0}\put(0,0){\framebox(1,1){}}}
					{\color[rgb]{1,0,0}\put(0,0){\line( 1,1){1}}}
					{\color[rgb]{1,0,0}\put(0,1){\line(1,-1){1}}}
				\end{picture}
			}\hspace*{3pt}}} 
} 
\LetLtxMacro{\DIFOaddbegin}{\DIFaddbegin} 
\LetLtxMacro{\DIFOaddend}{\DIFaddend} 
\LetLtxMacro{\DIFOdelbegin}{\DIFdelbegin} 
\LetLtxMacro{\DIFOdelend}{\DIFdelend} 
\DeclareRobustCommand{\DIFaddbegin}{\DIFOaddbegin \let\includegraphics\DIFaddincludegraphics} 
\DeclareRobustCommand{\DIFaddend}{\DIFOaddend \let\includegraphics\DIFOincludegraphics} 
\DeclareRobustCommand{\DIFdelbegin}{\DIFOdelbegin \let\includegraphics\DIFdelincludegraphics} 
\DeclareRobustCommand{\DIFdelend}{\DIFOaddend \let\includegraphics\DIFOincludegraphics} 
\LetLtxMacro{\DIFOaddbeginFL}{\DIFaddbeginFL} 
\LetLtxMacro{\DIFOaddendFL}{\DIFaddendFL} 
\LetLtxMacro{\DIFOdelbeginFL}{\DIFdelbeginFL} 
\LetLtxMacro{\DIFOdelendFL}{\DIFdelendFL} 
\DeclareRobustCommand{\DIFaddbeginFL}{\DIFOaddbeginFL \let\includegraphics\DIFaddincludegraphics} 
\DeclareRobustCommand{\DIFaddendFL}{\DIFOaddendFL \let\includegraphics\DIFOincludegraphics} 
\DeclareRobustCommand{\DIFdelbeginFL}{\DIFOdelbeginFL \let\includegraphics\DIFdelincludegraphics} 
\DeclareRobustCommand{\DIFdelendFL}{\DIFOaddendFL \let\includegraphics\DIFOincludegraphics} 
\RequirePackage{listings} 
\RequirePackage{color} 
\lstdefinelanguage{DIFcode}{ 
	moredelim=[il][\color{white}\tiny]{\%DIF\ <\ }, 
	moredelim=[il][\sffamily\bfseries]{\%DIF\ >\ } 
} 
\lstdefinestyle{DIFverbatimstyle}{ 
	language=DIFcode, 
	basicstyle=\ttfamily, 
	columns=fullflexible, 
	keepspaces=true 
} 
\lstnewenvironment{DIFverbatim}{\lstset{style=DIFverbatimstyle}}{} 
\lstnewenvironment{DIFverbatim*}{\lstset{style=DIFverbatimstyle,showspaces=true}}{} 

\begin{document}
	\bstctlcite{IEEEexample:BSTcontrol}
	
	\title{Analyzing Human Observer Ability in Morphing Attack Detection - Where Do We Stand?}	
	\author{Sankini Rancha Godage$^\ddag$, Frøy Løvåsdal$^\dag$, Sushma Venkatesh$^\ddag$, \\ Kiran Raja$^\ddag$, Raghavendra Ramachandra$^\ddag$, Christoph Busch$^\ddag$ \\$^\ddag$Norwegian University of Science and Technology (NTNU), Norway \\ $^\dag$The Norwegian Police Directorate, Norway}

	\DIFdelbegin 
	\DIFdelend \DIFaddbegin \IEEEtitleabstractindextext{%
		\begin{abstract}
			Morphing attacks are based on the \Rev{technique} of digitally fusing two (or more) faces into one, with the final visage resembling the contributing faces. Morphed images not only pose a challenge to \Rev{Face-Recognition Systems} (FRS) but also challenge the experienced human observers due to high quality, postprocessing to eliminate any visible artifacts, and further, the printing and scanning process. \Rev{Few studies have concentrated on examining how people can recognize morphing attacks, even as several publications have examined the susceptibility of automated FRS and offered morphing attack detection (MAD) approaches. MAD approaches base their decisions either on a single image with no reference to compare against (Single-Image MAD (S-MAD)) or using a reference image (Differential MAD (D-MAD)).} \Rev{One prevalent misconception is that an} examiner’s or observer’s capacity for facial morph detection depends on their subject expertise, experience, and familiarity with the issue and that no works have reported the specific results of observers who regularly verify identity (ID) documents for their jobs. As human observers are involved in checking the ID documents having facial images, a lapse in their competence can have significant societal challenges. \Rev{To assess the observers’ proficiency, this research} first builds a new benchmark database of realistic morphing attacks from 48 different subjects, resulting in 400 morphed images. Unlike the previous works, we also capture images from Automated Border Control (ABC) gates to mimic the realistic border-crossing scenarios in the D-MAD setting with 400 probe images, to study the ability of human observers to detect morphed images. A new dataset of 180 morphing images is also \Rev{produced to research} human capacity in the S-MAD environment. In addition to creating a new evaluation platform to conduct S-MAD and D-MAD analysis, the study employs 469 observers for D-MAD and 410 observers for S-MAD who are primarily governmental employees from more than 40 countries, along with 103 subjects who are not examiners. The analysis offers \Rev{intriguing insights and highlights the lack of expertise and failure} to recognize a sizable number of morphing attacks by experienced observers. Human observers tend to detect morphed images to a lower accuracy as compared to the four automated MAD algorithms evaluated in this work. The results of this study are intended to aid in the development of training \Rev{programs that will prevent security failures while determining whether an image is bona fide or altered}.
		\end{abstract}
		}
	\DIFaddend 
	
	\maketitle
	\IEEEdisplaynontitleabstractindextext
	
	

	\IEEEraisesectionheading{\section{Introduction}}
	\label{sec:introduction}
	
	
	\begin{figure*}[h]
		\centering 
		\includegraphics[width=0.85\columnwidth]{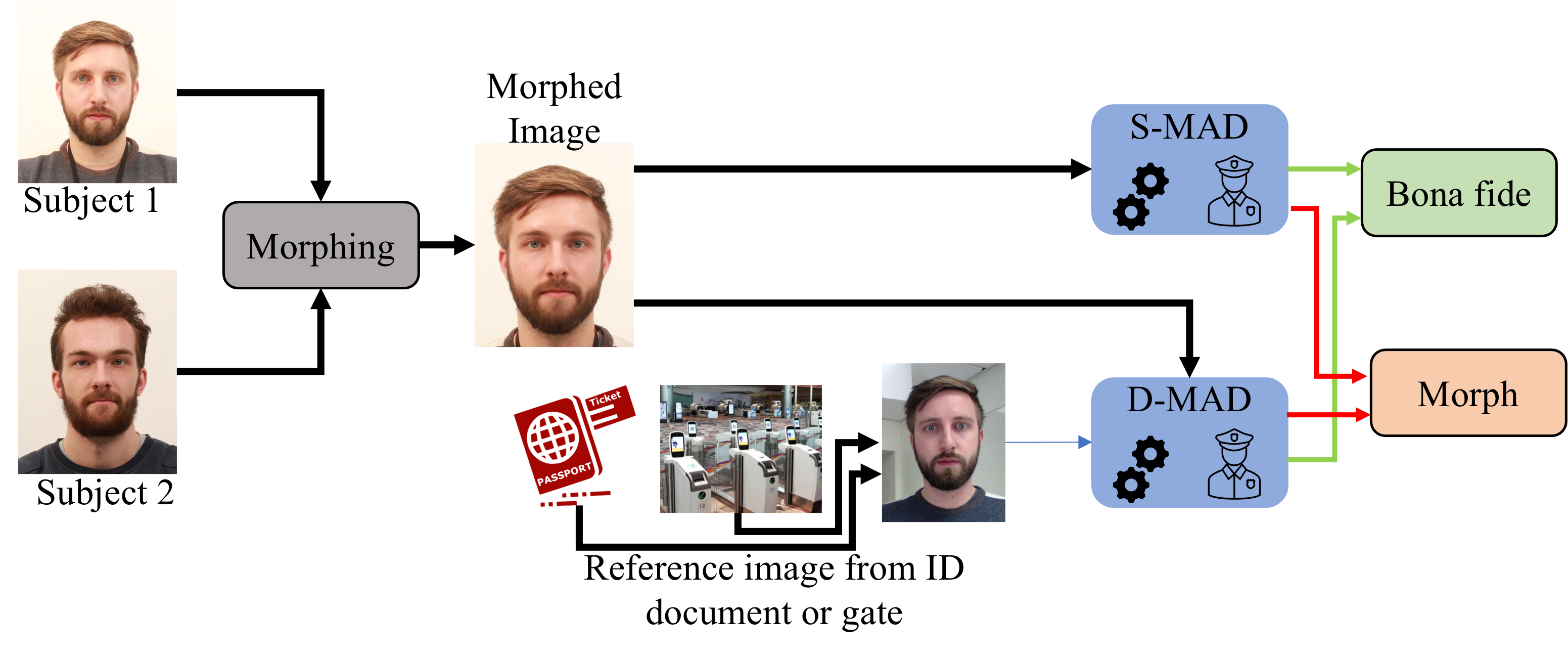}%
		\caption{Pipeline of Single-Image Morphing Attack Detection (S-MAD) and Differential MAD (D-MAD)}
		\label{fig:morphing-process-s-mad-d-mad}
	\end{figure*}

	Face recognition systems (FRS) are becoming increasingly common in identity management systems worldwide.  \DIFdelbegin \DIFdel{According}\DIFdelend \DIFaddbegin \DIFadd{A face image is mandated in all passports according }\DIFaddend to the guidelines of the International Civil Aviation Organization (ICAO), \DIFdelbegin \DIFdel{a face image is mandated in all passports, and in Europe, most countries }\DIFdelend \DIFaddbegin \DIFadd{and most European nations also }\DIFaddend have national ID cards \DIFdelbegin \DIFdel{, also using a }\DIFdelend \DIFaddbegin \DIFadd{with }\DIFaddend facial image as \DIFdelbegin \DIFdel{the primary modality.}\DIFdelend \DIFaddbegin \DIFadd{their main form of identification. }\DIFaddend 
	Most border controls globally already use automated facial recognition as the primary modality for verifying the traveler's identity. While the practice of using live enrolment is encouraged when issuing travel and/or identity documents, many countries still allow the applicants to upload or submit a passport photo. Such a practice is vulnerable with respect to morphed or otherwise manipulated facial images.

	A morphed image results from digitally combining two or more face images of two or more independent subjects, which usually resembles all the contributing subjects. \DIFdelbegin \DIFdel{A malicious actor can choose a different morphing factor for each contributing subject in creating a morph}\DIFdelend \DIFaddbegin \DIFadd{Each contributing subject to a morph can have a separate morphing factor selected by a malevolent actor}\DIFaddend . 
	\DIFdelbegin \DIFdel{The magnitude of the problem increases as a good quality morph not only circumvents }\DIFdelend \DIFaddbegin \DIFadd{A high-quality morph can not only evade }\DIFaddend automated FRS but also \DIFdelbegin \DIFdel{can fool }\DIFdelend \DIFaddbegin \DIFadd{trick }\DIFaddend a human observer (for \DIFdelbegin \DIFdel{instance}\DIFdelend \DIFaddbegin \DIFadd{example}\DIFaddend, a trained border \DIFdelbegin \DIFdel{guard}\DIFdelend \DIFaddbegin \DIFadd{officer}\DIFaddend ) in both the passport application process and automatic border control\DIFaddbegin \DIFadd{, which exacerbates the situation}\DIFaddend. While a novice attacker may create a morphed image with ghost \DIFdelbegin \DIFdel{artefacts}\DIFdelend \DIFaddbegin \DIFadd{artifacts}\DIFaddend , an expert attacker can create morphed images \DIFdelbegin \DIFdel{\Rev{carefully} }\DIFdelend \DIFaddbegin \DIFadd{carefully }\DIFaddend and then process them further to eliminate all the spurious traces (morphing related \DIFdelbegin \DIFdel{artefacts }\DIFdelend \DIFaddbegin \DIFadd{artifacts }\DIFaddend in images) in the image to enhance the appearance. \DIFdelbegin \DIFdel{Such a process usually results in a superior quality morph image that can challenge a }\DIFdelend \DIFaddbegin \DIFadd{A higher-quality morph image produced by such a method typically makes it difficult for a }\DIFaddend human observer to \DIFdelbegin \DIFdel{spot }\DIFdelend \DIFaddbegin \DIFadd{identify }\DIFaddend the morphs \cite{ferrara2014magic}.

	
	
	
	While many \DIFdelbegin \DIFdel{countries accept digital images, some countries also allow printed images which is later re-digitized }\DIFdelend \DIFaddbegin \DIFadd{nations only accept digital photos, some also accept printed photos that are then redigitized }\DIFaddend by the passport application officer\DIFdelbegin \DIFdel{leaving a chance for image to be morphed before printingit}\DIFdelend \DIFaddbegin \DIFadd{, giving the photo the chance to be altered before printing}\DIFaddend. Such a process eliminates most subtle cues present in the morphed image in \DIFaddbegin \DIFadd{the }\DIFaddend digital domain, making it challenging to detect the morphs. Additionally, \DIFdelbegin \DIFdel{morphed images can pose a greater challenge }\DIFdelend \DIFaddbegin \DIFadd{finding morphs in morphed photographs can be more difficult }\DIFaddend for human observers \DIFdelbegin \DIFdel{in detecting a morph }\DIFdelend than detecting impostors in \DIFdelbegin \DIFdel{regular }\DIFdelend \DIFaddbegin \DIFadd{typical }\DIFaddend facial images. Addressing such concerns of operational FRS, several works have started proposing various automated approaches for detecting morphing attacks \cite{venkatesh2021morphsurvey}. \DIFdelbegin \DIFdel{Generally, these Morphing Attack Detection }\DIFdelend \DIFaddbegin \DIFadd{To recognize the morphed images, these morphing attack detection }\DIFaddend (MAD) \DIFdelbegin \DIFdel{approaches rely on learning }\DIFdelend \DIFaddbegin \DIFadd{techniques often train a classifier on }\DIFaddend hand-crafted or deep features  \DIFdelbegin \DIFdel{to detect the morphed images by training a classifier }\DIFdelend \cite{venkatesh2021morphsurvey,raja2020morphing}. While one spectrum of solutions focuses on investigating machine-driven MAD, another set of works \DIFdelbegin \DIFdel{have focused on analysing }\DIFdelend \DIFaddbegin \DIFadd{has focused on analyzing }\DIFaddend the human observers\DIFdelbegin \DIFdel{' }\DIFdelend \DIFaddbegin \DIFadd{’ }\DIFaddend ability to detect morphing attacks. 
	\DIFdelbegin \DIFdel{\Rev{Complementing} the rapid progress in machine-driven MAD solutions, we focus on an understudied aspect of assessing human observers' ability }\DIFdelend \DIFaddbegin \DIFadd{We concentrate on a little-researched element of evaluating human observers’ capacity }\DIFaddend to detect the morphs \DIFaddbegin \DIFadd{to complement the rapid development of machine-driven MAD solutions}\DIFaddend . Unlike the previous studies  \cite{robertson2018detecting,ferrara2014magic,makrushin2017automatic}, we intend to study human ability on high-quality morphed images 
	\DIFdelbegin \DIFdel{\Rev{that} }\DIFdelend \DIFaddbegin \DIFadd{images that }\DIFaddend are carefully processed in a controlled manner to eliminate \DIFdelbegin \DIFdel{artefacts. Our strong assertion is that understanding the perceptual process of observers experienced in the }\DIFdelend \DIFaddbegin \DIFadd{artifacts. We firmly believe that by comprehending how observers perceive information during }\DIFaddend face and/or document examination\DIFdelbegin \DIFdel{can help build more accurate }\DIFdelend \DIFaddbegin \DIFadd{, we can better develop }\DIFaddend automated face morph detection software and \DIFdelbegin \DIFdel{design necessary training programs to improve the human ability }\DIFdelend \DIFaddbegin \DIFadd{create the training materials required to enhance people’s capacity }\DIFaddend for face morph detection.
	
	\subsection{Societal and Technological Aspects}
	\DIFdelbegin \DIFdel{Despite MAD algorithms providing a viable alternative to detect morphed images, in most of the }\DIFdelend \DIFaddbegin \DIFadd{Although MAD algorithms offer a practical option to detect altered photos, most }\DIFaddend ID control systems \DIFdelbegin \DIFdel{such as }\DIFdelend \DIFaddbegin \DIFadd{like }\DIFaddend border control, \DIFaddbegin \DIFadd{still involve }\DIFaddend a human expert \DIFdelbegin \DIFdel{is involved }\DIFdelend to make a \DIFdelbegin \DIFdel{decision}\DIFdelend \DIFaddbegin \DIFadd{judgment}\DIFaddend . A wrong decision by \DIFdelbegin \DIFdel{an }\DIFdelend \DIFaddbegin \DIFadd{a }\DIFaddend human can put the security of a country and society at risk in the worst case. \DIFaddbegin \DIFadd{To prevent ethical issues, a portion of the population should not always be treated as genuine or as malicious actors. }\DIFaddend On the other hand, \DIFdelbegin \DIFdel{any decision }\DIFdelend \DIFaddbegin \DIFadd{each choice made }\DIFaddend by an algorithm should be \DIFdelbegin \DIFdel{analyzed by human expert to avoid ethical challenges of not treating a subset of population always as bona fide or morphed images}\DIFdelend \DIFaddbegin \DIFadd{reviewed by a human}\DIFaddend . The clear interplay between the technology and human \DIFdelbegin \DIFdel{decision }\DIFdelend \DIFaddbegin \DIFadd{decision }\DIFaddend making aspect needs to be understood to avoid \DIFaddbegin \DIFadd{the }\DIFaddend justifiable use of technology. \DIFdelbegin \DIFdel{As a first step towards this, we motivate this work }\DIFdelend \DIFaddbegin \DIFadd{We are motivated }\DIFaddend to understand and analyze the \DIFdelbegin \DIFdel{ability }\DIFdelend \DIFaddbegin \DIFadd{skills }\DIFaddend of trained ID control experts working in various agencies as a first step in this direction. This \DIFaddend work is expected to serve as a \DIFdelbegin \DIFdel{basis for designing }\DIFdelend \DIFaddbegin \DIFadd{foundation for creating }\DIFaddend better training programs and policies for ID control experts to avoid \DIFaddbegin \DIFadd{the }\DIFaddend negative perception of MAD technologies \DIFaddbegin \DIFadd{by understanding the strengths and limitations}\DIFaddend .
	
	\subsection{Contributions}
	\DIFdelbegin \DIFdel{\Rev{Considering this inter-play of society and technology, we set out to answer if ID control experts (border guards, case handlers (for Passport, visas, ID, etc.), document examiners - 1st line, document examiner- 2nd line, expert document examiners- 3rd line and face comparison experts (Manual examination), 3rd line) have the ability to detect morphs in a reliable way \Rev{(Refer Table~\ref{tab:human-observer-expertise} for overview of expertise and their employment fields)}. The study is designed for  quantitative analysis of the results obtained from psycho-visual experiments.} 
	}\DIFdelend \DIFaddbegin \DIFadd{In light of this interaction between society and technology, we sought to determine if ID control specialists (border officers, case handlers) (for Passport, visas, ID, etc.), document examiners - 1st line, document examiner- 2nd line, expert document examiners- 3rd line and face comparison experts (Manual examination), 3rd line) have the ability to detect morphs reliably. The details of the experts recruited and their typical employment areas are provided in Table~\ref{tab:human-observer-expertise}. The study is intended for the quantitative examination of findings from psycho-visual experiments of these expert observer groups.
	}\DIFaddend 
	
	\Rev{We pose five critical questions to help analyze the findings} as listed below:
	\begin{itemize}
		\item How good are ID document examiners at detecting morphing attacks?
		\item Are human observers with certain types of training better than others at detecting morphing attacks?
		\item Does long experience working in a certain field (for instance, facial examination in ID control) positively impact MAD detection performance?
		\item \Rev{Are expert observers with training or experience in checking identity/identity documents better than those without training?}
		\item How do machine-based MAD algorithms perform compared to human observers \Rev{in} D-MAD and S-MAD?
	\end{itemize}
	
		\begin{table*}[htbp]
		\color{black}
		\centering
		\caption{\Rev{Expertise domain of human observers recruited for the experiments in this work.}}
		\resizebox{0.8\textwidth}{!}{
		\begin{tabular}{ll}
			\hline
			\hline
			Line of work & Expertise and tasks  \\
			\hline
			\hline
			\multicolumn{1}{l}{\multirow{3}[0]{*}{Border guard 1st line}} &  Check and ascertain whether travelers are eligible to cross the border \\      
			&  Perform basic document examination, basic facial comparison, and basic interview \\      
			&  Carry out checks against relevant databases and interpret the results \\
			\hline
			\multicolumn{1}{l}{\multirow{4}[0]{*}{Border guard 2nd line}} &  Support 1st line in their tasks \\ 
			&  Handle more complex or time-consuming cases \\      
			&  Perform more advanced document examination, facial comparison, interview \\      
			&  Decide on whether the traveler may be allowed to cross the border \\
			\hline
			&  Receive, process and consider applications for passports, ID cards, visas, \\
			& residence permits, national personal ID number, etc. \\
			&  Provide information about the application procedure according \\
			& to the legal framework, policy, etc. \\     
			Case handler - passport, &  Receive and register applications \\      
			visas, ID, etc. - 1st line &  Perform basic document examination, basic facial comparison, and basic interview \\      
			&  Carry out checks against relevant databases and interpret the results \\      
			&  Decide on the application / refer the application to a 2nd line officer \\
			\hline
			&  Support 1st line in their tasks \\      
			Case handler - passport, &  Handle more complex or time-consuming cases \\      
			visas, ID, etc. - 2nd line &  Perform more advanced checks of documents and person \\      
			&  Decide on applications \\
			\hline
			& Support 1st and 2nd line in checks of document and person in various \\
			& types of cases. e.g. passports, visas, border crossings. \\      
			&  Perform advanced document examination \\      
			&  Perform advanced facial comparison \\      
			ID Expert &  Perform in-depth interviews \\      
			(Embassies, Police  &  Decide on applications \\      
			Districts, ID Offices, etc.) 
			&  Produce and provide training to 1st and 2nd lines in the document \\ 
			& examination, facial comparison, impostor detection, etc. \\      
			&  Give evidence in court \\
			\hline
			&  Support 1st line in comparing documents \\      
			Document examiner &  Perform advanced document examinations \\      
			2nd line&  Produce and provide training \\      
			&  Give evidence in court \\
			\hline
			&  Support 1st and 2nd line in comparing faces \\      
			Document examiner &  Perform specialized face examinations/comparisons \\      
			3rd line (Central unit)&  Produce and provide training \\      
			&  Give evidence in court \\
			\hline
			 &  Support 1st and 2nd line in comparing faces \\
			Face examiner &  Perform specialized face examinations/comparisons \\
			(Central unit) &  Produce and provide training \\      
			&  Give evidence in court \\
			\hline
			&  Support 1st and 2nd line in comparing fingerprints \\ 
			Fingerprint examiner &  Perform specialized fingerprint examinations/comparisons \\      
			3rd line (Central unit) &  Produce and provide training \\      
			&  Give evidence in court \\
			\hline

			\hline
			\hline
		\end{tabular}%
	}
		\label{tab:human-observer-expertise}%
	\end{table*}%

	\DIFdelbegin \DIFdel{\Rev{In the course of answering the question, we contribute a new database and an evaluation platform.} As the first contribution, we create }\DIFdelend \DIFaddbegin \DIFadd{We develop }\DIFaddend a new database of morphed \DIFdelbegin \DIFdel{images referred to as }\DIFdelend \DIFaddbegin \DIFadd{images called the }\DIFaddend Human Observer Analysis Morphing Image Database \DIFaddbegin \DIFadd{as our initial contribution }\DIFaddend (HOMID). The newly created database consists of images from \DIFdelbegin \DIFdel{$48$ }\DIFdelend \DIFaddbegin \DIFadd{48 }\DIFaddend unique bona fide subjects and correspondingly \DIFdelbegin \DIFdel{$400$ }\DIFdelend \DIFaddbegin \DIFadd{400 }\DIFaddend morphed images created by a combination of two resembling subjects (in terms of age, gender, ethnicity\DIFaddbegin \DIFadd{, }\DIFaddend and skin color) with \DIFdelbegin \DIFdel{human expert intervention. The morphed images are further post-processed to eliminate any artefacts by employing two experts (researchers highly familiar with morphing) to create a realistic and high quality morphed image set. \Rev{As an additional measure for realistic evaluation}}\DIFdelend \DIFaddbegin \DIFadd{a human expert’s intervention. Two specialists (researchers with extensive morphing experience) further postprocessed the morphed images to remove any artifacts and provided the curated set of realistic}\DIFaddend , \DIFaddbegin \DIFadd{high-quality morphed images. As an additional measure for realistic evaluation, }\DIFaddend the newly created database is supplemented with corresponding images after printing and scanning. \DIFdelbegin \DIFdel{Along with morphed images, the database also contains images captured from an Automated Border Control }\DIFdelend \DIFaddbegin \DIFadd{The database also includes photographs that were taken as a probe set from an automated border control }\DIFaddend (ABC) gate \DIFdelbegin \DIFdel{as a probe set. The first set is referred to }\DIFdelend \DIFaddbegin \DIFadd{in addition to morphed images. The D-MAD situation can be evaluated using the first collection, known }\DIFaddend as the HOMID-D dataset\DIFdelbegin \DIFdel{suitable for evaluating the D-MAD scenario. \Rev{Another set of 30 bona fide subjects from the FRGC v2 dataset \cite{phillips2005overview} is further used to generate 30 morphed images}. }\DIFdelend \DIFaddbegin \DIFadd{. Another set of 30 bona fide subjects from the FRGC v2 dataset \mbox{
			\cite{phillips2005overview} }\hspace{0pt}
		is further used to generate 30 morphed images. }\DIFaddend All the morphed images are further carefully \DIFdelbegin \DIFdel{post-processed}\DIFdelend \DIFaddbegin \DIFadd{postprocessed}\DIFaddend , printed, and scanned. The second \DIFdelbegin \DIFdel{sub-set }\DIFdelend \DIFaddbegin \DIFadd{subset }\DIFaddend is referred to as the HOMID-S dataset \DIFaddbegin \DIFadd{is }\DIFaddend suitable for the S-MAD scenario and consists of \DIFdelbegin \DIFdel{$90$ }\DIFdelend \DIFaddbegin \DIFadd{90 }\DIFaddend digital images and \DIFdelbegin \DIFdel{$90$ }\DIFdelend \DIFaddbegin \DIFadd{90 }\DIFaddend printed-and-scanned images. \DIFdelbegin \DIFdel{Both database subsets comprise }\DIFdelend \DIFaddbegin \DIFadd{To produce realistic morph attacks, both database subsets contain }\DIFaddend a good diversity \DIFdelbegin \DIFdel{from different ethnicities, age groups and gender to create realistic morph attacks \Rev{(Please see Section~\ref{sec:database})}.
	}\DIFdelend \DIFaddbegin \DIFadd{of people of various ages, genders, and races (Please see Section~\ref{sec:database}).
	}\DIFaddend 
	
	\DIFdelbegin \DIFdel{In order to facilitate the large scale }\DIFdelend \DIFaddbegin \DIFadd{To facilitate the large-scale }\DIFaddend psycho-visual \DIFdelbegin \DIFdel{experiments}\DIFdelend \DIFaddbegin \DIFadd{experiments}\DIFaddend , we also present a customized Human Observer Evaluation Platform for morph image detection. The new evaluation \DIFdelbegin \DIFdel{platform mimics the realistic operational scenario where the images were provided to observers to determine if the image is bona fide or morphed}\DIFdelend \DIFaddbegin \DIFadd{tool simulates a practical operational situation in which observers were given photographs to assess whether they were bona fide or morphed}\DIFaddend. Furthermore, the platform is designed following the strict guidelines of \DIFdelbegin \DIFdel{GDPR }\DIFdelend \DIFaddbegin \DIFadd{General Data Protection Regulation (GDPR) }\DIFaddend to protect and preserve participants\DIFdelbegin \DIFdel{' }\DIFdelend \DIFaddbegin \DIFadd{’ }\DIFaddend privacy with full \DIFdelbegin \DIFdel{considerations }\DIFdelend \DIFaddbegin \DIFadd{consideration }\DIFaddend to the anonymity of participants. \DIFdelbegin \DIFdel{\Rev{The participants were asked for the consent before beginning the experiment in accordance to ethical guidelines laid down by the university.} }\DIFdelend \DIFaddbegin \DIFadd{In compliance with the university’s established ethical standards, the participants were contacted before the experiment began to obtain their consent }\DIFaddend \Rev{(refer Section~\ref{sec:web-platform} for details)}.
	
	While the earlier works focused on analyzing human \DIFdelbegin \DIFdel{observers' }\DIFdelend \DIFaddbegin \DIFadd{observers’ }\DIFaddend ability to detect morphing attacks, observers have been predominantly selected \DIFdelbegin \DIFdel{in a random manner }\DIFdelend \DIFaddbegin \DIFadd{randomly }\DIFaddend without carefully choosing ID control experts. In this study, we \DIFdelbegin \DIFdel{employ }\DIFdelend \DIFaddbegin \DIFadd{use }\DIFaddend a total of \DIFdelbegin \DIFdel{$469$ observers with a specific domain }\DIFdelend \DIFaddbegin \DIFadd{469 observers who have a specific area }\DIFaddend of expertise in face-based identity verification, \DIFdelbegin \DIFdel{such as }\DIFdelend \DIFaddbegin \DIFadd{including }\DIFaddend border guards, passport officers, visa officers, ID experts, \DIFaddbegin \DIFadd{and }\DIFaddend forensic examiners (face, document\DIFaddbegin \DIFadd{, }\DIFaddend and fingerprints), to name a few, and \DIFdelbegin \DIFdel{benchmark it against observers ($103$) }\DIFdelend \DIFaddbegin \DIFadd{compare their results to those of (100) observers }\DIFaddend who are not familiar with facial comparison or forensic examination of any kind. Such an observer base of experts and \DIFdelbegin \DIFdel{non-experts }\DIFdelend \DIFaddbegin \DIFadd{nonexperts }\DIFaddend in our experiments also provided an insight into the role of vast experience in face comparison\footnote{Face comparison in this context refers to comparing two face images to make a decision on mated or non-mated pair of subjects.} 
	and morphed image detection, which is currently missing in \DIFdelbegin \DIFdel{the }\DIFdelend the state-of-the-art studies \DIFdelbegin \DIFdel{. \Rev{The expertise of human observers and their typical employment is provided in Table~\ref{tab:human-observer-expertise}.} }\DIFdelend \DIFaddbegin \DIFadd{\Rev{Table~\ref{tab:human-observer-expertise} lists the qualifications of human observers and their typical jobs.} }\DIFaddend Furthermore, we \DIFdelbegin \DIFdel{assert that such }\DIFdelend \DIFaddbegin \DIFadd{contend that using human observers as }\DIFaddend a realistic benchmark \DIFdelbegin \DIFdel{of human observers can lead to designing better training programs for detecting morphs, and the knowledge }\DIFdelend \DIFaddbegin \DIFadd{can help create better training schemes for morph detection and that the knowledge gained }\DIFaddend from such a study can \DIFdelbegin \DIFdel{further be used to develop stronger }\DIFdelend \DIFaddbegin \DIFadd{be applied to create more powerful }\DIFaddend machine-based MAD algorithms.

	\DIFdelbegin \DIFdel{In an operational setting, human observers can be presented }\DIFdelend \DIFaddbegin \DIFadd{Human observers may be given just one image to decide whether a presented image is morphed in an operational scenario (for instance, passport application), or they may be given }\DIFaddend a suspected image \DIFdelbegin \DIFdel{with }\DIFdelend \DIFaddbegin \DIFadd{and }\DIFaddend a reference image from a \DIFdelbegin \DIFdel{trusted source ( for instance, }\DIFdelend \DIFaddbegin \DIFadd{reliable source ( such as }\DIFaddend ABC gate),\DIFdelbegin \DIFdel{or an observer is just presented with one image to decide on a morphing attack }\DIFdelend . \DIFdelbegin \DIFdel{In order to }\DIFdelend \DIFaddbegin \DIFadd{To }\DIFaddend account for both scenarios, we study the performance of human observers under the D-MAD scenario and S-MAD. \DIFdelbegin \DIFdel{Specifically, for D-MAD human observer analysis, we employ the images captured }\DIFdelend \DIFaddbegin \DIFadd{In particular, we use the images acquired }\DIFaddend from an ABC gate as a trusted source image \DIFaddbegin \DIFadd{for D-MAD human observer analysis }\DIFaddend and ask the observers to \DIFdelbegin \DIFdel{determine if }\DIFdelend \DIFaddbegin \DIFadd{judge whether }\DIFaddend the second suspected image (\DIFdelbegin \DIFdel{unknown-capture}\DIFdelend \DIFaddbegin \DIFadd{unknown capture}\DIFaddend ) is morphed or not \DIFdelbegin \DIFdel{, }\DIFdelend using the HOMID-D subset. While in the S-MAD experiments, we employ the images from both the morphed set and bona fide set obtained from the HOMID-S dataset. As a third \DIFdelbegin \DIFdel{contribution, we also benchmark the human observer ability against }\DIFdelend \DIFaddbegin \DIFadd{contribution, we compare human observer performance to }\DIFaddend machine-driven MAD algorithms to \DIFdelbegin \DIFdel{illustrate where humans stand against machines in detecting }\DIFdelend \DIFaddbegin \DIFadd{show how well humans and machines compare when it comes to identifying }\DIFaddend morphing attacks.
	
	The key contributions of this work can therefore be listed as:
	\begin{itemize}
		\item \textbf{Human Observer Analysis Face Morph Database:}  Presents a new database (HOMID) for morphed \DIFdelbegin \DIFdel{images }\DIFdelend \DIFaddbegin \DIFadd{photos }\DIFaddend and ABC gate images for \DIFdelbegin \DIFdel{analyzing human observer ability in detecting morphed images. In addition, the dataset has images to analyze both }\DIFdelend \DIFaddbegin \DIFadd{studying how well humans can spot morphed images. The dataset also includes images for analyzing }\DIFaddend D-MAD and S-MAD scenarios.
		\item \textbf{Human Observer Evaluation Platform:} A new evaluation platform mimicking the realistic operational scenario is designed where the images are provided to officers to determine if the image is bona fide or morphed.
		\item \textbf{Human Observer Detection Performance:} \DIFaddbegin \DIFadd{The study is based on a sizable number of observers (469 in the D-MAD trials and 410 in the S-MAD experiment) who regularly verify identification documents and/or compare biometric data to present a realistic benchmark of human observers}\DIFaddend \footnote{A total of 790 observers have  participated in the experiments in different categories. However, we only consider the observers who have completed the experiments to conduct the analysis in analyze this work.}. The study also benchmarks it against the novice observers with no training or experience in facial comparison or document examination to establish the benefit of domain knowledge.
		\item \textbf{Human Observer versus Machine Detection Performance:} \DIFdelbegin \DIFdel{Presents a holistic analysis }\DIFdelend \DIFaddbegin \DIFadd{To determine the advantages of each peer, a comprehensive examination }\DIFaddend of the performance of four machine-driven MAD algorithms \DIFdelbegin \DIFdel{against human observers to identify the strengths of each peer}\DIFdelend \DIFaddbegin \DIFadd{vs human observers is presented}\DIFaddend . Two algorithms in S-MAD and two algorithms in D-MAD benchmarked in \DIFaddbegin \DIFadd{the }\DIFaddend NIST FRVT MORPH challenge \cite{NistFrvtMorph} are evaluated on the newly constructed database to benchmark the performance of human observers against algorithms.
	\end{itemize}
	
	\DIFdelbegin \DIFdel{In the rest of this paper, Section~\ref{sec:background-works} presents the preliminary concepts related to }\DIFdelend \DIFaddbegin \DIFadd{The preliminary ideas for }\DIFaddend morph image detection and \DIFdelbegin \DIFdel{related }\DIFdelend \DIFaddbegin \DIFadd{associated }\DIFaddend works on human observer analysis \DIFaddbegin \DIFadd{are presented in Section ~\ref{sec:background-works}}\DIFaddend. In Section~\ref{sec:methodology}, Section~\ref{sec:database} presents the details of a newly created database for human observer analysis and Section~\ref{sec:web-platform} presents the details on evaluation platform. \DIFaddbegin \DIFadd{Section~\ref{ssec:d-mad-experiment-design} and \ref{ssec:s-mad-experiment-design} present the details of two experiment corresponding to D-MAD and S-MAD respectively.}\DIFaddend \DIFdelbegin \DIFdel{Section~\ref{sec:human-observer-evaluation} presents the results }\DIFdelend \DIFaddbegin \DIFadd{ The findings }\DIFaddend of the human observer detection performance \DIFdelbegin \DIFdel{along with the details of participant background and expertise}\DIFdelend \DIFaddbegin \DIFadd{are presented in Section~\ref{sec:human-observer-evaluation} together with information on the qualifications and experience of the participants along with the details on recruitment of observers. Section~\ref{ssec:d-mad-experiment-design} and Section~\ref{ssec:s-mad-experiment-design} present the details of two different experiments in this work. }\DIFaddend Finally, \DIFdelbegin \DIFdel{section}\DIFdelend \DIFaddbegin \DIFadd{Section}\DIFaddend ~\ref {sec:findings-discussion} discusses the key findings from the conducted analysis and further provides conclusive remarks in Section~\ref{sec:conclusion} before listing the potential future works.

	\section{Background and state of the art}
	\label{sec:background-works}
	\DIFdelbegin \DIFdel{This section presents a set of related works analyzing human observer ability in general face recognition \Rev{and morphing attack detection. The summary is further supplemented} }\DIFdelend \DIFaddbegin \DIFadd{A collection of linked papers that analyze human observers’ capacity for MAD are presented in this section. All representative efforts for automated morphing detection are also added to the summary. }\DIFaddend 
	
	\begin{table*}[htp]
		\centering
		\resizebox{0.65\linewidth}{!}{%
			\begin{tabular}{lcccc}
				\hline
				\hline
				Reference & Post-processed  & Digital & Images & Number\\
				& morphed & Print-scan & & observers \bigstrut\\ \hline
				\hline
				\multicolumn{5}{c}{D-MAD}\\
				\hline
				M. Ferrara et al.\cite{ferrara2014magic} 2016 & No & Digital & Bona fide 8 & Experts 44  \\ 
				&  &  & Morphed 20 & Non experts 543   \bigstrut\\ \hline
				DJ Robertson.\cite{robertson2017fraudulent} 2017 & No & Digital & Bona fide 14 & Non experts 28 \\ 
				&  &  & Morphed 35 &  Trained non experts 42 \bigstrut\\ \hline
				Robertson et al.\cite{robertson2018detecting} 2018 & No & Digital & Bona fide 14 & Non experts 80 \\
				&  &  & Morphed 35 &  \bigstrut\\ \hline
				Kramer et al.\cite{kramer2019face} 2019 & Yes & Digital & Bona fide 20 & Trained non experts 80  \\ 
				&  &  & Morphed 20 &  \bigstrut\\ \hline
				A Makrushin et al. \cite{makrushin2020simulation} 2020 & No & Digital & Bona fide 5& Experts 21 \\ 
				&  &  & Morphed 10 & Non experts 168   \bigstrut\\
				&  &  &  Impostor 5 & \bigstrut\\
				\hline
				A Makrushin et al. \cite{makrushin2020simulation} 2020 & No & Digital & Bona fide 10 & Experts 21\\ 
				&  &  and videos & Morphed 10 & Non experts 168 \\
				&  &  &  Impostor 5 & \bigstrut\\
				\hline
				Zhang et al. \cite{zhang2021mipgan} 2021 & No & Digital & Bona fide 15 & Experienced 5\\ 
				&  &  Printed & Morphed 75 & Non experts 10 \bigstrut\\
				\hline 
				\rowcolor{gray!20}
				&  & \textbf{Digital} & \textbf{Bona fide - 416} & \textbf{Expert Examiners * - 469}\\
				\rowcolor{gray!20}
				\textbf{Ours} &  \textbf{Yes}   & \textbf{Post-processed} & \textbf{Morphed - 384} & \textbf{Non-Experts - 103}\\
				\rowcolor{gray!20}
				&     & \textbf{Printed}        & & \\
				\hline
				\hline
				\multicolumn{5}{c}{S-MAD}\\
				\hline
				\hline
				A Makrushin et el.\cite{makrushin2017automatic} 2017 & No & Printed & Bona fide - 7 & Non experts 42 \\
				&  &  & Morphed 23 &  \bigstrut\\ \hline
				A Makrushin et al. \cite{makrushin2020simulation} 2020 & No & Digital & Bona fide 5& Experts 21  \\ 
				&  &  & Morphed 10 & Non experts 168 \bigstrut\\
				\hline
				Nightingale et al. \cite{nightingale2021perceptual} 2021 & No & Digital & Bona fide 54 & Non experts 100\\
				&  &  & Morphed 54 & (crowd-sourced) \bigstrut\\
				\hline
				Zhang et al. \cite{zhang2021mipgan} 2021 & Yes & Digital & Bona fide 15 & Experienced 14\\ 
				&  &  Printed & Morphed 75 & Non experts 42 \\
				&  &  Post-processed &  &  \bigstrut\\
				\hline
				\hline
				\rowcolor{gray!20}
				&  & \textbf{Digital} & \textbf{Bona fide - 60} & \textbf{Expert Examiners * - 410}\\
				\rowcolor{gray!20}
				\textbf{Ours} &  \textbf{Yes}   & \textbf{Post-processed} & \textbf{Morphed - 120} & \textbf{Non-Experts - 0}\\
				\rowcolor{gray!20}
				&     & \textbf{Printed}        & & \\
				\hline
				\hline
			\end{tabular}%
		}
		\caption{Detailed analysis of existing state-of-art for human observer analysis for MAD. *Experts examiners in the context of this work include case handlers, examiners and expert examiners.}
		\label{tab:sota-human-mad}
	\end{table*}
	
	\subsection{Human observers for face morphing attack detection}
	\DIFdelbegin \DIFdel{Several works were conducted to analyse human observers' ability to detect }\DIFdelend \DIFaddbegin \DIFadd{Many studies have been conducted to examine how well humans can recognize }\DIFaddend digital morph attacks. Robertson et al. \cite{robertson2017fraudulent} studied the morph attack detection by human observers, in addition to \DIFdelbegin \DIFdel{the accuracy of smartphone face recognition systems}\DIFdelend \DIFaddbegin \DIFadd{face recognition system on a smartphone}\DIFaddend . Robertson et al. \cite{robertson2017fraudulent} conducted three sets of experiments using the Glasgow Face Matching Test - GFMT \cite{burton2010glasgow}. \DIFdelbegin \DIFdel{\Rev{The findings of the study was based on 19 female and 30 male observers}}\DIFdelend \DIFaddbegin \DIFadd{\Rev{The findings of the study were based on 19 female and 30 male observers}}\DIFaddend .
	
	\DIFdelbegin \DIFdel{In the experimental set-up, the first }\DIFdelend \DIFaddbegin \DIFadd{The initial }\DIFaddend face image of each \DIFdelbegin \DIFdel{subject was cropped \Rev{(without a torso) and} }\DIFdelend \DIFaddbegin \DIFadd{individual was cropped \Rev{(to remove the torso) and} }\DIFaddend pasted onto a white background \DIFdelbegin \DIFdel{of size }\DIFdelend \DIFaddbegin \DIFadd{that measured  }\DIFaddend $14cm \times 10.5 cm$ \DIFaddbegin \DIFadd{for the experimental setup}\DIFaddend . The second image was embedded in a \DIFdelbegin \DIFdel{\Rev{frame similar} as }\DIFdelend \DIFaddbegin \DIFadd{frame similar to }\DIFaddend a UK passport (same size). \DIFdelbegin \DIFdel{Five morphed images were created for each }\DIFdelend \DIFaddbegin \DIFadd{Each }\DIFaddend combination of subjects \DIFdelbegin \DIFdel{, and }\DIFdelend \DIFaddbegin \DIFadd{resulted in the creation of five morphed images, with }\DIFaddend morphing factors starting \DIFdelbegin \DIFdel{from }\DIFdelend \DIFaddbegin \DIFadd{at }\DIFaddend 90\% from the first subject and 10\% from the second subject \DIFdelbegin \DIFdel{was used with }\DIFdelend \DIFaddbegin \DIFadd{and }\DIFaddend a step size of 20\%. In Experiment-1, \DIFdelbegin \DIFdel{\Rev{the observers were asked to confirm whether the images were mated images for the shown trusted image and the second non-trusted image without the prior knowledge of manipulations}.
	}\DIFdelend \DIFaddbegin \DIFadd{the observers were asked to confirm whether the images were mated images for the shown trusted image and the second nontrusted image without the prior knowledge of manipulations.
	}

	\DIFaddend 
	\DIFaddbegin \DIFadd{Images of the same person or a modified image were used to randomize the tests. Each trial had two images, and the observer had to choose whether the shown images were the same person or not. Each participant in this experiment underwent 49 trials, with 35 morphed pair trials, 7 mated comparison trials, 7 nonmated comparison trials, and 49 randomly chosen trials from the image collection. 25 female and three male observers participated in this experiment where the false acceptance rate of 68\% was noted for morphed images (morphing factor of 50\%). It should be noted that the observers were not asked to affirm whether they had been morphed; instead, they were just given the choice of selecting “match” or “nonmatch.”}\DIFaddend 
	
	With a similar \DIFdelbegin \DIFdel{set-up}\DIFdelend \DIFaddbegin \DIFadd{setup}\DIFaddend , another experiment (referred \DIFaddbegin \DIFadd{to }\DIFaddend as Experiment-2) was conducted. \DIFdelbegin \DIFdel{Unlike }\DIFdelend \DIFaddbegin \DIFadd{In contrast to }\DIFaddend Experiment-1, the \DIFdelbegin \DIFdel{observers were \Rev{briefed about face morphing attacks and provided hint to identify the face morphs before starting the experiment}, and }\DIFdelend \DIFaddbegin \DIFadd{observers were informed about face morphing attacks and given tips to recognize the morphs both before and }\DIFaddend during the test. The observers were allowed to choose between three options \DIFdelbegin \DIFdel{\Rev{such as}  "match","mismatch",and "morph".Thirty-four }\DIFdelend \DIFaddbegin \DIFadd{such as “match,” “mismatch,” and “morph.” In this study, which included 34 }\DIFaddend female and 8 male observers\DIFdelbegin \DIFdel{participated in this study, and }\DIFdelend \DIFaddbegin \DIFadd{, }\DIFaddend the authors concluded that \DIFdelbegin \DIFdel{the brief introduction to morphing could increase }\DIFdelend \DIFaddbegin \DIFadd{a quick exposure to morphing might improve }\DIFaddend morph detection accuracy. Specifically, they noted that the morph detection error rate was significantly lower (21\%) than \DIFdelbegin \DIFdel{the first experiment }\DIFdelend \DIFaddbegin \DIFadd{in the first experiment }\DIFaddend (32\%). \DIFdelbegin \DIFdel{\Rev{In the Experiment-3, mobiles phones were randomly enrolled with morphed or bona fide images of the observers and were asked to unlock them without any prior knowledge. The FRS on mobile phones did not perform well even for mated comparison trials (91.8\% false rejects), and it did not allow false accepts (0\% false accepts). The authors concluded that it was safe to use mobile phones under face morphing attacks, however, one can notice high false rejects at the expense of being robust to morphing attacks.}
	}\DIFdelend \DIFaddbegin \DIFadd{In Experiment 3, participants were asked to randomly select and unlock mobile phones that had altered or real photos of the observers. The FRS on mobile phones did not perform well even for mated comparison trials (91.8\% false rejects), and it did not allow false accepts (0\% false accepts). The authors noted significant false rejects at the expense of being resilient to morphing assaults. 
	}\DIFaddend 
	
	In 2016, Ferrara et al. \cite{ferrara2016effects} studied face morph attack detection with both human observers and algorithms. \DIFdelbegin \DIFdel{80 face morphs were generated using the free GNU Image Manipulation Program }\DIFdelend Using the free GIMP Animation Package v2.6 \cite{gnu2020gimp} and the GIMP Animation Package v2.8 (GAP) \cite{gnu2020gimp2} software, \DIFaddbegin \DIFadd{80 face morphs were produced. }\DIFaddend The observers were \DIFdelbegin \DIFdel{shown two face images in each trial and the observers were asked to indicate if the morphed images }\DIFdelend \DIFaddbegin \DIFadd{instructed to report whether the morphing photographs }\DIFaddend matched against a trusted bona fide (i.e.\DIFaddbegin \DIFadd{, }\DIFaddend original) face image \DIFdelbegin \DIFdel{. }\DIFdelend \DIFaddbegin \DIFadd{in each trial, which included two face images. In this experiment, there were 543 nonexpert observers and }\DIFaddend Forty-four experts (border guards)\DIFdelbegin \DIFdel{and 543 non-expert observers participated in this experiment}\DIFdelend . The results showed that morphed images were not correctly detected (i.e., \DIFdelbegin \DIFdel{\Rev{classified} }\DIFdelend \DIFaddbegin \DIFadd{classified }\DIFaddend as a bona fide image).
	
	\DIFaddbegin \DIFadd{In four studies akin to those conducted by Robertson et al. \mbox{
			\cite{robertson2017fraudulent}, }\hspace{0pt}
	}\DIFaddend Kramer et al.\cite{kramer2019face} \DIFdelbegin \DIFdel{studied }\DIFdelend \DIFaddbegin \DIFadd{investigated }\DIFaddend the human and \DIFdelbegin \DIFdel{\Rev{algorithmic morph detection performance with four experiments similar} to Robertson et al.\mbox{
			\cite{robertson2017fraudulent}}\hspace{0pt}
		, but using }\DIFdelend \DIFaddbegin \DIFadd{algorithmic morph detection performance while utilizing }\DIFaddend a high-quality morph database. \DIFdelbegin \DIFdel{\Rev{Overall morph detection accuracy of experts (56\%) was higher than non-experts (52.2\%) as analysed using signal detection measures}. After }\DIFdelend \DIFaddbegin \DIFadd{Overall morph detection accuracy of experts (56\%) was higher than nonexperts (52.2\%) as analyzed using signal detection measures. There was no discernible performance gain in morph detection following }\DIFaddend the morph detection training \DIFdelbegin \DIFdel{program, there was no notable performance \Rev{improvement in detecting morphs. Further, single image based morphing attack detection (S-MAD) was studied using 120 images and forty participants (21 women)}}\DIFdelend \DIFaddbegin \DIFadd{session. Further, single image-based morphing attack detection (S-MAD) was studied using 120 images and 40 participants (21 women)}\DIFaddend . Kramer et al.\cite{kramer2019face} \DIFdelbegin \DIFdel{analysed the data using hit (correct }\DIFdelend \DIFaddbegin \DIFadd{used hit (right }\DIFaddend morph identification) and false alarm \DIFdelbegin \DIFdel{\Rev{(bona fide identified as morphed image)}}\DIFdelend \DIFaddbegin \DIFadd{to analyze the data (bona fide identified as morphed image)}\DIFaddend . There were no notable differences between the experts and \DIFdelbegin \DIFdel{non-experts }\DIFdelend \DIFaddbegin \DIFadd{nonexperts }\DIFaddend in morph detection accuracy for S-MAD. \DIFdelbegin \DIFdel{\Rev{In addition, it was noted that no reference morph detection (S-MAD) algorithms performed better than human observers. The study also concluded that it is difficult} }\DIFdelend \DIFaddbegin \DIFadd{No reference morph detection (S-MAD) algorithms outperformed human observers, it was also noted. The study also concluded that it is difficult }\DIFaddend to detect morphed images without using the cues of watermarks, hidden printed letters, ink, etc. \cite{kramer2019face}. 
	
	Ferrara et al. \cite{ferrara2014magic} studied face morph attacks under \DIFdelbegin \DIFdel{Automated Border Control }\DIFdelend \DIFaddbegin \DIFadd{automated border control }\DIFaddend (ABC) gate settings. \DIFdelbegin \DIFdel{Eighteen morphed images were provided to trained face comparison experts, where it was established that all the images went undetected as morphed }\DIFdelend \DIFaddbegin \DIFadd{It was discovered that all 18 altered photographs were not recognized as being morphed by qualified face comparison experts.}\DIFaddend Makrushin et al. \cite{makrushin2020simulation} conducted \DIFdelbegin \DIFdel{the }\DIFdelend \DIFaddbegin \DIFadd{a }\DIFaddend web-based simulation of a border control scenario. The \DIFdelbegin \DIFdel{observers }\DIFdelend \DIFaddbegin \DIFadd{observers }\DIFaddend were provided with an introduction to face morphing, morphing attacks\DIFaddbegin \DIFadd{, }\DIFaddend and hints to identify the morphed images. The observers were \DIFdelbegin \DIFdel{asked to conduct }\DIFdelend \DIFaddbegin \DIFadd{required to complete }\DIFaddend 15 trials, \DIFdelbegin \DIFdel{where each trial consisted of }\DIFdelend \DIFaddbegin \DIFadd{each of which contained 10 morphing face photos and }\DIFaddend 5 \DIFdelbegin \DIFdel{bona fide and 10 morphed face images. First, the }\DIFdelend \DIFaddbegin \DIFadd{real face photographs. The }\DIFaddend observer had to \DIFdelbegin \DIFdel{compare the passport facial image against the given bona fide or morphed image . 
	}\DIFdelend \DIFaddbegin \DIFadd{first contrast the face image from the passport with the genuine or altered image that had been provided. }\DIFaddend The second experiment was for face comparison, with 25 trials (10 mated comparisons, 5 \DIFdelbegin \DIFdel{non-mated comparisons}\DIFdelend \DIFaddbegin \DIFadd{nonmated comparisons, }\DIFaddend and 10 morphing comparison trials). \DIFdelbegin \DIFdel{In each of these trails, two images were presented to observers with one image }\DIFdelend \DIFaddbegin \DIFadd{Two images were shown to observers one }\DIFaddend on a frame \DIFdelbegin \DIFdel{of a sample }\DIFdelend \DIFaddbegin \DIFadd{from a model }\DIFaddend passport to be \DIFdelbegin \DIFdel{compared against a small video clip where the traveler rotated his /}\DIFdelend \DIFaddbegin \DIFadd{contrasted with a brief video clip in which the traveler rotated his or }\DIFaddend her head. Finally, the observer had to decide on the passport face image, to decide if the \DIFdelbegin \DIFdel{traveler'}\DIFdelend \DIFaddbegin \DIFadd{traveler’}\DIFaddend s image was morphed or not looking at the video. \DIFaddbegin \DIFadd{In both trials, }\DIFaddend 189 \DIFdelbegin \DIFdel{observers participated in both experiments, of which }\DIFdelend \DIFaddbegin \DIFadd{observers took part, and }\DIFaddend 21 \DIFdelbegin \DIFdel{observers had no previous experience with face morphing}\DIFdelend \DIFaddbegin \DIFadd{of them had never seen a face transform before}\DIFaddend . In the first experiment (no reference), unskilled observers\DIFdelbegin \DIFdel{' }\DIFdelend \DIFaddbegin \DIFadd{’ }\DIFaddend true positive rate (TPR) was reported to be 65.65\%\DIFaddbegin \DIFadd{, }\DIFaddend and a TPR of 63.33\% for skilled observers. \DIFdelbegin \DIFdel{In }\DIFdelend \DIFaddbegin \DIFadd{The TPR of competent and untrained observers in }\DIFaddend the second experiment (\DIFaddbegin \DIFadd{a }\DIFaddend face comparison) \DIFdelbegin \DIFdel{, skilled observers' TPR was }\DIFdelend \DIFaddbegin \DIFadd{were }\DIFaddend reported as 93.45\% and \DIFdelbegin \DIFdel{unskilled observers as }\DIFdelend 95.25\%\DIFdelbegin \DIFdel{. }\DIFdelend \DIFaddbegin \DIFadd{, respectively }\DIFaddend Makrushin et al. \cite{makrushin2020simulation} claimed that training the unskilled observers would increase the detection ability to reach the performance of skilled observers. \DIFdelbegin \DIFdel{\Rev{Makrushin et al. \cite{makrushin2020simulation} also examined the accuracy a face recognition software from the Dermalog Face Recognition \cite{dermalog202identification, king2009dlib} where they noted the false acceptance rate (FAR) of human observers was higher (15.45\%) than the said software (0\%).}
	}\DIFdelend \DIFaddbegin \DIFadd{The accuracy of a face-recognition program from Dermalog Face Recognition \mbox{
			\cite{dermalog202identification, king2009dlib} }\hspace{0pt}
		was also investigated by Makrushin et al. \mbox{
			\cite{makrushin2020simulation} }\hspace{0pt}
		where they noted the false acceptance rate (FAR) of human observers was greater (15.45\%) than the algorithms’s (0\%).
	}\DIFaddend 
	
	\DIFaddbegin \DIFadd{A pair of 54 digital images was used in }\DIFaddend Nightingale et al. \cite{nightingale2021perceptual} \DIFdelbegin \DIFdel{recently examined the ability of human observers to detect morphed }\DIFdelend \DIFaddbegin \DIFadd{investigation of the human observer’s capacity to recognize altered }\DIFaddend images under S-MAD and D-MAD \DIFdelbegin \DIFdel{setting for a pair of 54 digital images. \Rev{However, the study employed one hundred workers on Amazon's Mechanical Turk (AMT) in a crowd-based approach who were neither expert examiners nor observers familiar with face recognition}. While the observers recorded an accuracy of }\DIFdelend \DIFaddbegin \DIFadd{conditions. However, the study employed one hundred workers on Amazon’s Mechanical Turk (AMT) in a crowd-based approach who were neither expert examiners nor observers familiar with face recognition. While the observer’s accuracy in distinguishing between genuine and fake images was }\DIFaddend 80.8\%\DIFdelbegin \DIFdel{ in identifying the bona fide v/s bona fide images , morphing detection accuracy was reported as }\DIFdelend \DIFaddbegin \DIFadd{\%, their success rate in detecting morphing was only }\DIFaddend 58.3\%. However, the observers who received some training obtained a similar accuracy of 59.2\%, but the work noted that participants made different mistakes as compared to \DIFaddbegin \DIFadd{the }\DIFaddend original experiment. \DIFaddbegin \DIFadd{Further testing of the face-recognition algorithm by }\DIFaddend Nightingale et al. \cite{nightingale2021perceptual} \DIFdelbegin \DIFdel{further evaluated the face recognition algorithm using deep VGG features resulting in a very low accuracy with }\DIFdelend \DIFaddbegin \DIFadd{revealed very poor performance in }\DIFaddend an Area Under the Curve (AUC) of 0.38.
	
	Zhang et al. \cite{zhang2021mipgan} \DIFdelbegin \DIFdel{studied the ability }\DIFdelend \DIFaddbegin \DIFadd{investigated the capacity }\DIFaddend of human observers \DIFdelbegin \DIFdel{employing }\DIFdelend \DIFaddbegin \DIFadd{using 42 inexperienced viewers and }\DIFaddend 14 experienced observers \DIFdelbegin \DIFdel{(highly }\DIFdelend \DIFaddbegin \DIFadd{using 15 genuine and 75 morphed photos. The experienced observers were extremely }\DIFaddend familiar with morphing \DIFdelbegin \DIFdel{, including }\DIFdelend \DIFaddbegin \DIFadd{and included }\DIFaddend ID experts in border control\DIFdelbegin \DIFdel{) and 42 inexperienced observers using 15 bona fide and 75 morphed images}\DIFdelend . Unlike the previous works, Zhang et al. \cite{zhang2021mipgan} studied five different types of morphing algorithms that included landmark-based morphs and Generative Adversarial Network (GAN) based morphs. The \DIFdelbegin \DIFdel{experienced group obtained a detection accuracy of }\DIFdelend \DIFaddbegin \DIFadd{detection accuracy for the skilled group was }\DIFaddend 97.14\%. \DIFdelbegin \DIFdel{In contrast, the non-experienced }\DIFdelend \DIFaddbegin \DIFadd{While detecting landmark-based morphs was difficult compared to detecting GAN-based morphing, the inexperienced }\DIFaddend obtained an accuracy of 79.21\%\DIFdelbegin \DIFdel{, and it was noted that detecting landmarks-based morphs was challenging compared to detecting GAN based morphing}\DIFdelend . \DIFaddbegin \DIFadd{Five experienced observers and ten novice observers were used in }\DIFaddend Zhang et al.\DIFdelbegin \DIFdel{\mbox{
			\cite{zhang2021mipgan} }\hspace{0pt}
		also studied }\DIFdelend \DIFaddbegin \DIFadd{'s \mbox{
			\cite{zhang2021mipgan} }\hspace{0pt}
		study of }\DIFaddend D-MAD detection \DIFdelbegin \DIFdel{ability with five experienced observers and ten inexperienced observers, and the }\DIFdelend \DIFaddbegin \DIFadd{abilities. The }\DIFaddend group with relevant \DIFdelbegin \DIFdel{experiences achieved an overall }\DIFdelend \DIFaddbegin \DIFadd{expertise had an overall accuracy of }\DIFaddend 86\%\DIFdelbegin \DIFdel{accuracy}\DIFdelend .
	
	A detailed analysis of all the existing state-of-the-art for human observer \DIFdelbegin \DIFdel{analysis }\DIFdelend \DIFaddbegin \DIFadd{analyses }\DIFaddend for MAD is presented in Table~\ref{tab:sota-human-mad}.  

	\subsection{Algorithms for morphing Attack Detection}
	\subsubsection{Differential morphing attack detection (D-MAD) }
	\DIFaddend In the lines of human observer-based morphing attack detection, \DIFdelbegin \DIFdel{a number of }\DIFdelend \DIFaddbegin \DIFadd{several }\DIFaddend approaches have been devised for automated morphing attack detection. \DIFdelbegin \DIFdel{A trusted }\DIFdelend \DIFaddbegin \DIFadd{The second suspicious image is examined to determine if it has been morphed or not using a trustworthy }\DIFaddend live capture image \DIFdelbegin \DIFdel{is provided }\DIFdelend from an ABC gate or \DIFdelbegin \DIFdel{otherwise trusted capture device, and the second suspected image is analysed to decide if it is morphed or not}\DIFdelend \DIFaddbegin \DIFadd{other reliable capture device}\DIFaddend . The first approach of D-MAD was based on inverting the morphing process in a reverse-engineered manner, which was termed \DIFdelbegin \DIFdel{as }\DIFdelend \DIFaddbegin \DIFadd{Demorphing }\DIFaddend \cite{Ferrara2018demorphing}. \DIFdelbegin \DIFdel{Similarly, several works have been reported where the difference of feature vectors from the bona fide image and the morphed image is used to determine if the suspected image is }\DIFdelend \DIFaddbegin \DIFadd{Similar to this, several studies have been published in which the difference between the feature vectors of the genuine and }\DIFaddend morphed \DIFaddbegin \DIFadd{images are utilized to assess whether an image is genuine or morphed }\DIFaddend \cite{scherhag2020deep,mohan2019robust}. \DIFdelbegin \DIFdel{The deep features from two different networks are employed to determine the difference in features in \mbox{
\cite{scherhag2020deep}}\hspace{0pt}
, and features from the }\DIFdelend \DIFaddbegin \DIFadd{Morphing attacks were also detected using features from the }\DIFaddend 3D shape and the diffuse reflectance component estimated directly from the image \DIFdelbegin \DIFdel{was employed to detect a morphing attack in \mbox{
\cite{mohan2019robust}}\hspace{0pt}
. Another set of works explored the shift }\DIFdelend in \DIFaddbegin \DIFadd{\mbox{
\cite{mohan2019robust}}\hspace{0pt}
, and the deep features from two different networks were used to determine the difference in features \mbox{
\cite{scherhag2020deep}}\hspace{0pt}
. Another group of works investigated morphing by examining the change in }\DIFaddend landmarks of \DIFdelbegin \DIFdel{bona fide }\DIFdelend \DIFaddbegin \DIFadd{genuine }\DIFaddend and suspected morph images in the \DIFdelbegin \DIFdel{face region to determine the morphing attack \mbox{
\cite{damer2018detecting,scherhag2019detection}}\hspace{0pt}
.
	}\DIFdelend \DIFaddbegin \DIFadd{region of the face \mbox{
\cite{damer2018detecting,scherhag2019detection}}\hspace{0pt}
.
	}\DIFaddend 


	\subsubsection{Single image morphing attack detection (S-MAD) }
	S-MAD algorithms largely rely on learning a classifier to distinguish the bona fide image from a morphed image. \DIFdelbegin \DIFdel{Given a suspected morph image, $I_s$, the texture information }\DIFdelend \DIFaddbegin \DIFadd{The texture data }\DIFaddend is extracted from the normalized and aligned face \DIFaddbegin \DIFadd{of a suspected morph image}\DIFaddend . The texture features such as Binarized Statistical Independent Features (BSIF) \DIFdelbegin \DIFdel{(BSIF) }\DIFdelend and Local Binary Patterns (LBP) are used to classify the images using a \DIFdelbegin \DIFdel{pre-trained SVM}\DIFdelend \DIFaddbegin \DIFadd{pretrained Support Vector Machine (SVM) }\DIFaddend classifier \cite{raghavendra2018detecting,raghavendra2017face,scherhag2017vulnerability, spreeuwers2018towards,scherhag2019detection} in the earlier works. While extending the works for MAD, another approach was proposed to exploit the \DIFdelbegin \DIFdel{colour }\DIFdelend \DIFaddbegin \DIFadd{color }\DIFaddend spaces and the scale spaces jointly \cite{raghavendra2018detecting, ramachandra2019towards}. \DIFdelbegin \DIFdel{With the intent to address and also detect the post-processed morphed images, pre-trained }\DIFdelend \DIFaddbegin \DIFadd{Pretrained }\DIFaddend deep networks for \DIFdelbegin \DIFdel{extraction of texture features were employed }\DIFdelend \DIFaddbegin \DIFadd{texture feature extraction were used }\DIFaddend to detect the morphing attacks not only in the digital domain but also in the \DIFdelbegin \DIFdel{re-digitized }\DIFdelend \DIFaddbegin \DIFadd{redigitized }\DIFaddend domain (after a print-scan transformation) \cite{raghavendra2017transferable} \DIFaddbegin \DIFadd{to address and detect the postprocessed morphed images}\DIFaddend . Notably, the earlier works have employed two deep neural networks\DIFdelbegin \DIFdel{, }\DIFdelend \DIFaddbegin \DIFadd{; }\DIFaddend including VGG19 \cite{simonyan2014very} and AlexNet \cite{krizhevsky2012imagenet}, where they perform \DIFdelbegin \DIFdel{feature level }\DIFdelend \DIFaddbegin \DIFadd{a feature-level }\DIFaddend fusion of the first fully connected layers from both the networks  \cite{raghavendra2017transferable}. \DIFdelbegin \DIFdel{In a continued effort, other }\DIFdelend \DIFaddbegin \DIFadd{Other }\DIFaddend deep networks have been \DIFdelbegin \DIFdel{investigated for detecting morph attacks \mbox{
\cite{ferrara2019face}}\hspace{0pt}
. Another }\DIFdelend \DIFaddbegin \DIFadd{researched for morph attack detection in an ongoing effort  \mbox{
\cite{ferrara2019face}}\hspace{0pt}
. Another }\DIFaddend approach to detecting morphing attacks was proposed by extracting the features from the “Photo Response \DIFdelbegin \DIFdel{Non-Uniformity“,}\DIFdelend \DIFaddbegin \DIFadd{Non Uniformity (PRNU),” }\DIFaddend where the characteristics of the image sensor were employed to determine if the image was morphed or not \cite{debiasi2018prnu}. \DIFdelbegin \DIFdel{Motivated by the effectiveness of the noise modelling, }
\DIFdel{better performing }
\DIFdelend \DIFaddbegin \DIFadd{Better performing }\DIFaddend algorithms have been reported\DIFaddbegin \DIFadd{, including dedicated context aggregation networks to automatically model the noise \mbox{
\cite{venkatesh2019morphed} }\hspace{0pt}
}\DIFaddend using color spaces to seek \DIFdelbegin \DIFdel{for }\DIFdelend residuals of the morphing process.\DIFdelbegin \DIFdel{\mbox{
\cite{venkatesh2019morphed} }\hspace{0pt}
including dedicated context aggregation networks to automatically model the noise \mbox{
\cite{venkatesh2020detecting}}\hspace{0pt}
	}\DIFdelend 


	\subsection{Limitations of the existing works}
	\label{ssec:limitations-of-sota}
	With a detailed review of the existing state-of-the-art in previous section, we note a set of limitations in the existing works in this section.

	\begin{itemize}
		\item The \DIFdelbegin \DIFdel{previously reported works focused mostly on }\DIFdelend \DIFaddbegin \DIFadd{majority of the previously mentioned works were primarily concerned with }\DIFaddend D-MAD \DIFdelbegin \DIFdel{settings }\DIFdelend or S-MAD settings. \DIFdelbegin \DIFdel{Thus}\DIFdelend \DIFaddbegin \DIFadd{As a result}\DIFaddend , they do not \DIFdelbegin \DIFdel{realistically benchmark the }\DIFdelend \DIFaddbegin \DIFadd{accurately benchmark the ability of the }\DIFaddend same human observer \DIFdelbegin \DIFdel{'s capability to assess the morphed }\DIFdelend \DIFaddbegin \DIFadd{to evaluate the altered }\DIFaddend images in D-MAD and S-MAD \DIFdelbegin \DIFdel{settings}\DIFdelend \DIFaddbegin \DIFadd{situations}\DIFaddend .
		\item \DIFdelbegin \DIFdel{Further, the morphed images used in those experiments are not carefully post-processed to resemble the realistic passport images in ID cardsleading to a somewhat biased observation}\DIFdelend \DIFaddbegin \DIFadd{Furthermore, the morphed images utilized in those studies were not carefully postprocessed to mimic the authentic passport images on ID cards, which resulted in a somewhat skewed conclusion}\DIFaddend . For instance, Kramer et al. \cite{kramer2019face} used a \DIFdelbegin \DIFdel{high quality }\DIFdelend \DIFaddbegin \DIFadd{high-quality }\DIFaddend morphed database, but the images themselves do not have a realistic background for face images. \DIFdelbegin \DIFdel{Only the cropped face is presented without any presence of neck or background for the human face }\DIFdelend \DIFaddbegin \DIFadd{Without a neck or other human face background, only the cropped face is shown}\DIFaddend . Another set of works  \cite{robertson2017fraudulent,Ferrara2016} directly use the morphed image, without any modification/\DIFdelbegin \DIFdel{post-processing }\DIFdelend \DIFaddbegin \DIFadd{postprocessing }\DIFaddend despite many visible \DIFdelbegin \DIFdel{artefacts }\DIFdelend \DIFaddbegin \DIFadd{artifacts }\DIFaddend leading to biased observations and thereby conclusions.
		\item \DIFdelbegin \DIFdel{Most of the previous works related to morphing detection with human observers employed a smaller set of images}\DIFdelend \DIFaddbegin \DIFadd{The majority of earlier studies on morphing detection using human observers used a smaller collection of images}\DIFaddend . However, the small size of the dataset does not lead to a statistically conclusive result.  The reported works employ a small set of expert observers to validate the initial observations made by the previous studies. \DIFdelbegin \DIFdel{Brysbaert \mbox{
\cite{brysbaert2019many} }\hspace{0pt}
have explained the importance }\DIFdelend \DIFaddbegin \DIFadd{To obtain precise results in subjective studies, Brysbaert \mbox{
\cite{brysbaert2019many} }\hspace{0pt}
has discussed the significance }\DIFaddend of the sample size and number of participants\DIFdelbegin \DIFdel{to get clear results in subjective experiments}\DIFdelend . The large sample size experiment gives a more precise statistical analysis.
	\end{itemize}

	 \DIFdelbegin \DIFdel{In order }\DIFdelend \DIFaddbegin \DIFadd{We perform empirical research depending on the information acquired from human observers }\DIFaddend to address the \DIFdelbegin \DIFdel{limitations }\DIFdelend \DIFaddbegin \DIFadd{constraints }\DIFaddend of the current \DIFdelbegin \DIFdel{state-of-art and answer the research questions noted above, we conduct \Rev{empirical research relying on the data gathered from human observers.}
	}\DIFdelend \DIFaddbegin \DIFadd{state-of-the-art and respond to the research questions listed previously.
	}\DIFaddend \section{\Rev{Methodology}}
	\label{sec:methodology}
	\begin{figure}[H]
		\centering
		\subfloat{\includegraphics[width = 1in]{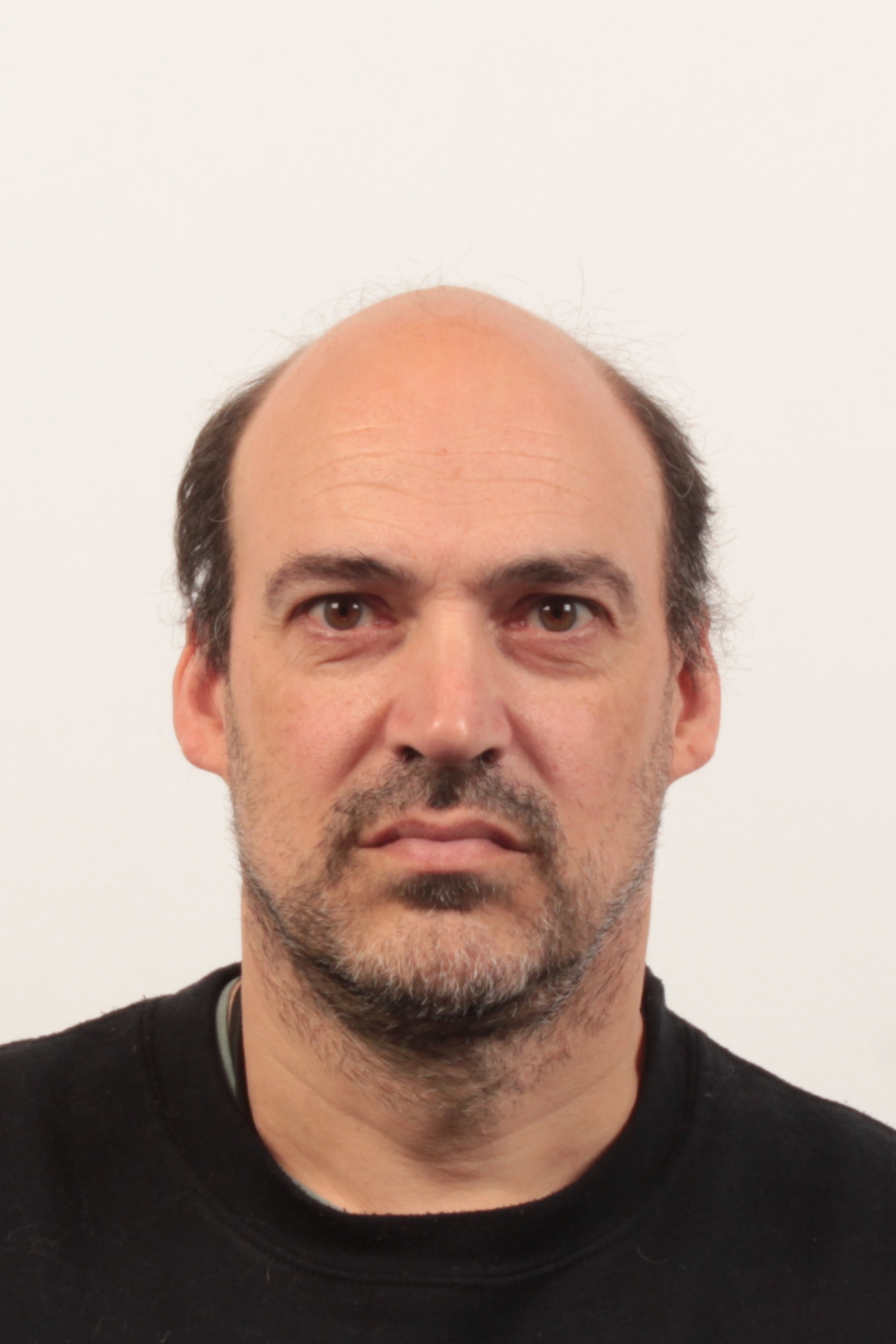}}
		\subfloat{\includegraphics[width = 1in]{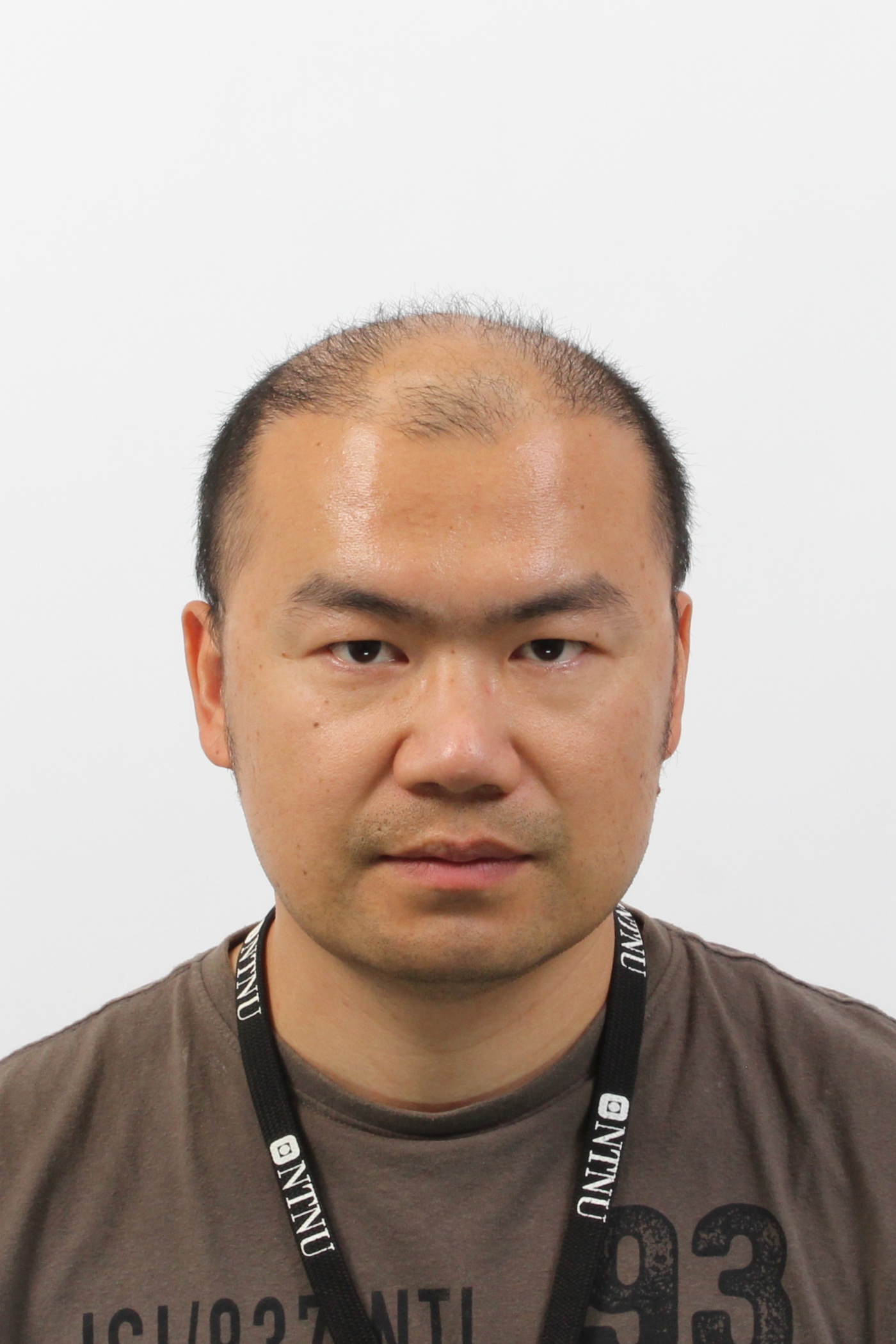}}
		\subfloat{\includegraphics[width = 1in]{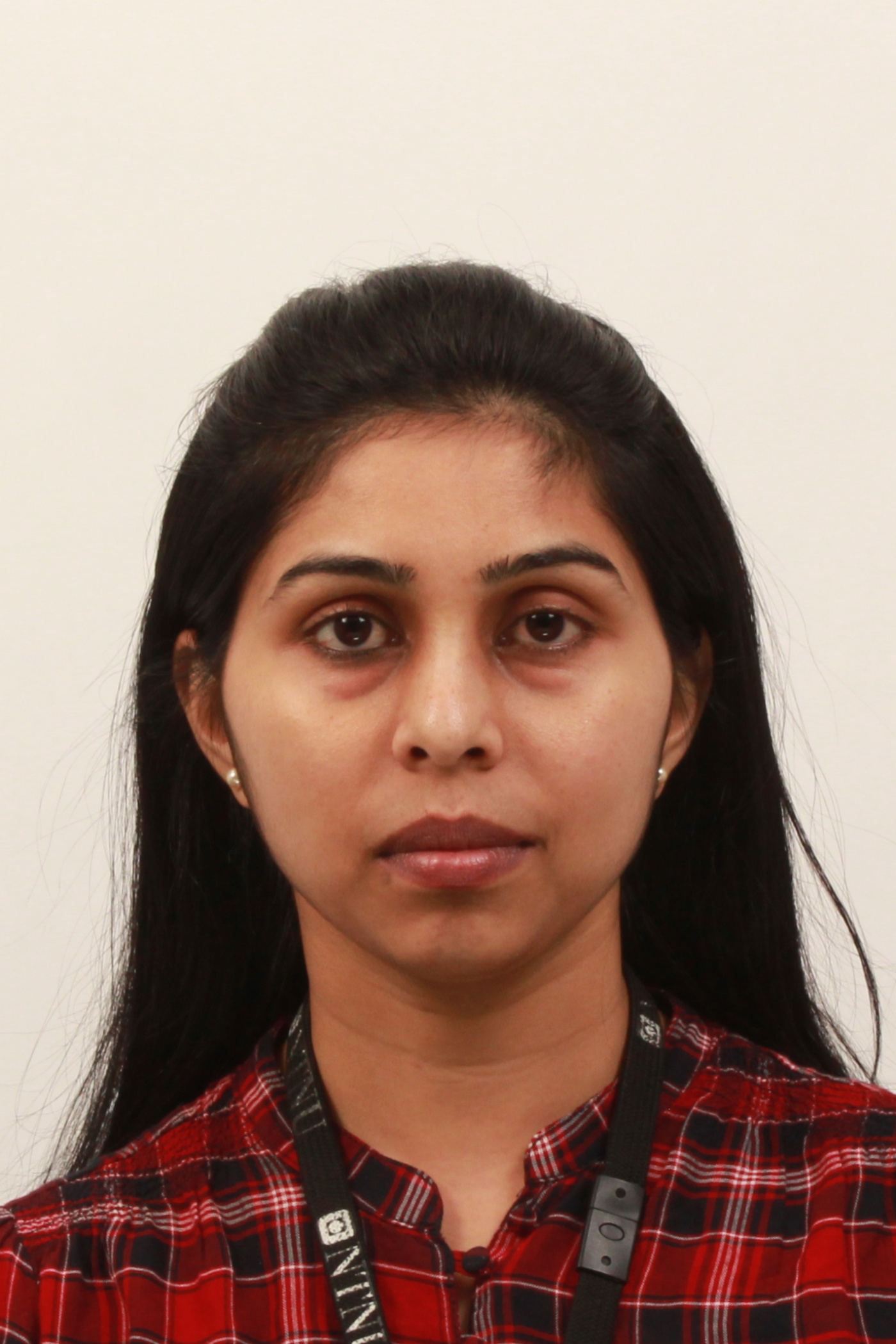}}\\
		\caption{Bona fide subset in HOMID-D.}
		\label{tab:d-mad-bonafide-images}
	\end{figure}

	\DIFdelbegin \DIFdel{\Rev{Our methodology consists of three core components such as creation of the new datasets; development of the evaluation platform to gather the data on human observers; and the data collection associated with the human observer evaluation and analysis as presented in subsequent sections. We first present the analysis at a high level for detecting D-MAD and S-MAD, following which various sub-analysis is presented measuring the performance w.r.t duration of training, field of expertise and detection ability in print-scan setting. In addition, we present detailed analysis of detection accuracy from two different algorithms for S-MAD and D-MAD respectively.}
	}\DIFdelend \DIFaddbegin \DIFadd{\Rev{Our methodology consists of three main parts, which are as follows: the development of new datasets; the building of an evaluation platform to collect data on human observers; and the data collecting related to the evaluation and analysis of the human observers. We first present the analysis at a high level for detecting D-MAD and S-MAD, following which various subanalysis is presented measuring the performance w.r.t duration of the training, field of expertise, and detection ability in the print-scan setting. Additionally, we provide a thorough examination of the detection accuracy for S-MAD and D-MAD using two alternative automated techniques.}}\DIFaddend 

\subsection{\Rev{Datasets:} Human Observer Analysis Morphing Image Database (HOMID)}
	\label{sec:database}
	\DIFdelbegin \DIFdel{Noting the non-availability of the face images }\DIFdelend \DIFaddbegin \DIFadd{We first create a new database called the HOMID to address the lack of face images }\DIFaddend for human observer analysis of morphing \DIFdelbegin \DIFdel{attack detection, we first generate a new database referred to as Human Observer Analysis Morphing Image Database (HOMID)}\DIFdelend \DIFaddbegin \DIFadd{attack detection}\DIFaddend . As a noted limitation in the earlier works, the face images employed for the human observer analysis do not have images respecting the ICAO standards for Machine Readable Travel Documents (MRTD) \cite{ICAO-9303-p9-2015}. \DIFdelbegin \DIFdel{We, therefore, create }\DIFdelend \DIFaddbegin \DIFadd{As a result, we develop }\DIFaddend a new database that \DIFdelbegin \DIFdel{is both realistic and compliant with the ICAO standards. The database has two main subsets , each with both bonafide }\DIFdelend \DIFaddbegin \DIFadd{complies with ICAO criteria while still being realistic. There are two primary subsets of the database, each including bona fide }\DIFaddend and morph subsets. The first subset, HOMID-D, corresponds to D-MAD settings where the bona fide images are captured from a regular \DIFdelbegin \DIFdel{photo-studio }\DIFdelend \DIFaddbegin \DIFadd{photo studio }\DIFaddend setting, and bona fide probe images are captured from a real ABC gate. The \DIFdelbegin \DIFdel{morphed images in this subset are created using the procedure explained subsequently}\DIFdelend \DIFaddbegin \DIFadd{process described later is used to create the morphed photos in this subgroup}\DIFaddend . The second subset corresponds to S-MAD settings where the bona fide images are captured in a photo studio setting, and morphed images are created, as explained further below. \DIFdelbegin \DIFdel{Both subsets of data have digitally post-processed and printed-scanned images to study all the aspects }\DIFdelend \DIFaddbegin \DIFadd{To explore all the facets }\DIFaddend of MAD by human observers\DIFaddbegin \DIFadd{, both subsets of data have printed-scanned photos that have undergone digital postprocessing}\DIFaddend .

	\begin{table}[H]
		\centering
		\resizebox{0.85\linewidth}{!}{%
			\begin{tabular}{lcccc}
				\hline
				Ethnicity & Caucasian & South Asian & Middle East & Chinese  \\ 
				\hline
				\hline
				Female    & 2         & 2            & 3           & 0       \\ 
				Male      & 26        & 10           & 2           & 3       \\ 
				Total     & 28        & 12           & 5           & 3       \\ 
				\hline
				\hline
			\end{tabular}%
		}
		\newline
		\caption{Demographic distribution of HOMID-D subset for bona fide images}
		\label{tab:d-mad-bonafide-distribution}
	\end{table}
\begin{figure}[H]
	\centering
	\subfloat{\includegraphics[width = 1in]{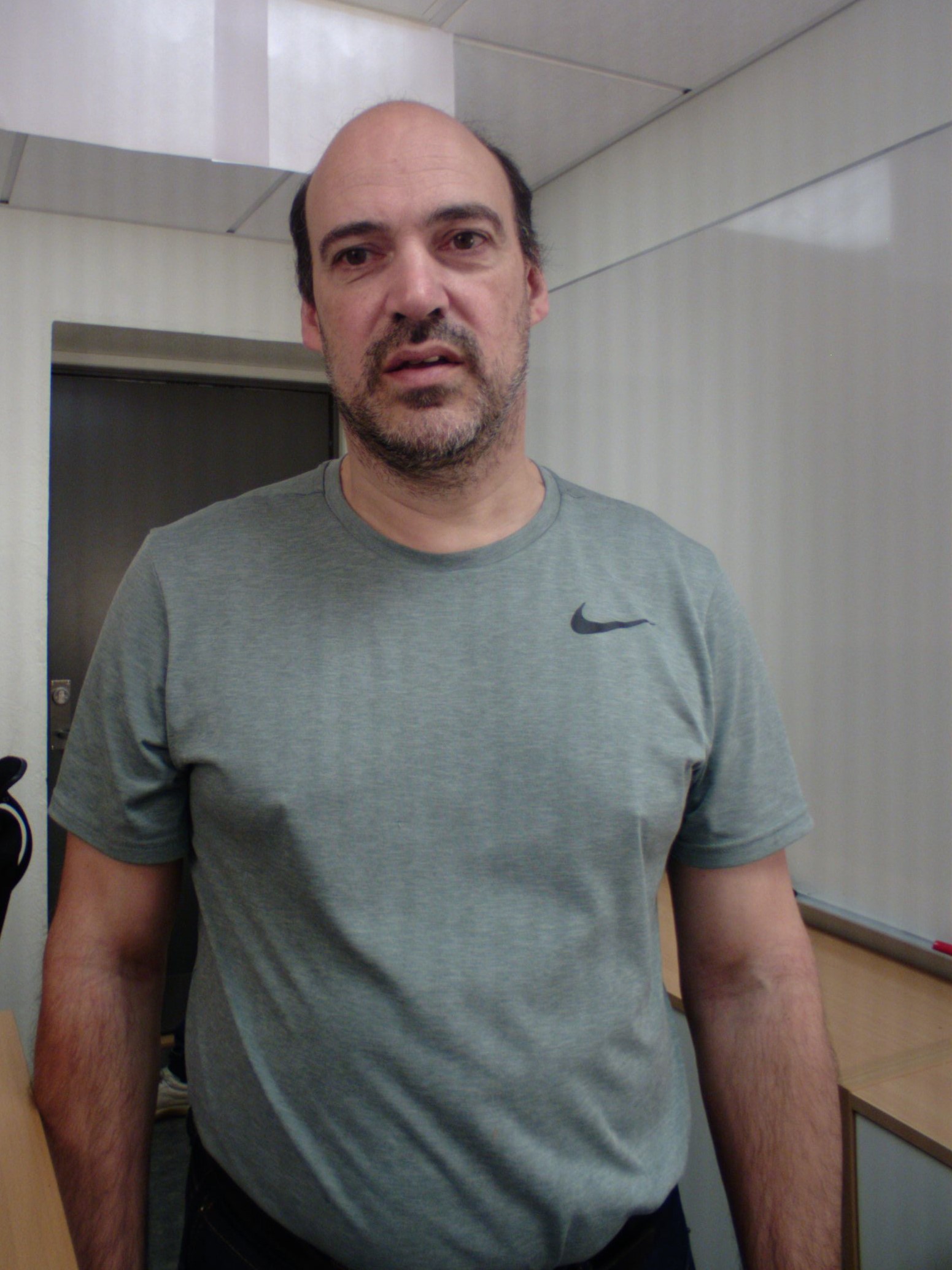}}
	\subfloat{\includegraphics[width = 1in]{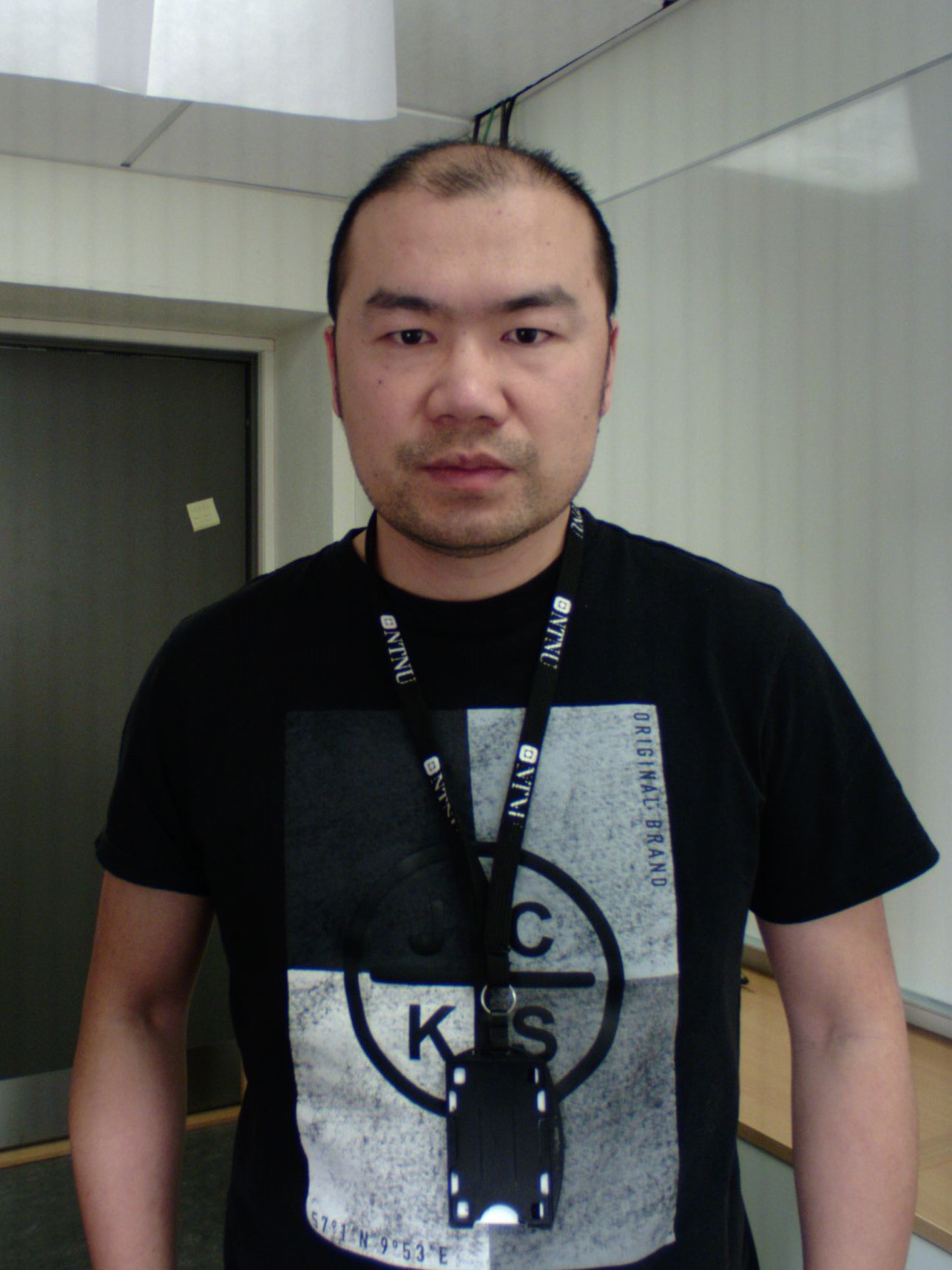}}
	\subfloat{\includegraphics[width = 1in]{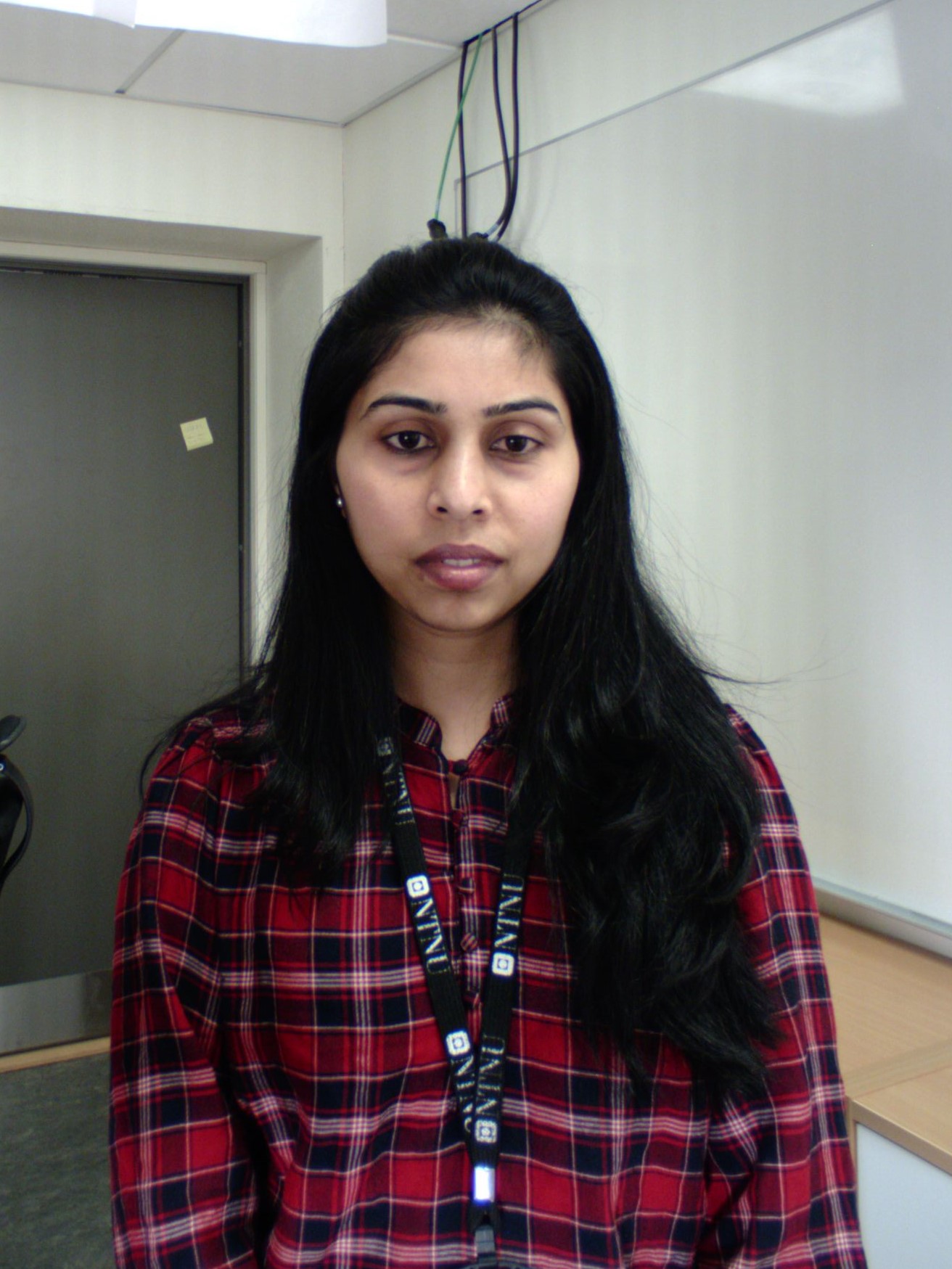}}\\
	\caption{Example images captured under trusted capture settings from ABC gate images corresponding to subjects in Figure~\ref{tab:d-mad-bonafide-images}. ABC gate images were collected on a different day to reflect real life scenarios.}
	\label{tab:d-mad-abc-images}
\end{figure}

	\subsubsection{HOMID Differential MAD Subset}
	\DIFdelbegin \DIFdel{In order to study the human observer ability to detect morphing under }\DIFdelend \DIFaddbegin \DIFadd{We produce the first subset of data that contains genuine and morphed images to explore how well human observers can recognize morphing in }\DIFaddend a differential (i.e., reference-based) \DIFdelbegin \DIFdel{setting, we create the first subset of data that has bona fide and morphed images}\DIFdelend \DIFaddbegin \DIFadd{environment}\DIFaddend . The bona fide images are collected from \DIFdelbegin \DIFdel{$48$ }\DIFdelend \DIFaddbegin \DIFadd{48 }\DIFaddend unique subjects with different age groups, gender\DIFaddbegin \DIFadd{, }\DIFaddend and ethnicity. The demographic distribution of the dataset is provided in  Table~\ref{tab:d-mad-bonafide-distribution}.

	\textbf{HOMID-D Bona fide Subset}: \DIFdelbegin \DIFdel{The }\DIFdelend \DIFaddbegin \DIFadd{To replicate the ICAO passport photo, the }\DIFaddend face images were \DIFdelbegin \DIFdel{captured in a photo-studio setting with good illumination}\DIFdelend \DIFaddbegin \DIFadd{taken in a photo studio with adequate lighting}\DIFaddend , no shadows, no poses, \DIFdelbegin \DIFdel{eyes fully open without glasses, closed mouth and neutral expression in }\DIFdelend \DIFaddbegin \DIFadd{and neutral expressions on }\DIFaddend a white background\DIFdelbegin \DIFdel{to mimic the ICAO standard passport photo}\DIFdelend . The face image of each data subject was captured multiple times, from which one image was chosen for creating morphs, and other images were used for studying the vulnerability of FRS. Each \DIFdelbegin \DIFdel{image was captured in }\DIFdelend \DIFaddbegin \DIFadd{photo was taken at a }\DIFaddend high resolution and \DIFdelbegin \DIFdel{later cropped to an image of $413 \times 513$  pixels }\DIFdelend \DIFaddbegin \DIFadd{afterward reduced to a 413 × 513 pixels size}\DIFaddend . Further, the cropped face images in the bona fide set were carefully cropped to respect the minimum Inter-Eye Distance \DIFdelbegin \DIFdel{(IED) }\DIFdelend according to passport image standards in various countries. \DIFdelbegin \DIFdel{The bona fide subset consists of $480$ images in total . A sample illustration of the bona fide images }\DIFdelend \DIFaddbegin \DIFadd{There are 480 total photos in the genuine subset. Figure~\ref{tab:d-mad-bonafide-images} offers an example representation of the genuine photos }\DIFaddend in the HOMID-D subset\DIFdelbegin \DIFdel{is provided in Figure~\ref{tab:d-mad-bonafide-images}}\DIFdelend .

	\textbf{HOMID-D Trusted Capture Subset}: In the D-MAD setting, a trusted live capture image is compared against a suspect image. We \DIFdelbegin \DIFdel{create a set of probe images acquired }\DIFdelend \DIFaddbegin \DIFadd{use an ABC gate to build a series of probe pictures that correspond to }\DIFaddend trusted capture \DIFdelbegin \DIFdel{settings using an Automated Border Control (ABC) gate}\DIFdelend \DIFaddbegin \DIFadd{conditions}\DIFaddend. Images from each of the 48 subjects were captured using the ABC gate resulting in 10 different attempts. \DIFdelbegin \DIFdel{Out }\DIFdelend \DIFaddbegin \DIFadd{, five }\DIFaddend of the 10 \DIFdelbegin \DIFdel{images, 5 images were used to assess }\DIFdelend \DIFaddbegin \DIFadd{photos were utilized to evaluate }\DIFaddend the vulnerability of \DIFdelbegin \DIFdel{provided }\DIFdelend \DIFaddbegin \DIFadd{supplied }\DIFaddend morphs. Specifically, we have employed a state-of-art Commercial-Off-The-Shelf (COTS) system to verify the quality of the morphed images against the captured ABC gate images. \DIFdelbegin \DIFdel{A high comparison score obtained confirms the high }\DIFdelend \DIFaddbegin \DIFadd{The good }\DIFaddend quality of the \DIFdelbegin \DIFdel{created morph images and post-processing quality }\DIFdelend \DIFaddbegin \DIFadd{produced morph pictures and postprocessing quality is confirmed by the high comparative score }\DIFaddend \footnote{A high score here refers to a score crossing the predefined threshold by COTS SDK for a given FAR. For instance, Neurotech defines the threshold of 36 at FAR=0.1\%, and any score crossing the value is considered a high score.}. The rest of the 5 images from each subject were further scrutinized for appearance and quality \DIFdelbegin \DIFdel{score }\DIFdelend \DIFaddbegin \DIFadd{scores by an expert researcher}\DIFaddend based on which the best image was chosen for subjective experiments. Figure ~\ref{tab:d-mad-abc-images} \DIFdelbegin \DIFdel{represents few images }\DIFdelend \DIFaddbegin \DIFadd{shows a selection of photographs }\DIFaddend from our database \DIFdelbegin \DIFdel{corresponding to bona fide images represented }\DIFdelend \DIFaddbegin \DIFadd{that correlate to the real images seen }\DIFaddend in Figure~\ref{tab:d-mad-bonafide-images}.

\begin{figure}[h]
		\centering 
		\includegraphics[width=0.8\textwidth]{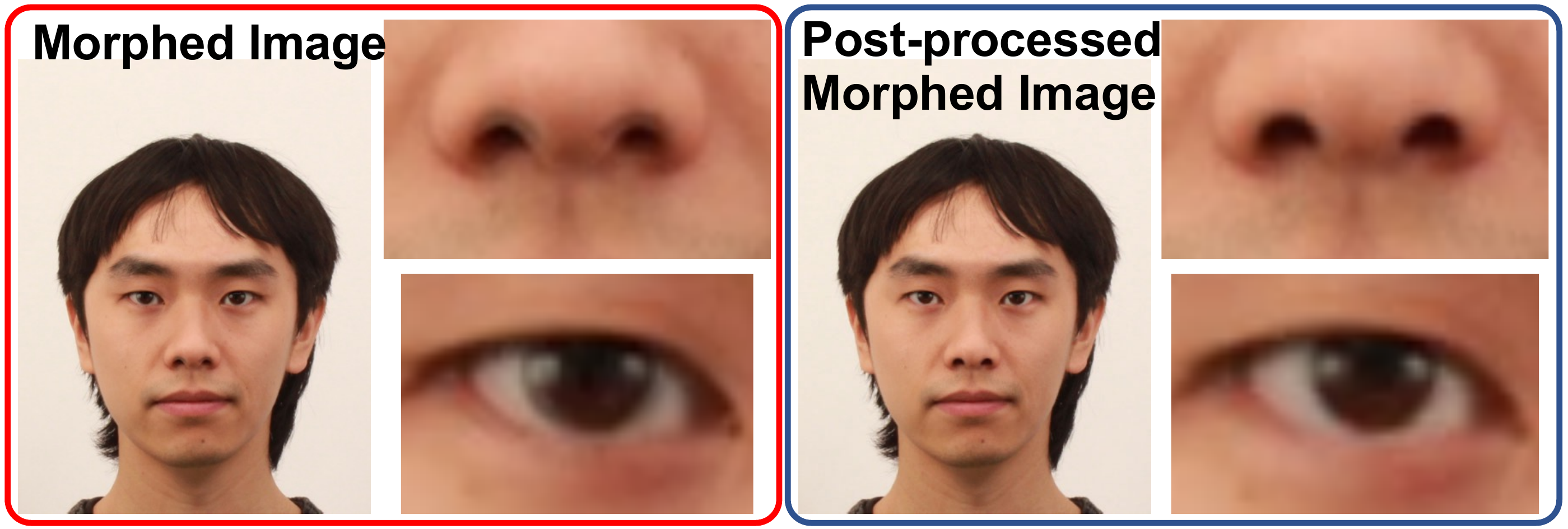}%
		\caption{Morphed image which has visible artifacts near the iris area and the nose areas and the corresponding post processed image with artifacts eliminated.}
		\label{fig:post-processing-morph}
	\end{figure}

	\textbf{HOMID-D Morph Subset}:	\DIFdelbegin \DIFdel{With }\DIFdelend \DIFaddbegin \DIFadd{We combine two similar subjects to generate 400 morphed images using }\DIFaddend the newly created \DIFdelbegin \DIFdel{images of $48$ bona fide subjects , we create a $400$ morphed image by combining two resembling subjects }\DIFdelend \DIFaddbegin \DIFadd{photographs of 48 real subjects }\DIFaddend (in terms of age, gender, ethnicity, and skin color). Based on the earlier works indicating high attack strength of morphed images with a morphing factor of \DIFdelbegin \DIFdel{$0.5$ }\DIFdelend \DIFaddbegin \DIFadd{0.5 }\DIFaddend \cite{ferrara2014magic, ferrara2016effects, raja2020morphing,  venkatesh2021morphsurvey}(i.e., 50\% contribution from each subject), we create the morphed images using a morphing factor of $0.5$. \DIFdelbegin \DIFdel{\Rev{The selection of subjects for morphing was done by a human expert to avoid unrealistic morphing combinations. Further, to create highly realistic morphed images to challenge the human observers, each of the morphed images was post-processed to eliminate any artefacts in the image  stemming from the morphing process, such as double contours in the iris region, incorrect lip region, etc. Each post-processed image was further scrutinized by two experts ensuring the high quality of created morphed images.} }\DIFdelend \DIFaddbegin \DIFadd{To prevent the creation of implausible morphing combinations the subjects for morphing were chosen by a human expert. Further, to create highly realistic morphed images to challenge the human observers, each of the morphed images was postprocessed to eliminate any artifacts in the image signal stemming from the morphing process, such as double contours in the iris region, incorrect lip region, etc. Two specialists also examined each postprocessed image to ensure the excellent quality of created morphed images produced. }\DIFaddend A sample illustration of a \DIFdelbegin \DIFdel{post-processed }\DIFdelend \DIFaddbegin \DIFadd{postprocessed }\DIFaddend image is provided in Figure \DIFdelbegin \DIFdel{~\ref{fig:post-processing-morph} }\DIFdelend \DIFaddbegin \DIFadd{4 }\DIFaddend where one can observe the naive morph generation resulting in visible \DIFdelbegin \DIFdel{artefacts }\DIFdelend \DIFaddbegin \DIFadd{artifacts }\DIFaddend in the iris and nose areas while the \DIFdelbegin \DIFdel{post-processed }\DIFdelend \DIFaddbegin \DIFadd{postprocessed }\DIFaddend image eliminates such \DIFdelbegin \DIFdel{artefacts}\DIFdelend \DIFaddbegin \DIFadd{artifacts}\DIFaddend , resulting in a realistic face image.

	\DIFdelbegin \DIFdel{Furthermore, as it is common practice in many countries to provide }\DIFdelend \DIFaddbegin \DIFadd{Additionally, since providing }\DIFaddend a printed image that is \DIFdelbegin \DIFdel{subsequently scanned , we have }\DIFdelend \DIFaddbegin \DIFadd{afterward scanned is a standard procedure in many nations, we }\DIFaddend created another subset of \DIFdelbegin \DIFdel{morphed images }\DIFdelend \DIFaddbegin \DIFadd{morphed photographs }\DIFaddend by printing and scanning them \DIFdelbegin \DIFdel{using }\DIFdelend \DIFaddbegin \DIFadd{with }\DIFaddend an Epson Expression Premium XP-7100 \cite{hp2020printer} photo printer. Thus the morphed image set has a total of 400 digital images and 400 printed-scanned images, both after \DIFaddbegin \DIFadd{postprocessing.}\DIFaddend

\begin{table*}[htp]
	\centering
	\resizebox{\linewidth}{!}{%
		\begin{tabular}{llcccccccc}
			\hline
			\multicolumn{2}{c}{\multirow{2}{*}{}} & \multicolumn{4}{c}{Digital} & \multicolumn{4}{c}{Print and scanned} \\ \cline{3-10} 
			\multicolumn{2}{c}{} & Bona fide & Morph & \begin{tabular}[c]{@{}c@{}}Post-processed   \\ morph\end{tabular} & \begin{tabular}[c]{@{}c@{}}ABC Gate   \\ images\end{tabular} &  Bona fide & Morph & \begin{tabular}[c]{@{}c@{}}Post-processed   \\ morph\end{tabular} & \begin{tabular}[c]{@{}c@{}}ABC Gate   \\ images\end{tabular} \\ \hline
			\hline
			\multirow{2}{*}{Male} & Individual subject & 40 & 40 & 40 & 40 & 40 & 40 & 40 & 40 \\
			& Total  images & 80 & 80 & 80 & 80 &  80 & 80 & 80 & 80 \\ \hline
			\multirow{2}{*}{Female} & Individual  subject & 8 + 8* & 8 & 8 & 8 + 8*  & 8 & 8 & 8 & 8 \\ 
			& Total images & 16 + 16* & 16 & 16 & 16 + 16* &  16 & 16 & 16 & 16 \\ \hline
			\multicolumn{2}{c}{Total}  & 112 & 96 & 96 & 96 &  96 & 96 & 96 & 96 \\ \hline
			\hline
			\multicolumn{2}{c}{Total} &  \multicolumn{4}{c}{416} &  \multicolumn{4}{c}{384} \\ \hline
			\hline
		\end{tabular}%
	}
	\caption{Distribution of images in the database. *Note: Additional bona fide images used for checking the consistency of user participation, i.e., if the observer is randomly selecting images as bona fide or morph in continuous trials.}
	\label{tab:d-mad-data-distribution}
\end{table*}

	\subsubsection{HOMID Single Image MAD Subset}
	While \DIFdelbegin \DIFdel{in certain application scenarios such as border control, a decision on the image as a bona fide or }\DIFdelend \DIFaddbegin \DIFadd{decisions about whether an image is genuine or a }\DIFaddend morph can be \DIFaddbegin \DIFadd{made }\DIFaddend based on an image pair \DIFaddbegin \DIFadd{in some application settings, such as border control }\DIFaddend (e.g.\DIFaddbegin \DIFadd{, }\DIFaddend with trusted image capture), \DIFdelbegin \DIFdel{multiple scenarios such as visa applications }\DIFdelend \DIFaddbegin \DIFadd{other scenarios, like visa }\DIFaddend or passport applications\DIFdelbegin \DIFdel{have only one single }\DIFdelend \DIFaddbegin \DIFadd{, just have one }\DIFaddend image without any reference \DIFdelbegin \DIFdel{image}\DIFdelend \DIFaddbegin \DIFadd{images}\DIFaddend . For instance, applying for a \DIFdelbegin \DIFdel{B1 }\DIFdelend \DIFaddbegin \DIFadd{Bl }\DIFaddend USA visa needs the applicant to upload the passport image, and this could potentially be a morphed image on which the decision has to be made. \DIFdelbegin \DIFdel{In order to }\DIFdelend \DIFaddbegin \DIFadd{To }\DIFaddend study the human observer performance for \DIFdelbegin \DIFdel{morphing attack detection }\DIFdelend \DIFaddbegin \DIFadd{MAD }\DIFaddend in such scenarios, we create a Single-Image Morphing Attack dataset which we refer to as the HOMID-S dataset. \DIFdelbegin \DIFdel{Similar to the }\DIFdelend \DIFaddbegin \DIFadd{The }\DIFaddend HOMID-S subset \DIFdelbegin \DIFdel{, the HOMID-S subset }\DIFdelend has bona fide and \DIFdelbegin \DIFdel{morph }\DIFdelend \DIFaddbegin \DIFadd{morphed }\DIFaddend image subsets, \DIFdelbegin \DIFdel{as explained }\DIFdelend \DIFaddbegin \DIFadd{similar to HOMID-D subset, as discussed }\DIFaddend below. In addition, to consider image capture conditions other than our dataset, we
	make use of \DIFaddbegin \DIFadd{the }\DIFaddend FRGC v2 dataset \DIFdelbegin \DIFdel{\mbox{
\cite{phillips2005overview} }\hspace{0pt}
}\DIFdelend \DIFaddbegin \DIFadd{\cite{phillips2005overview}} \DIFaddend to derive HOMID S-MAD dataset as explained further. 

	\begin{figure}[H]
		\centering
		\begin{minipage}[c]{\textwidth}
			\subfloat[]{%
				\includegraphics[width=0.13\columnwidth]{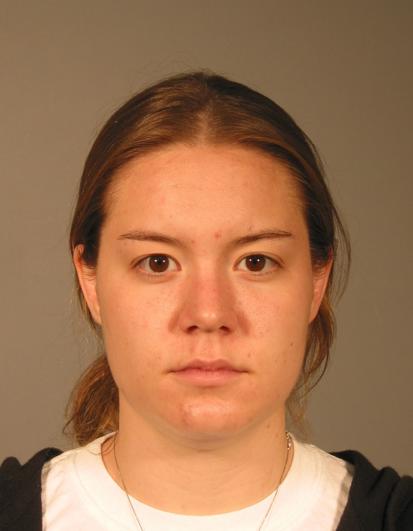}%
				\label{fig:s-mad-database-a}%
			} ~
			\subfloat[]{%
				\includegraphics[width=0.13\columnwidth]{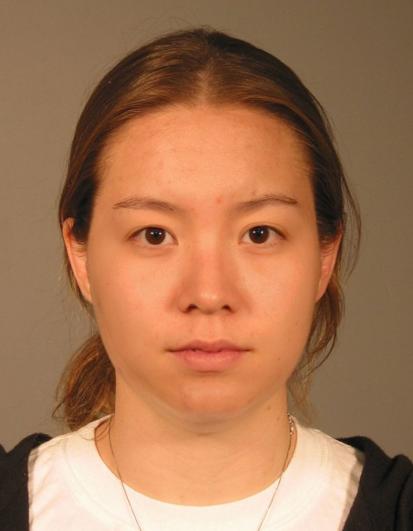}%
				\label{fig:s-mad-database-b}%
			} ~
			\subfloat[]{%
				\includegraphics[width=0.13\columnwidth]{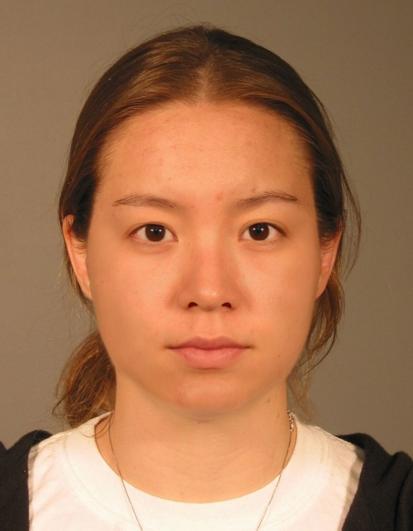}%
				\label{fig:s-mad-database-c}%
			} ~
			\subfloat[]{%
				\includegraphics[width=0.13\columnwidth]{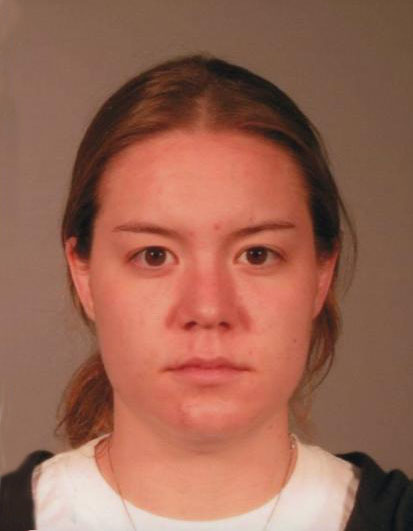}%
				\label{fig:s-mad-database-d}%
			} ~
			\subfloat[]{%
				\includegraphics[width=0.13\columnwidth]{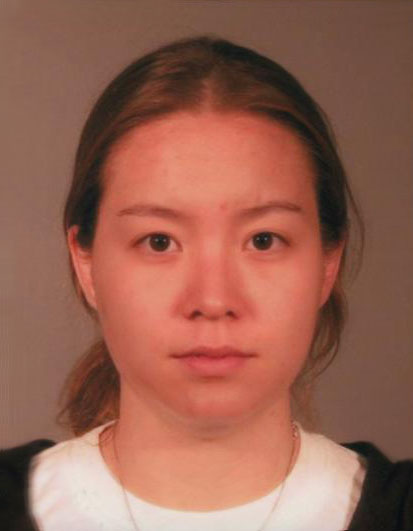}%
				\label{fig:s-mad-database-e}%
			}~ 
			\subfloat[]{%
				\includegraphics[width=0.13\columnwidth]{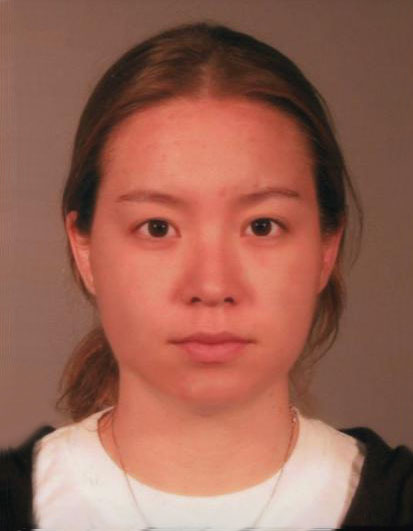}%
				\label{fig:s-mad-database-f}%
			}
		\end{minipage}
		\caption{Sample images for HOMID-S database. (a) Digital bona fide image, (b) Digital morphed image, (c) Digital post processed image, (d) Print and scanned bona fide image, (e) Print and scanned morphed image, (f) Print and scanned post processed image.}
		\label{fig:s-mad-database}
	\end{figure}

	\begin{figure*}[b]
		\centering 
		\subfloat[iHOPE D-MAD experimental setup. The participant is asked to make a decision and choose "Bona fide" or "Morphed" images given the "Trusted live capture" image and an "Unknown capture" image.]{%
			\includegraphics[width=0.48\textwidth]{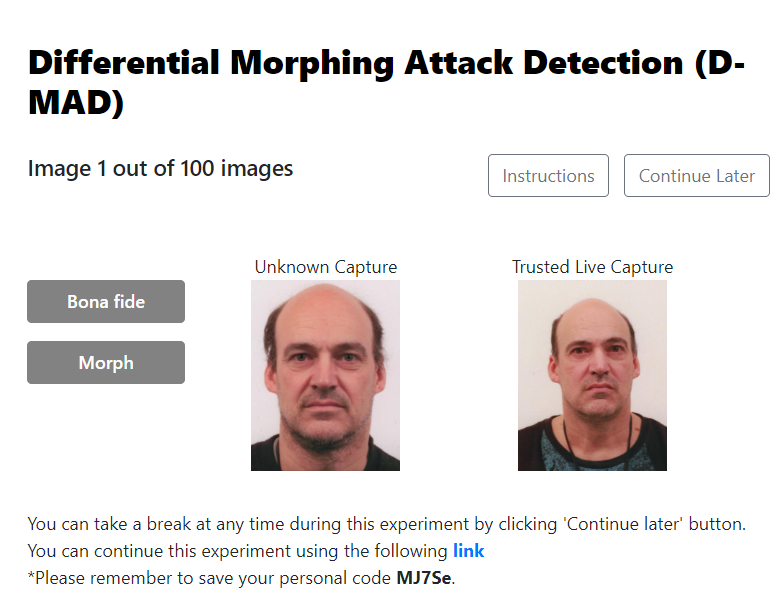}
			\label{fig:iHOPE-d-mad-page}
		}
		\quad
		\subfloat[iHOPE S-MAD experimental setup. The participant is asked to make a decision and choose "Bona fide" or "Morphed" images given one image from "Unknown capture" setting.]{%
			\includegraphics[width=0.40\textwidth]{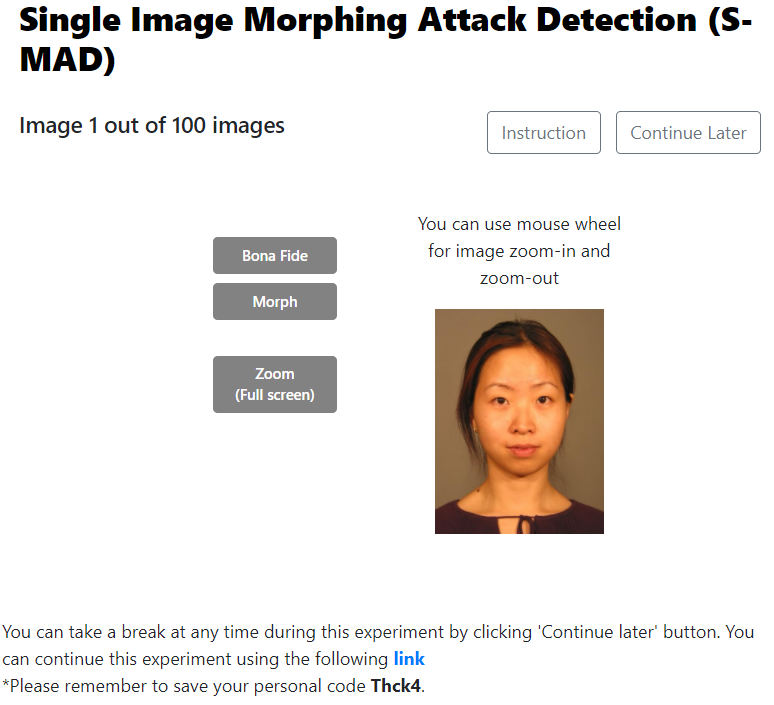}
			\label{fig:iHOPE-s-mad-page}
		}
		\caption{Graphical user interface for the human observers experimental setup used in this work.}
		\label{fig:iHOPE-platform}
	\end{figure*}

	\textbf{HOMID-S Bona fide Subset}: \DIFdelbegin \DIFdel{To create a dataset of bona fide images, we choose }\DIFdelend \DIFaddbegin \DIFadd{We selected }\DIFaddend 30 subjects \DIFaddbegin \DIFadd{(15 men and 15 women) }\DIFaddend from the FRGC v2 dataset \cite{phillips2005overview} \DIFdelbegin \DIFdel{with 15 male and 15 female subjects}\DIFdelend \DIFaddbegin \DIFadd{to produce a dataset of bona fide images}\DIFaddend . As the FRGC v2 dataset provides multiple captures for each subject, we choose a high-quality enrolment image for the bona fide set and further process each image to conform to ICAO standards for the face image. \DIFdelbegin \DIFdel{Care has been exercised to correct any rotation or poses in the chosen bona fide subset}\DIFdelend \DIFaddbegin \DIFadd{The selected bona fide subset has been carefully checked to make sure that any rotations or poses are corrected}\DIFaddend . The demographic distribution of the chosen subjects is provided in Table~\ref{tab:s-mad-database}.

	\begin{table}[H]
		\resizebox{\linewidth}{!}{%
			\begin{tabular}{lcccccc}
				\hline
				& \multicolumn{3}{c}{Digital}  & \multicolumn{3}{c}{Print and scanned} \\
				\cline{2-7} & Bona fide & Morph & Post-processed &   Bona fide & Morph & Post-processed  \\
				& & & Morph & & & Morph \\\hline
				\hline
				Male  & 15   & 15  & 15 & 15  & 15 & 15   \\ 
				Female  & 15   & 15  & 15 & 15  & 15 & 15   \\
				\hline 
				Total & \multicolumn{3}{c}{90}  & \multicolumn{3}{c}{90} \\
				\hline
				\hline
			\end{tabular}
		}
		\caption{Demographic distribution in the HOMID-S database}
		\label{tab:s-mad-database}
	\end{table}

	\textbf{HOMID-S Morph Subset}: We further generate morphed images for the chosen 30 subjects using a morphing factor of \DIFdelbegin \DIFdel{$0.5$. Each subject is morphed with another resembling subject with respect to }\DIFdelend \DIFaddbegin \DIFadd{0.5. To produce realistic morphed photos, each subject is merged with another subject that closely resembles it in terms of }\DIFaddend age, gender\DIFdelbegin \DIFdel{and ethnicity to create realistic morphed images}\DIFdelend \DIFaddbegin \DIFadd{, and race}\DIFaddend . Further, to assure the high quality of the morphed images, we carry out manual \DIFdelbegin \DIFdel{post-processing }\DIFdelend \DIFaddbegin \DIFadd{postprocessing }\DIFaddend with human expert intervention. \DIFdelbegin \DIFdel{As a result, we have created }\DIFdelend \DIFaddbegin \DIFadd{We consequently produced }\DIFaddend 15 morphed \DIFdelbegin \DIFdel{images for male subjects }\DIFdelend \DIFaddbegin \DIFadd{images for male individuals }\DIFaddend and 15 morphed images for female subjects. Each of the morphed images, which is \DIFdelbegin \DIFdel{post-processed}\DIFdelend \DIFaddbegin \DIFadd{postprocessed}\DIFaddend , is also printed and scanned in the lines of previous works \cite{raghavendra2018detecting, raja2020morphing, venkatesh2021morphsurvey} using an Epson Expression Premium XP-7100 \cite{hp2020printer} photo printer. 

	A sample illustration of the images in the HOMID-S subset is further provided in Figure~\ref{fig:s-mad-database} which depicts both bona fide and morphed images along with the post-processed and printed-and-scanned images.

	\subsection{\Rev{Platform:} iMARS Human Observer Platform for Evaluation - iHOPE}
	\label{sec:web-platform}
	\DIFdelbegin \DIFdel{In order to facilitate the large scale experiments, we have also created a customized }\DIFdelend \DIFaddbegin \DIFadd{We also developed an individualized }\DIFaddend Human Observer Platform for Evaluation (iHOPE) for \DIFdelbegin \DIFdel{morphed image detection}\DIFdelend \DIFaddbegin \DIFadd{morphing image identification to support large-scale trials}\DIFaddend . The new evaluation platform mimics the realistic operational scenario where the images are provided to observers to determine if the image is bona fide or morphed. \DIFdelbegin \DIFdel{Furthermore}\DIFdelend \DIFaddbegin \DIFadd{Additionally}\DIFaddend , the platform \DIFdelbegin \DIFdel{is designed following the strict guidelines of the General Data Protection Regulation (GDPR ) to protect and preserve participants' privacy with full considerations of the anonymity of participants}\DIFdelend \DIFaddbegin \DIFadd{was created under the tight restrictions of the GDPR to safeguard participants’ privacy while giving full regard to participants’ anonymity}\DIFaddend .

	\DIFdelbegin \DIFdel{The platform currently }\DIFdelend \DIFaddbegin \DIFadd{Currently, the platform }\DIFaddend supports two sets of experiments that correspond to \DIFaddbegin \DIFadd{the }\DIFaddend D-MAD and S-MAD scenarios\DIFdelbegin \DIFdel{and can be conducted }\DIFdelend \DIFaddbegin \DIFadd{. These experiments can be carried out }\DIFaddend on a desktop environment \DIFdelbegin \DIFdel{inspired by the current practices in operational scenarios}\DIFdelend \DIFaddbegin \DIFadd{that is modeled after the methods currently used in operational settings}\DIFaddend . The participants first need to register their details by providing their email address (voluntary), name, age, and gender. \DIFdelbegin \DIFdel{Each participant is provided with a short introduction to the experiment's goal on the homepage}\DIFdelend \DIFaddbegin \DIFadd{On the homepage, there is a summary of the experiment’s objectives for each participant}\DIFaddend . Moreover, each participant was asked to indicate whether they had any training in facial comparison, document examination and morphing, the length of said training, and their line of work. The platform first \DIFdelbegin \DIFdel{directs the participants }\DIFdelend \DIFaddbegin \DIFadd{takes users }\DIFaddend to the D-MAD experiment, where \DIFdelbegin \DIFdel{the participant can obtain scores and information about time spent after every 100 tasks}\DIFdelend \DIFaddbegin \DIFadd{they may view their results and learn how much time they spent on each task}\DIFaddend . Then, the platform presents a pair of images of a chosen subject by providing the trusted capture image (captured from an ABC gate) and an \DIFdelbegin \DIFdel{unknown capture image. While the trusted capture image is a bona fide image, the unknown capture image is randomly selected from the morphed }\DIFdelend \DIFaddbegin \DIFadd{unknown-capture image. The unknown-capture image is chosen at random from either a morphing }\DIFaddend set or a bona fide set of the \DIFdelbegin \DIFdel{respective subject}\DIFdelend \DIFaddbegin \DIFadd{relevant subject, whereas the trusted capture image is genuine}\DIFaddend . All the observers are presented same pairs in D-MAD experiments to derive the conclusions in this work. \DIFdelbegin \DIFdel{The images in the platform are presented in a random order for each participant}\DIFdelend \DIFaddbegin \DIFadd{Each participant sees the photos on the platform in a different order at random}\DIFaddend . The platform is also designed to collect meta-data on how much time the participants spend on each image pair. \DIFdelbegin \DIFdel{Care is exercised in the images used to avoid any decision bias due to the clothing of the presented images by showing the morphed images, bona fide images}\DIFdelend \DIFaddbegin \DIFadd{By providing the morphed images, bona fide images, }\DIFaddend and trusted capture (ABC) gate images with the same \DIFdelbegin \DIFdel{clothing }\DIFdelend \DIFaddbegin \DIFadd{clothes }\DIFaddend in both images for each pair of \DIFaddbegin \DIFadd{photographs, care is taken to avoid any decision bias caused by the clothing of the given }\DIFaddend images. Figure \DIFdelbegin \DIFdel{~\ref{fig:iHOPE-d-mad-page} }\DIFdelend \DIFaddbegin \DIFadd{6a }\DIFaddend illustrates the designed iHOPE-D-MAD evaluation page.
	\DIFaddbegin 

	\DIFadd{Considering various operational practices, the platform also allows the participants to zoom in and enlarge the images before taking a decision. The platform also enables the participants to take a break and resume the experiments as needed, which lessens the strain on their minds during lengthy cognitive experiments. The total score is shown in the D-MAD experiments every time the participant clicks on “Continue later.” Once the D-MAD experiment is over, the platform then refers users to the S-MAD experiment. Figure ~\ref{fig:iHOPE-d-mad-page} presents the iHOPE S-MAD evaluation platform where the participant is presented with one image from an unknown-capture setting of a subject. The subgroups of bona fide and morphed images from which the photographs are drawn are presented at random.
	}\DIFaddend 


	\subsection{\Rev{Participants/Observers:} iMARS Human Observer Evaluation}
	\label{sec:human-observer-evaluation}
	The online evaluation platform \DIFdelbegin \DIFdel{for benchmarking the human observer ability in detecting morphed images consists of two different setups corresponding to D-MAD and S-MAD. \Rev{The platform was designed to allow the participants to pause and resume the experiment at their own pace to accommodate busy schedules, avoid fatigue and cognitive load in making decisions for the experiment. }
	}\DIFdelend \DIFaddbegin \DIFadd{uses two alternative configurations, called D-MAD and S-MAD, to compare how well humans are at spotting morphed images. The platform was designed to allow the participants to pause and resume the experiment at their own pace to accommodate busy schedules, and avoid fatigue and cognitive load in making decisions for the experiment.
	}\DIFaddend 

	\paragraph{\Rev{\textbf{Recruitment of Observers}}}
	\DIFdelbegin \DIFdel{\Rev{The observers were recruited from trusted channels through national agencies who deal with ID management and verification on a daily basis. Invitations were sent through secure channels through liaisons at various agencies to create an observer pool of Border Guards (30), Case handlers - Passport, visas, ID, etc (150), Document examiners - 1st line (38), Document examiners - 2st line (40), Document examiners - 3rd line (30), Face comparison experts - Manual examination (44), ID Experts (53) and Others (84). The observers were made aware of the goals of the experiment through brief introduction and instructions. The participants were asked for explicit consent according the national guidelines and GDPR. In addition to acquiring the consent, the demographic information of the participants, their work experience and training duration was noted for the sake of our analysis through dedicated questionnaire. Further, each participant was given a unique code to anonymize the participant information, which could also be used to get the data deleted if the participant did not wish to allow his/her data to be used for analysis.}
	}\DIFdelend \DIFaddbegin \DIFadd{The observers were recruited from national organizations that regularly deal with ID management and verification. Invitations were sent through secure channels through liaisons at various agencies to create an observer pool of Border Guards (30), Case handlers - Passport, visas, ID, etc (150), Document examiners - 1st line (38), Document examiners - 2nd line (40), Document examiners - 3rd line (30), Face comparison experts - Manual examination (44), ID Experts (53) and Others (84). The experiment’s objectives were explained to the observers in a brief introduction and set of instructions. The participants were asked for explicit consent according to the national guidelines and GDPR. In addition to obtaining consent, a specific questionnaire was used to record participant's demographics, employment history, and training duration for our study. Further, each participant was given a unique code to anonymize the participant information, which could also be used to get the data deleted if the participant did not wish to allow his/her data to be used for analysis.
	}\DIFaddend 

	In the set of questionnaire, the participants were asked to provide the following information:
	\begin{itemize}
		\item Age, gender, email address and nationality
		\item Profession/Line of work
		\begin{itemize}
			\item Experience with facial examination ($< 6$ months, 6-12 months, 1-3 years, 3-5 years and more than 5 years)
			\item Experience document examination ($< 6$ months, 6-12 months, 1-3 years, 3-5 years and more than 5 years)
			\item Is the participant a documented super recogniser?
			\item Experience in morphing attack detection (Half a day, Half-day to one day, 2-3 days, 4-5 days, more than 5 days)
			\item Line of work (Face comparison, Fingerprint expert, Document examiner (1st, 2nd and 3rd line), ID Expert (Embassy, Police and other government authorities), Case handler (ID, Visa, Passport), Border guard (1st line, 2nd line))
		\end{itemize}
	\end{itemize}

		\begin{table}[H]
		\centering
		\resizebox{0.95\textwidth}{!}{%
			\begin{tabular}{lcc}
				\hline
				\hline
				Comparison & Digital images & Print and scanned \\ \hline
				Morph  vs Bona fide                  & 48             & 48  \\
				Morph   vs ABC gate image                 & 48             & 48   \\
				Post processed morph     & 48             & 48        \\ 
				image vs Bona fide  & & \\
				Post processed morph  & 48             & 48                \\ 
				vs ABC gate image & & \\
				Bona fide  vs Bona fide       & *16          & 0                 \\ \hline
				Total                                        & 208            & 192               \\ \hline\hline
			\end{tabular}%
		}
		\caption{Statistics of different comparison trials in D-MAD evaluation. * indicates the total trials to check the consistency of the participants.}
		\label{tab:d-mad-comparisons-trial}
	\end{table}

	\subsection{\Rev{Experiment-I:} D-MAD Human Observer Evaluation}
	\label{ssec:d-mad-experiment-design}
	\DIFdelbegin \DIFdel{This studyconsisted of }\DIFdelend \DIFaddbegin \DIFadd{In this study’s }\DIFaddend 400 image \DIFdelbegin \DIFdel{comparisons, and in each trial, two face images were shown }\DIFdelend \DIFaddbegin \DIFadd{comparison trials, two face photos were displayed }\DIFaddend side by side as \DIFdelbegin \DIFdel{depicted }\DIFdelend \DIFaddbegin \DIFadd{illustrated }\DIFaddend in Figure~\ref{fig:iHOPE-d-mad-page}. Each trial was randomized by choosing the image pairs randomly. 400 trials were divided into four parts, where each part consisted of 100 unique comparisons. \DIFdelbegin \DIFdel{The pair of images consisted of one image from a trusted live capture, and then the participants }\DIFdelend \DIFaddbegin \DIFadd{Participants }\DIFaddend were asked to \DIFdelbegin \DIFdel{decide }\DIFdelend \DIFaddbegin \DIFadd{determine }\DIFaddend if the second \DIFdelbegin \DIFdel{suspected image(from unknown capture) was 'bona fide' or 'morph image'}\DIFdelend \DIFaddbegin \DIFadd{image, which was assumed to be from an unreliable capture, was a “bona fide” or “morphed image” after viewing the first image from a trusted live capture}\DIFaddend . An optional zoom-in function was provided on the platform for participants who wished to enlarge the images before deciding, mimicking the real-life deployment solution. \DIFdelbegin \DIFdel{The achieved accuracy was displayed periodically after }\DIFdelend \DIFaddbegin \DIFadd{After }\DIFaddend 100 \DIFdelbegin \DIFdel{comparisons to motivate }\DIFdelend \DIFaddbegin \DIFadd{comparisons, the acquired accuracy was frequently displayed to encourage }\DIFaddend the participants.

	The D-MAD experiment consisted of 48 comparison trials in 5 different categories as listed in Table~\ref{tab:d-mad-comparisons-trial}. \DIFdelbegin \DIFdel{In addition to morphed image detection, the evaluation also introduced 8 }\DIFdelend \DIFaddbegin \DIFadd{The evaluation added eight }\DIFaddend random bona fide v/s bona fide comparisons \DIFdelbegin \DIFdel{to }\DIFdelend \DIFaddbegin \DIFadd{in addition to the identification of morphing images to }\DIFaddend conduct a consistency check \DIFdelbegin \DIFdel{if }\DIFdelend \DIFaddbegin \DIFadd{on whether }\DIFaddend a user was \DIFdelbegin \DIFdel{randomly clicking }\DIFdelend \DIFaddbegin \DIFadd{arbitrarily selecting }\DIFaddend the same image \DIFdelbegin \DIFdel{over and again }\DIFdelend \DIFaddbegin \DIFadd{repeatedly }\DIFaddend or if the clicks were random. The final analysis has removed all such data where a random click \DIFdelbegin \DIFdel{behaviour }\DIFdelend \DIFaddbegin \DIFadd{behavior }\DIFaddend was observed.

	\subsection{\Rev{Experiment-II:} S-MAD Human Observer Evaluation}
	\label{ssec:s-mad-experiment-design}
	In the S-MAD human observer evaluation, a total of 180 individual images were presented to each observer. The participants were \DIFdelbegin \DIFdel{provided with just one imagebased on which a decision had to be made if the image was bona fide or morphed}\DIFdelend \DIFaddbegin \DIFadd{only given one image}\DIFaddend , as shown in Figure~\ref{fig:iHOPE-s-mad-page} \DIFaddbegin \DIFadd{and were required to decide whether it was a genuine photograph or a morphed image}\DIFaddend . The face images displayed in the evaluation platform \DIFdelbegin \DIFdel{was }\DIFdelend \DIFaddbegin \DIFadd{were }\DIFaddend rescaled to EU passport image standards (\DIFdelbegin \DIFdel{$413\times531$ }\DIFdelend \DIFaddbegin \DIFadd{413 × 531 }\DIFaddend pixels and JPEG2000 compression) \cite{bako2006jpeg}. The participants \DIFdelbegin \DIFdel{were provided with a zoom functionality where the image was blown to full screen , considering an operational scenario where a border guard/examiner would do a similar operation }\DIFdelend \DIFaddbegin \DIFadd{had access to a zoom feature that enlarges the image to fill the entire screen as they analyze how a border officer or examiner might conduct a comparable operation in real life}\DIFaddend . The participants further received the final scores after the completion of 180 trials and intermediate scores after 90 trials.

	\section{Findings, Analysis and Discussion}
	\label{sec:findings-discussion}
	This section \DIFdelbegin \DIFdel{presents the findings from experiments and provides a detailed }\DIFdelend \DIFaddbegin \DIFadd{covers the results of the tests and offers a thorough }\DIFaddend analysis of the \DIFdelbegin \DIFdel{trends observed in human observer evaluation }\DIFdelend \DIFaddbegin \DIFadd{patterns seen in the evaluation of human observers}\DIFaddend . The analysis is based on 469 observers in the D-MAD experiment and 410 observers in the S-MAD experiment. The distribution of a different line of work is presented in Table ~\ref{tab:d-mad-s-mad-observer-distribution} in D-MAD and S-MAD experiments. \DIFdelbegin \DIFdel{Further, we present a detailed discussion on each subset of findingsfor the benefit of the reader}\DIFdelend \DIFaddbegin \DIFadd{For the reader’s benefit, we also provide a thorough discussion of each subgroup of findings}\DIFaddend .

	\begin{table*}[htbp]
		\centering
		\caption{Distribution of participants in D-MAD and S-MAD experiments}
		\begin{tabular}{lcccc}
			\hline
			\hline
			& \multicolumn{2}{c}{D-MAD} & \multicolumn{2}{c}{S-MAD} \\
			\cline{2-5}
			& Number of & Average & Number of & Average \\
			Line of work & participants &  Accuracy  & participants &  Accuracy\\
			\hline
			\hline
			Border Guard & 30 & 64.66 & 26 & 55.17 \\ 
			Case handler- Passport, visas, ID, etc & 150 & 63.45 & 137 & 56.65\\
			Document examiner- 1st line & 38 & 60.79 & 30 & 57.63 \\
			Document examiner- 2st line & 40 & 68.64 & 34 & 62.56 \\
			Document examiner- 3rd line & 30 & 65.74 & 25 & 61.51 \\
			Face comparison expert (Manual examination) & 44 & 72.56 & 39 & 64.63 \\ 
			ID Expert & 53 & 63.09 & 50 & 57.21 \\
			Other & 84 & 64.66 & 69 & 55.17 \\
			Student & 103 & 56.91 & - & - \\ 
			\hline
			\hline
			Total participants & 572 &   & 410 &  \\
			Experts & 469 &   & 410 &  \\
			\hline
			\hline
		\end{tabular}%
		\label{tab:d-mad-s-mad-observer-distribution}%
	\end{table*}%

	\subsection{Metrics for Evaluations}
	 \DIFdelbegin \DIFdel{We employ }\DIFdelend \DIFaddbegin \DIFadd{To show the results, we use Attack Presentation Classification Error Rate (APCER) and Bona fide Presentation Classification Error Rate (BPCER). APCER is defined as as the proportion of attacks classified as bona fide and BPCER is the proportion of bona fide images classified as attacks.}\DIFaddend 
	The accuracy is reported as an average of Attack Presentation Classification Error Rate (APCER) and Bona fide Classification Error Rate (BPCER) aligning the results to NIST FRVT MORPH challenge \cite{NistFrvtMorph} and is defined as below:
	\begin{equation}
		Accuracy = \frac{(1-APCER)+(1-BPCER)}{2}
	\end{equation}

	\subsection{\Rev{Overall Accuracy in D-MAD and S-MAD}}
\begin{figure}[h]
		\centering 
		\includegraphics[trim=0 0 0 0, width=.85\linewidth]{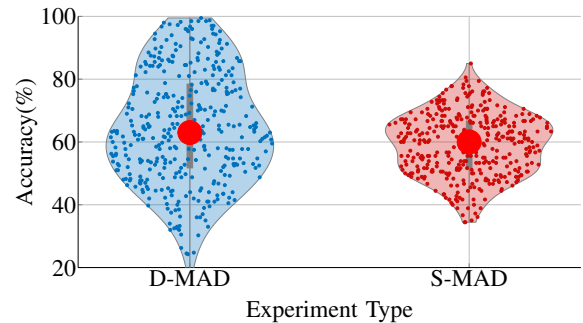}%
		\caption{Overall accuracy of morphed image detection for D-MAD and S-MAD experiments}
		\label{fig:d-mad-s-mad-accuracy}
	\end{figure}

	\DIFdelbegin \DIFdel{\Rev{We first present the overall detection accuracy for D-MAD and S-MAD before analyzing each of them individually.  As it can be noted from Figure~\ref{fig:d-mad-s-mad-accuracy}, the average accuracy of detecting the morphing attack in D-MAD setting is \Rev{64.10\% (\textbf{min} = 19.75\%, \textbf{max}= 99.50\%, SD=17.71\%, \textit{Mdn}=62.75\%)} while few observers have more than 95\% accuracy. We note a slightly lower average accuracy in S-MAD setting corresponding to \Rev{58.98\% (min = 33.33\%, max = 85\%, SD=9.75\%)}. Unlike the D-MAD setting, we note that none of the observers have more than 82\% detection accuracy in S-MAD and this can be attributed to challenging decision making process in the absence of reference image. With the noted observations, we analyse multiple factors as explained in upcoming sections.}
	}\DIFdelend \DIFaddbegin \DIFadd{We first present the overall detection accuracy for D-MAD and S-MAD before analyzing each of them individually. Figure 7 shows that while few observers have more than 95\% accuracy, the average accuracy of identifying the morphing attack in the D-MAD condition is 64.10\% (\textit{min}=19.75\%, \textit{max}=99.50\%, \textit{SD}=17.71\%, \textit{Mdn}=62.75\%). We note a slightly lower average accuracy in the S-MAD setting corresponding to 58.98\%
	(\textit{min}=33.33\%, \textit{max}=85\%, \textit{SD}=9.75\%, \textit{Mdn}=59.44\%). Unlike the D-MAD setting, we note that none of the observers have more than 82\% detection accuracy in S-MAD and this can be attributed to the challenging decision-making process in the absence of a reference image. We analyze and present several factors in the following subsections based on these observations.
	}\DIFaddend

	\subsection{Training in facial comparison v/s MAD}
	\DIFdelbegin \DIFdel{As }\DIFdelend \DIFaddbegin \DIFadd{Since }\DIFaddend facial examination expert \DIFdelbegin \DIFdel{observers are expected to master the task of }\DIFdelend \DIFaddbegin \DIFadd{viewers are considered to be skilled in }\DIFaddend facial comparison, we \DIFdelbegin \DIFdel{anticipated that factors such as length of }\DIFdelend \DIFaddbegin \DIFadd{predicted that factors like the amount of time spent on }\DIFaddend topic-specific training and \DIFdelbegin \DIFdel{length of experience in facial examination play a role in their detection accuracy. Therefore}\DIFdelend \DIFaddbegin \DIFadd{the amount of facial examination experience a person has will affect how accurately they detect face morphs. Therefore, }\DIFaddend we analyze the MAD accuracy of human observers for both D-MAD and S-MAD experiments. \DIFdelbegin \DIFdel{Specifically, we categorize the observers }\DIFdelend \DIFaddbegin \DIFadd{We specifically, group the observers into six groups }\DIFaddend based on the \DIFdelbegin \DIFdel{amount of training undergone in six different categories}\DIFdelend \DIFaddbegin \DIFadd{length of training they have received}\DIFaddend : no prior experience, less than 6 months, \DIFdelbegin \DIFdel{6-12 months, 1-3 years, 3-5 }\DIFdelend \DIFaddbegin \DIFadd{6–12 months, 1–3 years, 3-5 }\DIFaddend years, and more than 5 years. We make the following observations from the results obtained:
	\subsubsection{\Rev{Results for Experiment-I - D-MAD}}
	\begin{itemize}
		\item \DIFdelbegin \DIFdel{\Rev{The observers with no specific training on face comparison obtain a MAD accuracy of \Rev{62.81\% (min = 19.75\%, max = 98.50\%, SD=17.69\%, Med=61.50\%)}. In contrast, the observers with experience obtain an average accuracy of 64.89\% (averaged over five different experience groups, \Rev{\textit{min}=23.75\%, \textit{max}=99.50\%, \textit{SD}=17.71\%, \textit{Mdn}=63.37\%)} in D-MAD settings as noted in Figure~\ref{fig:d-mad-training-accuracy}.}
		}\DIFdelend \DIFaddbegin \DIFadd{The observers with no specific training on face comparison obtain a MAD accuracy of 62.81\% (\textit{min}=19.75\%, \textit{max}=98.50\%, \textit{SD}=17.69\%, \textit{Mdn}=61.50\%). In contrast, experienced observers in D-MAD settings, as shown in Figure~\ref{fig:d-mad-training-accuracy}, achieve an average accuracy of 64.89\% (averaged over five different experience groups, (\textit{min}=23.75\%, \textit{max}=99.50\%, \textit{SD}=17.71\%, \textit{Mdn}=63.37\%).
		}\DIFaddend \item \DIFdelbegin \DIFdel{\Rev{To evaluate the differences of morphing attack detection across observers with different training length for facial examination, we conduct Kruskal-Wallis Test \cite{kruskal1952use}. The test revealed no significant differences (\Rev{\textit{H}(5)=7.85, \textit{p}=.167}) across the groups (n = 173, 74, 30, 77, 44, 71).}
		}\DIFdelend \DIFaddbegin \DIFadd{We use the Kruskal–Wallis test }[\DIFadd{41}] \DIFadd{to assess how different observers with various levels of face examination training detect morphing attacks. The test revealed no significant differences (\textit{H}(5)=7.85, \textit{p}=.167) across the groups (n = 173, 74, 30, 77, 44, 71).
		}\DIFaddend \item \DIFdelbegin \DIFdel{\Rev{We further investigate if longer training length in facial examination helps in detecting morphing attacks and we conduct another  Kruskal-Wallis Test \cite{kruskal1952use}. The test revealed no significant differences (\Rev{\textit{H}(1)=1.01, \textit{p}=.315}) across the groups (Less than 1 year training = 277, More than 1 year training = 192). And a similar observation is also made for observers with more than 5 years of training with a (\Rev{\textit{H}(1)=0.5, \textit{p}=.418}, Less than 5 years = 398, More than 5 years = 71).}
	}\DIFdelend \DIFaddbegin \DIFadd{We further investigate if longer training length in facial examination helps in detecting morphing attacks and we conduct another Kruskal–Wallis test }[\DIFadd{41}]\DIFadd{. The test found no significant differences between the groups (Less than 1 year of training = 277, More than 1 year of training = 192) (\textit{H}(1)=1.01, \textit{p}=.315). And a similar observation is also made for observers with more than 5 years of training with a (\textit{\textit{H}}(1)=0.5, \textit{p}=.418, Less than 5 years = 398, More than 5 years = 71).}\DIFaddend 
\end{itemize}

	\subsubsection{\Rev{Results for Experiment-II - S-MAD}}
	\begin{itemize}
		\item \DIFdelbegin \DIFdel{\Rev{The average MAD accuracy with respect to facial examination training in the S-MAD setting is 59.77\% (averaged over five different length categories, \Rev{\textit{min}=36.11\%, \textit{max}=80.56\%, \textit{SD}=9.09\%, \textit{Mdn}=60\%)}. At the same time, observers with no experience had an accuracy of 57.98\% \Rev{(\textit{min}=33.33\%, \textit{max}=85\%, \textit{SD}=10.70\%, \textit{Mdn}=57.78\%)} as presented in Figure~\ref{fig:s-mad-training-accuracy}.}
		}\DIFdelend \DIFaddbegin \DIFadd{The average MAD accuracy with respect to facial examination training in the S-MAD setting is 59.77\% (averaged over five different length categories, (\textit{min}=36.11\%, \textit{max}=80.56\%, \textit{SD}=9.09\%, \textit{Mdn}=60\%)). At the same time, observers with no experience had an accuracy of 57.98\%	(\textit{min}=33.33\%, \textit{max}=85\%, \textit{SD}=10.70\%, \textit{Mdn}=57.78\%) as presented in Figure~\ref{fig:s-mad-training-accuracy}.
		}\DIFaddend \item \DIFdelbegin \DIFdel{\Rev{Observations for different training length for facial examination in S-MAD setting indicated no significant differences (\Rev{\textit{H}(5)=13.63, \textit{p}=.018}) across the groups (n = 155, 62, 27, 65, 41, 60) with no-training to more than 5 years of training. No significant differences (\Rev{H(1)=0.88, \textit{p}=.348}) across the groups (Less than 1 year training = 244, More than 1 year training = 166) were further observed and in the similar lines, no significant differences were observed for observers with more than 5 years of training as compared to less than 5 years of training (\Rev{\textit{H}(1)=2.39, \textit{p}=.122}, Less than 5 years = 350, More than 5 years = 60).}
	}\DIFdelend \DIFaddbegin \DIFadd{Observations for different training lengths for facial examination in the S-MAD setting indicated no significant differences (\textit{H}(5)=13.63, \textit{p}=.018) across the groups (n = 155, 62, 27, 65, 41, 60) with no training to more than 5 years of training. No significant differences (\textit{H}(1)=0.88, \textit{p}=.348) across the groups (Less than 1-year training = 244, More than 1-year training = 166) were further observed and in similar lines, when comparing observers with more than 5 years of training to those with fewer than 5 years, no discernible changes were found (\textit{H}(1)=2.39, \textit{p}=.122, Less than 5 years = 350, More than 5 years = 60).
	}\DIFaddend \end{itemize}

	\subsubsection{\Rev{Common Observations}}
	\DIFdelbegin \DIFdel{\Rev{While we do not see major differences amongst various observers with prior experience, we note that the participants (74 in D-MAD and 62 in S-MAD) with less than 6 months of training in facial comparison (62.52\% in D-MAD, 61.53\% in S-MAD) and 6-12 months (67.59\% in D-MAD, 62.53\% in S-MAD) obtain the highest accuracy.  Our observation indicates that training in facial comparison does not necessarily lead to better MAD. To our knowledge, no comprehensive training programme exists yet \Rev{for MAD by humans}. Generally, MAD is just one part of document examination training or face comparison training. }
}\DIFdelend \DIFaddbegin \DIFadd{The participants (74 in D-MAD and 62 in S-MAD) with less than 6 months of training in face comparison (62.52\% in D-MAD, 61.53\% in S-MAD) showed no significant differences among the different observers to observers with 6–12 months of training (67.59\% in D-MAD, 62.53\% in S-MAD).  However, the observers with 6–12 months of training performed best in both categories. Our observation indicates that training in facial comparison does not necessarily lead to better MAD. As far as we are aware, there is currently no comprehensive training program for MAD by humans. Generally, MAD is just one part of document examination training or face comparison training.
	}\DIFaddend

	\begin{figure*}[h]
		\centering 
		\subfloat[D-MAD Accuracy]{%
			\includegraphics[trim=0 0 0 0, width=.5\linewidth]{figures-210915/d_mad_facial_trainig_vs_acc_violin.tex}%
			\label{fig:d-mad-training-accuracy}%
		}
		\subfloat[S-MAD Accuracy]{%
			\includegraphics[trim=0 0 0 0, width=.5\linewidth]{figures-210915/s_mad_facial_trainig_vs_acc_violin.tex}%
			\label{fig:s-mad-training-accuracy}%
		}
		\caption{Impact of training in facial examination on MAD accuracy of human observers.}
		\label{fig:mad-vs-training}
		\resizebox{0.65\textwidth}{!}{
		\begin{tabular}{lcccccccccc}
			\hline
			& \multicolumn{5}{c}{D-MAD}  & \multicolumn{5}{c}{S-MAD} \\
			\cline{2-11}& \#  & Acc & Min & Max & SD & \# & Acc & Min & Max & SD \\
			\hline
			No 			& 173 & 62.81 & 19.75 & 98.50 &	17.69		& 155 & 57.98 & 33.30 & 85.00 & 10.79\\
			< 6 months	& 74  & 62.52 & 36.00 &	98.25 &	16.60		& 62 & 61.53  & 38.89 & 78.33 & 8.52\\
			6-12 months & 30  & 67.59 & 26.25 & 98.75 &	19.49		& 27 & 62.53  & 38.89 & 80.56 & 10.48\\
			1-3 years 	& 77  & 64.65 & 24.25 & 99.50 &	18.29		& 65 & 60.21  & 41.67 & 76.11 & 7.83\\
			3-5 years 	& 44  & 61.14 & 33.00 & 91.25 &	14.85		& 41 & 57.68  & 40.00 & 76.67 & 8.81\\
			> 5 years	& 71  & 62.50 & 23.75 & 98.50 &	18.65		& 56 & 56.94  & 36.11 & 75.00 & 9.77\\
			\hline
		\end{tabular}
		}
	\end{figure*}

	\subsection{Experience in document examination v/s MAD}
	We also evaluate the impact of the experience of document examiners on the accuracy \DIFdelbegin \DIFdel{in }\DIFdelend \DIFaddbegin \DIFadd{of }\DIFaddend detecting the morphed images. We \DIFdelbegin \DIFdel{categorize }\DIFdelend \DIFaddbegin \DIFadd{divide }\DIFaddend the document examiners \DIFaddbegin \DIFadd{into six groups }\DIFaddend based on the \DIFdelbegin \DIFdel{training duration, }\DIFdelend \DIFaddbegin \DIFadd{length of their training }\DIFaddend specifically in document examination, \DIFdelbegin \DIFdel{in six different categories: }\DIFdelend no prior experience, less than 6 months, \DIFdelbegin \DIFdel{6-12 months, 1-3 years, 3-5 }\DIFdelend \DIFaddbegin \DIFadd{6–12 months, 1–3 years, 3–5 }\DIFaddend years, and more than 5 years. We make the following observations from the results obtained.
	\subsubsection{\Rev{Results for Experiment-I - D-MAD}}
	\begin{itemize}
		\item \DIFdelbegin \DIFdel{\Rev{The observers with specific training on document examination reached an average accuracy of 65.12\% in D-MAD (258 participants, \textit{min}=19.75\%, \textit{max}=98.50\%, \textit{SD}=16.51\%, \textit{Mdn}=62.25\%). 
		\item The observers with less than 6 months (43 participants) and 6-12 months (22 participants) training in document examination obtained a better average accuracy of 65.31\% and 67.57\% in the D-MAD setting respectively. However, the average accuracy of all the observers with training is 63.95\%, as noted in Figure~\ref{fig:d-mad-document-training-accuracy}.}
		}\DIFdelend \DIFaddbegin \DIFadd{The observers with specific training on document examination reached an average accuracy of 65.12\% in D-MAD (258 participants, \textit{min}=19.75\%, \textit{max}=98.50\%, \textit{SD}=16.51\%, \textit{Mdn}=62.25\%). Comparatively, the untrained observers had an average accuracy of 59.27\% (186 participants, \textit{min}=65.05\%, \textit{max}=99.50\%, \textit{SD}=23.75\%, \textit{Mdn}=63.50\%).
		}\DIFaddend \item \DIFdelbegin \DIFdel{\Rev{To evaluate the differences of morphing attack detection across observers with different training length in document examination, we conduct Kruskal-Wallis Test similar to previous section\cite{kruskal1952use}. The test revealed no significant differences (\Rev{\textit{H}(5)=5.23, \textit{p}=.388}) across the groups (n = 211, 43, 22, 66, 45, 82 in the order of no training to more than 5 years of training, refer Fig~\ref{fig:d-mad-document-training-accuracy}.}
		}\DIFdelend \DIFaddbegin \DIFadd{The observers with less than 6 months (43 participants) and 6–12 months (22 participants) training in document examination obtained a better average accuracy of 65.31\% and 68.12\% in the D-MAD setting. However, the average accuracy of all the observers with training is 63.95\%, as noted in Figure ~\ref{fig:d-mad-document-training-accuracy}.
		}\DIFaddend \item \DIFdelbegin \DIFdel{\Rev{Further investigations on if longer training length in document examination helps in detecting morphing attacks revealed no significant differences (\textit{H}(1)=1.94, \textit{p}=0.164) across the groups (Less than 1 year training = 254, More than 1 year training = 193). And a similar observation was also noted for observers with more than 5 years of training with a \Rev{\textit{H}(5)=0.02, \textit{p}=.879} (Less than 5 years = 387, More than 5 years = 82) using Kruskal-Wallis test.}
	}\DIFdelend \DIFaddbegin \DIFadd{We use the Kruskal–Wallis test, as described in the preceding section \mbox{
\cite{kruskal1952use}}\hspace{0pt}
, to assess the variations in MAD across observers with varying training lengths in document analysis. The test revealed no significant differences (\textit{H}(5)=5.23, \textit{p}=.388) across the groups (n = 211, 43, 22, 66, 45, 82 in the order of no training to more than 5 years of training, refer to Fig~\ref{fig:d-mad-document-training-accuracy}.
		}\item \DIFadd{Further research into whether longer training durations in document examination aid in the detection of morphing attacks found no appreciable changes (\textit{H}(1)=1.94, \textit{p}=0.164) across the groups (Less than 1 year training = 254, More than 1 year training = 193). Additionally, a similar finding was made with observation with more than 5 years of training
		with an \textit{H}(5)=0.02, \textit{p}=.879 (Less than 5 years = 387, More than 5 years = 82) using the Kruskal–Wallis test. 
	}\DIFaddend \end{itemize}

	\DIFaddend \begin{figure*}[h]
	\centering 
	\subfloat[D-MAD Accuracy]{%
		\includegraphics[trim=0 0 0 0, width=.5\linewidth]{figures-210915/d_mad_document_training_vs_violin.tex}
		\label{fig:d-mad-document-training-accuracy}%
	}
	\subfloat[S-MAD Accuracy]{%
		\includegraphics[trim=0 0 0 0, width=.5\linewidth]{figures-210915/s_mad_document_training_vs_violin.tex}
		\label{fig:s-mad-document-training-accuracy}%
	}
	\caption{Impact of training in document examination on MAD accuracy of human observers.}
	\label{fig:mad-vs-training-document}
	\resizebox{0.65\textwidth}{!}{
		\begin{tabular}{lcccccccccc}
			\hline
			& \multicolumn{5}{c}{D-MAD}  & \multicolumn{5}{c}{S-MAD} \\
			\cline{2-11}& \#  & Acc & Min & Max & SD & \# & Acc & Min & Max & SD \\
			\hline
			No 			& 211 & 65.05   &  23.75 &  99.50  & 19.08  & 191 & 59.35  & 33.30 & 85.00 & 10.59\\
			< 6 months	& 43  & 65.31   &  32.50 &  97.50  & 17.06 	& 36 & 59.72   & 38.89 & 73.89 & 9.15\\
			6-12 months & 22  & 67.57   &  41.00 &  92.50  & 13.57	& 20 & 62.53   & 48.33 & 79.44 & 8.68\\
			1-3 years 	& 66  & 60.03   &  24.50 &  96.00  & 16.66	& 55 & 57.72   & 41.67 & 73.33 & 7.46\\
			3-5 years 	& 45  & 62.76   &  33.00 &  95.25  & 15.72	& 41 & 59.62   & 43.33 & 77.22 & 8.44\\
			> 5 years	& 82  & 64.10   &  19.75 &  98.50  & 16.86	& 67 & 57.12   & 34.44 & 79.44 & 10.24\\
			\hline
		\end{tabular}
	}
\end{figure*}

	\subsubsection{\Rev{Results for Experiment-II - S-MAD}}
	\begin{itemize}
		\item \DIFdelbegin \DIFdel{\Rev{We note a drop in average accuracy in the S-MAD setting for observers with no prior training (186 participants) in document examination with an accuracy of 59.27\% compared to 65.12\% in the D-MAD setting. The drop in accuracy indicates the challenging nature of the S-MAD setting without any reference image to compare against.}
		}\DIFdelend \DIFaddbegin \DIFadd{We note a drop in average accuracy in the S-MAD setting for observers with no prior training (186 participants) in document examination with an accuracy of 59.27\% compared to 65.12\% in the D-MAD setting. The decrease in accuracy highlights how difficult the S-MAD setting is when there is no reference image to compare it against.
		}\DIFaddend \item \DIFdelbegin \DIFdel{\Rev{Not different from D-MAD is the observation for S-MAD setting where no significant differences were observed among groups of observers with different training length in document examination with a (\Rev{\textit{H}(5)=6.61, \textit{p}=.250}, n = 191, 36, 20, 55, 41, 67, Refer~\ref{fig:s-mad-document-training-accuracy}) with no-training to more than 5 years of training. No significant differences were further observed for groups with less than 1 year training (n=56) and more than 1 year training (n=163) where (\Rev{\textit{H}(1)=3.79, \textit{p}=.051}. Along the same lines, no significant differences were observed for observers with more than 5 years of training as compared to less than 5 years of training ((\Rev{\textit{H}(1)=2.82, \textit{p}=.093}, Less than 5 years = 343, More than 5 years = 67).}
	}\DIFdelend \DIFaddbegin \DIFadd{The observation for the S-MAD setting where no significant variations were seen among groups of observers with various training lengths in document evaluation with a similar methodology to D-MAD (\textit{H}(5)=6.61, \textit{p}=.250, n = 191, 36, 20, 55, 41, 67) with no training to more than 5 years of training as shown in Figure~\ref{fig:s-mad-document-training-accuracy}. No significant differences were further observed for groups with less than 1 year training (n = 56) and more than 1 year training (n = 163) where (\textit{H}(1)=3.79, \textit{p}=.051). In a similar manner, no discernible variations were found between observers with more than 5 years of training and those with less than 5 years of instruction (\textit{H}(1)=2.82, \textit{p}=.093, Less than 5 years = 343, More than 5 years = 67).
		}

	\DIFaddend \end{itemize}

	\subsubsection{Common Observations}
	\DIFdelbegin \DIFdel{\Rev{Similar to D-MAD setting, the accuracy of observers with 6-12 months (20 participants) training in document examination obtain higher average accuracy of 62.53\%. While the average accuracy of all observers with training in document examination is 58.66\% as noted in Figure~\ref{fig:s-mad-document-training-accuracy} and is lower than the accuracy in the D-MAD setting, the observation reiterates the challenge of detecting the morphed images in an S-MAD setting. Despite no major differences in average accuracy for observers with face comparison training and document examination, the individual observers with face comparison training perform better in detecting morphs than document examiners. While the initial indications point to better competence of one category over the other, it should be noted with the caveat that there is a need for a detailed investigation.}
	}\DIFdelend 

	\DIFaddbegin \DIFadd{Similar to the D-MAD setting, the accuracy of observers with 6–12 months (20 participants) training in document examination obtain a higher average accuracy of 62.53\%. The finding confirms the difficulty of recognizing the morphing images in an S-MAD setting, although the average accuracy of all observers with training in document examination is 58.66\% as shown in Figure~\ref{fig:s-mad-document-training-accuracy}, which is lower than the accuracy in the D-MAD setting, the observation reiterates the challenge of detecting the morphed images in an S-MAD setting. Despite no major differences in average accuracy for observers with face comparison training and document examination, the individual observers with face comparison training perform better in detecting morphs than document examiners. Although the early signs suggest that one category is more competent than the other, it should be highlighted that there is a need for a detailed investigation.
	}

	\subsection{Line of Work v/s MAD Accuracy}
	\DIFdelbegin \DIFdel{Further, we analyse the }\DIFdelend \DIFaddbegin \DIFadd{Additionally, we examine the categories of border guards and case handlers to determine the }\DIFaddend accuracy of MAD based on the line of work \DIFdelbegin \DIFdel{by looking at categories of border guards, case handlers }\DIFdelend (for Passport, visas, ID, etc.), document examiners - 1st line, document examiner- 2nd line, expert document examiners- 3rd line, face comparison experts (Manual examination), 3rd line, ID experts and miscellaneous as another category. \DIFaddbegin \DIFadd{According to their area of work, Figures }\DIFaddend Figure~\ref{fig:d-mad-line-of-work} and Figure~\ref{fig:s-mad-line-of-work} \DIFdelbegin \DIFdel{presents }\DIFdelend \DIFaddbegin \DIFadd{show }\DIFaddend the accuracy of human observers\DIFdelbegin \DIFdel{based on the line of work. Our observations from this set of analyses are noted below}\DIFdelend . \DIFaddbegin \DIFadd{Below are some observations we made as a result of these analyses.
	}\DIFaddend 

	\subsubsection{\Rev{Results for Experiment-I - D-MAD}}
	\begin{itemize}
		\item \DIFdelbegin \DIFdel{\Rev{The observers who work on face comparison (manual examination) have the highest average accuracy of 72.56\% (44 participants, \textit{min}=23.75\%, \textit{max}=99.50\%, \textit{SD}=20.97\%, \textit{Mdn}=77-25\%), with 12 participants obtaining more than 90\% accuracy in the D-MAD setting.}
		}\DIFdelend \DIFaddbegin \DIFadd{The observers who work on face comparison (manual examination) have the highest average accuracy of 72.56\% (44 participants, \textit{min}=23.75\%, \textit{max}=99.50\%, \textit{SD}=20.97\%, \textit{Mdn}=77.25\%), with 12 participants obtaining more than 90\% accuracy in the D-MAD setting.
		}\DIFaddend \item \DIFdelbegin \DIFdel{\Rev{While a high accuracy is noted for case handlers of passports, visas, and ID with 11 participants obtaining more than 90\% accuracy, 26\% of participants (total 150 participants) obtain less than 50\% accuracy in detecting morphs in the D-MAD setting.}
		}\DIFdelend \DIFaddbegin \DIFadd{While a high accuracy is noted for case handlers of passports, visas, and ID with 11 participants obtaining more than 90\% accuracy, 26\% of participants (a total of 150 participants) obtain less than 50\% accuracy in detecting morphs in the D-MAD setting.
		}\DIFaddend \item \DIFdelbegin \DIFdel{\Rev{It should also be noted that the border guards have reasonable accuracy, albeit a small portion being unable to detect the morphs in the D-MAD setting.}
	}\DIFdelend \DIFaddbegin \DIFadd{Additionally, it should be mentioned that the border patrol agents have respectable accuracy, despite a tiny percentage of them being unable to identify the morphs in the D-MAD environment.
	}\DIFaddend \end{itemize}

	\subsubsection{\Rev{Results for Experiment-II - S-MAD}}
	\begin{itemize}
		\item \DIFdelbegin \DIFdel{\Rev{In contrast to D-MAD, the accuracy is significantly lower in the S-MAD setting for border guards.}
		}\DIFdelend \DIFaddbegin \DIFadd{In contrast to D-MAD, the accuracy is significantly lower in the S-MAD setting for border guards.
		}\DIFaddend \item \DIFdelbegin \DIFdel{\Rev{Similar to the previous set of analyses, the average accuracy drops across irrespective of lines of work in the S-MAD setting when no reference images are available.} 
	}\DIFdelend \DIFaddbegin \DIFadd{Similar to the last set of analyses, the average accuracy decreases in the S-MAD scenario when no reference photos are available, independent of the lines of work. 
	}\DIFaddend \end{itemize}

	\subsubsection{\Rev{Common Observations}}
	\DIFdelbegin \DIFdel{\Rev{Special attention should also be given to miscellaneous observers without a specific line of work performing equally well as face comparison experts in detecting morphs in the D-MAD settings, while the average performance drops in this group in S-MAD settings.}
	}\DIFdelend \DIFaddbegin \DIFadd{Special consideration should also be given to various observers outside of a specific field of expertise who match face comparison specialists in morph detection in the D-MAD settings, while the average performance drops in this group in S-MAD settings.
	}\DIFaddend 

	\begin{figure*}[h]
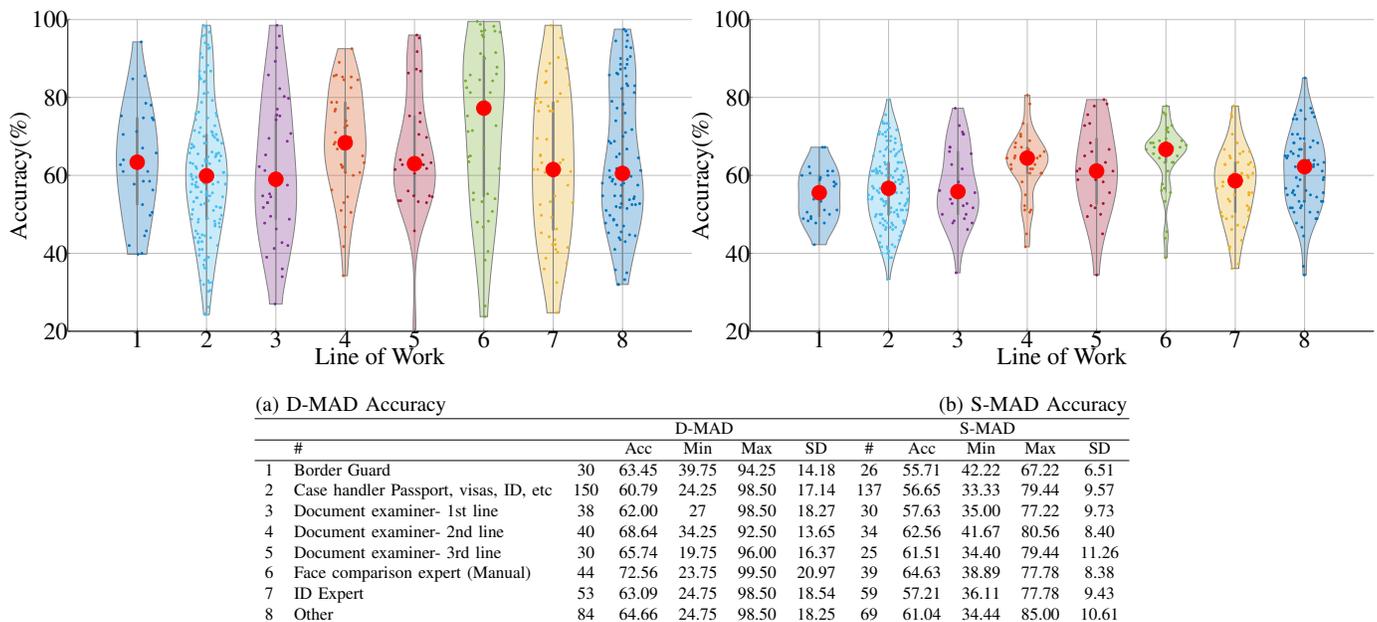

		\centering 
		\subfloat[D-MAD Accuracy]{%
			\includegraphics[trim=0 6cm 0 0, width=.5\linewidth]{figures-210915/d_mad_line_of_work_violin.tex}%
			\label{fig:d-mad-line-of-work}%
		}
		\subfloat[S-MAD Accuracy]{%
			\includegraphics[trim=0 6cm 0 0, width=.5\linewidth]{figures-210915/s_mad_line_of_work_violin.tex}%
			\label{fig:s-mad-line-of-work}%
		}\\
		\resizebox{0.65\textwidth}{!}{
			\begin{tabular}{llcccccccccc}
				\hline
				& & \multicolumn{5}{c}{D-MAD}  & \multicolumn{5}{c}{S-MAD} \\
				\cline{1-12}& \# &   & Acc & Min & Max & SD & \# & Acc & Min & Max & SD \\
				\hline
				1 & Border Guard 						  & 30  & 63.45  & 39.75 & 94.25 & 14.18 & 26  & 55.71 & 42.22 & 67.22 & 6.51\\
				2 & Case handler Passport, visas, ID, etc & 150 & 60.79  & 24.25 & 98.50 & 17.14 & 137 & 56.65 & 33.33 & 79.44 & 9.57\\
				3 & Document examiner- 1st line 		  & 38  & 62.00  & 27    & 98.50 & 18.27 & 30  & 57.63 & 35.00 & 77.22 & 9.73\\
				4 & Document examiner- 2nd line 		  & 40  & 68.64  & 34.25 & 92.50 & 13.65 & 34  & 62.56 & 41.67 & 80.56 & 8.40\\
				5 & Document examiner- 3rd line 		  & 30  & 65.74  & 19.75 & 96.00 & 16.37 & 25  & 61.51 & 34.40 & 79.44 & 11.26\\
				6 & Face comparison expert (Manual) 	  & 44  & 72.56  & 23.75 & 99.50 & 20.97 & 39  & 64.63 & 38.89 & 77.78 & 8.38\\
				7 & ID Expert 							  & 53  & 63.09  & 24.75 & 98.50 & 18.54 & 59  & 57.21 & 36.11 & 77.78 & 9.43\\
				8 & Other  								  & 84  & 64.66  & 24.75 & 98.50 & 18.25 & 69  & 61.04 & 34.44 & 85.00 & 10.61\\
				\hline
			\end{tabular}
		}
	\caption{Impact of line of work on MAD accuracy of human observers.}
	\label{fig:mad-vs-line-of-work}
	\end{figure*}

	\DIFdelbegin 
\DIFdel{Motivated by subtle differences, we }\DIFdelend \DIFaddbegin \DIFadd{We }\DIFaddend further conduct a statistical analysis \DIFdelbegin \DIFdel{using different line of work anddue to different }\DIFdelend \DIFaddbegin \DIFadd{utilizing a different field of study and, due to the variable }\DIFaddend number of observers, we \DIFdelbegin \DIFdel{use }\DIFdelend \DIFaddbegin \DIFadd{utilize }\DIFaddend ranks in one-criterion variance analysis \cite{kruskal1952use} \DIFaddbegin \DIFadd{since we are motivated by small distinctions}\DIFaddend . With the given number of observers in each different line of work corresponding to Figure \DIFaddbegin \DIFadd{Figure}\DIFaddend ~\ref{fig:d-mad-line-of-work} (n = 30, 150, 38, 40, 30, 44, 53), we note the \DIFdelbegin \DIFdel{\Rev{\textit{H}(6)=18.99, \textit{p}=.004} }\DIFdelend \DIFaddbegin \DIFadd{\textit{H}(6)=18.99, \textit{p}=.004 }\DIFaddend indicating significance between the observers in \DIFaddbegin \DIFadd{a }\DIFaddend different line of work. However, \DIFdelbegin \DIFdel{compared }\DIFdelend \DIFaddbegin \DIFadd{in contrast }\DIFaddend to observers in \DIFdelbegin \DIFdel{different line of work against face comparison experts }\DIFdelend \DIFaddbegin \DIFadd{other fields, experts in face comparison }\DIFaddend (Manual examination), we note \DIFdelbegin \DIFdel{a \Rev{\textit{H}(1)=10.53, \textit{p}=.001} suggesting }\DIFdelend \DIFaddbegin \DIFadd{an \textit{H}(1)=10.53, \textit{p}=.001 suggesting the }\DIFaddend better performance of face comparison experts.
	\DIFdelbegin 
\DIFdelend

	\subsection{MAD for Digital v/s Print-Scan Images}
\begin{figure*}[h]
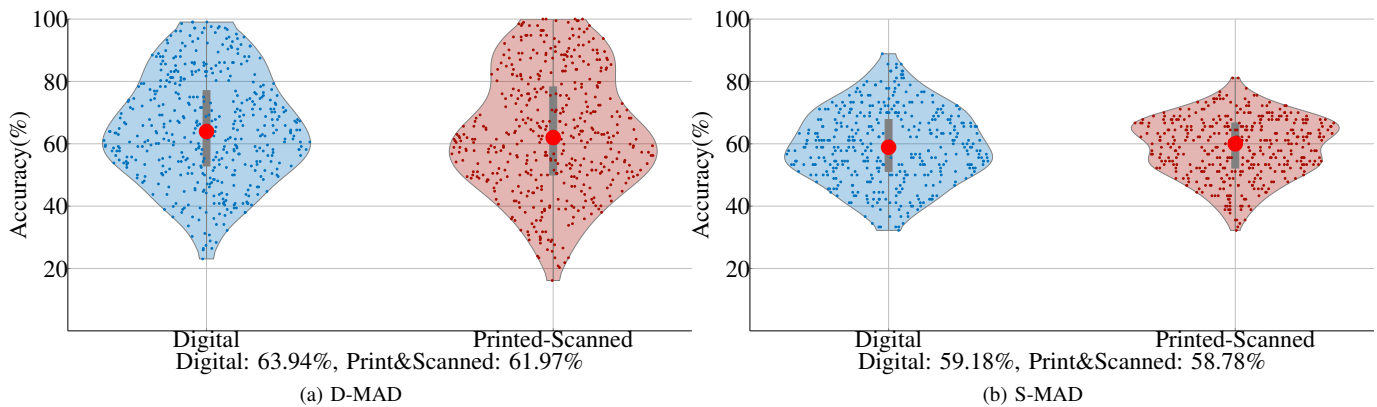

		\centering 
		\subfloat[D-MAD]{%
			\includegraphics[trim=0 0 0 0, width=.5\linewidth]{figures-210915/d_mad_PrintVsDigital_average_violinplot.tex}%
			\label{fig:d-mad-print-vs-digital}%
		}
		\subfloat[S-MAD]{%
			\includegraphics[trim=0 0 0 0, width=.5\linewidth]{figures-210915/s_mad_PrintVsDigital_average_violinplot.tex}%
			\label{fig:s-mad-print-vs-digital}%
		}
		\caption{Impact of digital v/s print-scan morphed images on MAD accuracy of human observers.}
		\label{fig:mad-print-vs-digital}%
	\end{figure*}

	\DIFdelbegin \DIFdel{In addition to the analysis provided above, we note the severity }\DIFdelend \DIFaddbegin \DIFadd{We note the difficulty }\DIFaddend of the MAD \DIFdelbegin \DIFdel{task }\DIFdelend \DIFaddbegin \DIFadd{job }\DIFaddend for digital and printed-scanned \DIFdelbegin \DIFdel{images, which correspond to }\DIFdelend \DIFaddbegin \DIFadd{photos, which relate to a }\DIFaddend real-life \DIFdelbegin \DIFdel{scenarios}\DIFdelend \DIFaddbegin \DIFadd{setting, in addition to the analysis that was previously provided}\DIFaddend . While the former is common in applying for visas by uploading digital images, the latter can also be observed in different countries where the printed-scanned images can be uploaded to obtain valid ID documents. \DIFdelbegin \DIFdel{In this set of analyses , we combine all different categories of observers and analyse the accuracy for }\DIFdelend \DIFaddbegin \DIFadd{This series of analyses compare the accuracy of }\DIFaddend digital versus printed-scanned morphed images\DIFdelbegin \DIFdel{as presented }\DIFdelend \DIFaddbegin \DIFadd{, as seen }\DIFaddend in Figure~\ref{fig:mad-print-vs-digital}. 

	\subsubsection{\Rev{Results for Experiment-I - D-MAD}}
	\DIFdelbegin \DIFdel{\Rev{It can be noted that the observers spot the morphed images in the digital domain with an average accuracy of \Rev{64.22\% (\textit{min}=42.05\%, \textit{max}=95.73\%, \textit{SD}=10.35\%, \textit{Mdn}=63.05\%)} in D-MAD settings. In contrast, the accuracy drops slightly for printed-scanned images to 63.39\% \Rev{(\textit{min}=38.80\%, \textit{max}=94.40\%, \textit{SD}=10.55\%, \textit{Mdn}=64.41\%)}. {\color{black} To further understand if the detection accuracy is statistically different in detecting digital and print-scan attacks, we conduct Levene’s test \cite{levene1961robust} by performing ANOVA on the absolute deviations of the data values from their group means and we note a \Rev{\textit{t}(1.93)=7.26, \textit{p}=.007}. The low p-value obtained clearly indicates no significant difference in observers in detecting either of the category.}}
	}\DIFdelend \DIFaddbegin \DIFadd{It can be noted that the observers spot the morphed images in the digital domain with an average accuracy of 64.22\% (\textit{min}=42.05\%, \textit{max}=95.73\%, \textit{SD}=10.35\%, \textit{Mdn}=63.05\%) in D-MAD settings. In contrast, the accuracy drops slightly for printed-scanned images to 63.39\% (\textit{min}=38.80\%, \textit{max}=94.40\%, \textit{SD}=10.55\%, \textit{Mdn}=64.41\%). To further comprehend whether the detection accuracy for digital and print-scan attacks differs statistically, we conduct Levene’s test \cite{levene1961robust} by performing ANOVA on the absolute deviations of the data values from their group means and we note a \textit{t}(1.93)=7.26, \textit{p}=.007. The analysis indicates no discernible difference between observers in identifying either category (\textit{p}>.001).
	}\DIFaddend 

	\subsubsection{\Rev{Results for Experiment-II - S-MAD}}
	\DIFdelbegin \DIFdel{\Rev{However, the average accuracy drops to 59.18\% (\textit{min}=32.22\%, \textit{max}=88.88\%, \textit{SD}=11.70\%, \textit{Mdn}=58.88\%) and 58.78\% (\textit{min}=32.22\%, \textit{max}=81.11\%, \textit{SD}=9.05\%, \textit{Mdn}=60\%) for digital and printed-scanned images, respectively, in S-MAD settings. It has to be noted that none of the observers crosses an accuracy of 90\% in the S-MAD setting, reinforcing the challenge of detecting the morphed images when no reference image is unavailable. The lower quality of images in S-MAD can be a factor for lower accuracy. A further analysis if needed to understand the drop in accuracy for both S-MAD and digital v/s printed-scanned images. }
	}\DIFdelend \DIFaddbegin \DIFadd{However, the average accuracy drops to 59.18\% (\textit{min}=32.22\%, \textit{max}=88.88\%, \textit{SD}=11.70\%, \textit{Mdn}=58.88\%) and 58.78\% (\textit{min}=32.22\%, \textit{max}=81.11\%, \textit{SD}=9.05\%, \textit{Mdn}=60\%) for digital and printed-scanned images, respectively, in S-MAD settings. It should be observed that no observer exceeds an accuracy of 90\% in the S-MAD setting, highlighting the difficulty in identifying altered images in the absence of a reference image. The lower quality of images in S-MAD can be a factor for lower accuracy. Further analysis is needed to understand the drop in accuracy for both S-MAD and digital v/s printed-scanned images.
	}\DIFaddend

	\subsection{MAD v/s Demographics of Observers}
	\DIFdelbegin \DIFdel{We further analyse }\DIFdelend \DIFaddbegin \DIFadd{To see whether these traits contribute to improved MAD accuracy, we also examine }\DIFaddend the demographics of the observers\DIFdelbegin \DIFdel{to understand if such factors play a role in obtaining better MAD accuracy}\DIFdelend . Specifically, we look at age, gender\DIFaddbegin \DIFadd{, }\DIFaddend and country of work (not ethnicity) for S-MAD and D-MAD. Figure ~\ref{fig:mad-age-vs-accuracy} presents the correlation of MAD versus \DIFaddbegin \DIFadd{the }\DIFaddend age of the observers. 

	\subsubsection{\Rev{Results for Experiment-I - D-MAD}}
	\DIFdelbegin \DIFdel{\Rev{As noted from Figure~\ref{fig:d-mad-age-vs-accuracy}, there appears to be no strong correlation between the age of observers and MAD accuracy.}
	}\DIFdelend \DIFaddbegin \DIFadd{\Rev{As noted in Figure~\ref{fig:d-mad-age-vs-accuracy} and Figure~\ref{fig:s-mad-age-vs-accuracy}, there appears to be no strong correlation between the age of observers and MAD accuracy.}
	}\DIFaddend 

	\subsubsection{\Rev{Results for Experiment-II - S-MAD}}
\begin{figure*}[h]
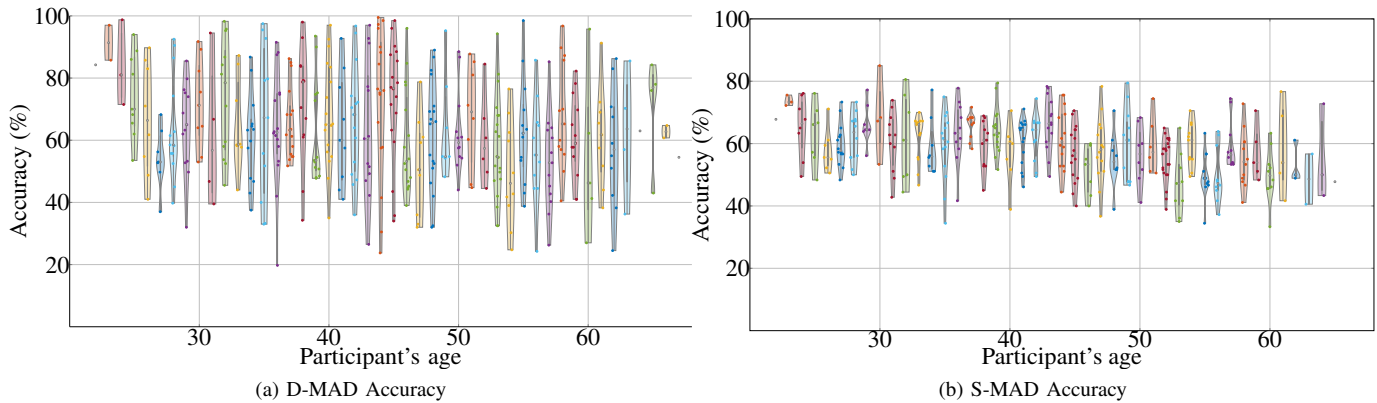

		\centering 
		\subfloat[D-MAD Accuracy]{%
			\includegraphics[trim=0 0 0 0, width=.5\linewidth]{figures-210915/d_mad_age_vs_acc_violin.tex}%
			\label{fig:d-mad-age-vs-accuracy}%
		}
		\subfloat[S-MAD Accuracy]{%
			\includegraphics[trim=0 0 0 0, width=.5\linewidth]{figures-210915/s_mad_age_vs_acc_violin.tex}%
			\label{fig:s-mad-age-vs-accuracy}%
		}
		\caption{MAD accuracy versus age of human observers.}
		\label{fig:mad-age-vs-accuracy}%
	\end{figure*}

	\DIFdelbegin \DIFdel{\Rev{However, the general trend of drop in MAD accuracy in S-MAD settings compared to D-MAD settings can be easily observed. The absence of an image pair with a trusted live capture as a reference makes it a challenge to determine if a suspected image is morphed or not.}
	}\DIFdelend \DIFaddbegin \DIFadd{It is simple to see the general trend of decreased MAD accuracy in S-MAD settings compared to D-MAD settings by comparing Figure~\ref{fig:s-mad-age-vs-accuracy} to Figure~\ref{fig:d-mad-age-vs-accuracy}. The absence of an image pair with a trusted live capture as a reference makes it a challenge to determine if a suspected image is morphed or not.
	}\DIFaddend 

	\DIFdelbegin \DIFdel{While we also analyse the detection accuracy with respect to nationality, we do not derive any conclusive results, as most }\DIFdelend \DIFaddbegin \DIFadd{Since the majority }\DIFaddend of the observers are from European \DIFdelbegin \DIFdel{countries}\DIFdelend \DIFaddbegin \DIFadd{nations, we do not come to any firm conclusions even if we also analyze the detection accuracy in terms of nationality}\DIFaddend . Also, no significant differences were observed \DIFdelbegin \DIFdel{with respect to }\DIFdelend \DIFaddbegin \DIFadd{concerning }\DIFaddend gender, as noted in Figure~\ref{fig:gender-vs-mad} in Appendix.

	\subsection{Improvement of detection accuracy over the number of images}
	\label{sssec:progression-over-images}
	\DIFdelbegin \DIFdel{Noting }\DIFdelend \DIFaddbegin \DIFadd{We examine whether observers become more adept at seeing morphs after viewing a particular number of photos. Taking into account }\DIFaddend the low accuracy of the \DIFdelbegin \DIFdel{different categories }\DIFdelend \DIFaddbegin \DIFadd{various groups }\DIFaddend of observers in \DIFdelbegin \DIFdel{detecting }\DIFdelend \DIFaddbegin \DIFadd{identifying }\DIFaddend morphing attacks in both D-MAD and S-MAD \DIFdelbegin \DIFdel{settings, }\DIFdelend \DIFaddbegin \DIFadd{situations }\DIFaddend we also analyze if the observers get better at detecting morphs after looking at a certain number of images. We thus analyze the observer accuracy by dividing the whole set of images into 4 blocks in the D-MAD setting, where each consists of 100 image pairs\DIFaddbegin \DIFadd{, }\DIFaddend and the 2 blocks in the S-MAD setting, where each set consists of 92 images.

	\subsubsection{\Rev{Results for Experiment-I - D-MAD}}
	\DIFdelbegin \DIFdel{\Rev{As noted in the Figure~\ref{fig:mad-accuracy-vs-set}, average accuracy of each observer group increases for every 100 images completed in the case of D-MAD (Figure~\ref{fig:d-mad-accuracy-vs-set}). Although this is a preliminary indication, it can be easily deduced that the observers become better at detecting the morphed images by looking at a considerable amount of images. A positive interpretation of this observation is that a dedicated training program in examination-based morphing attack detection would generally efficiently increase the competence in detecting morphs. }
	}\DIFdelend \DIFaddbegin \DIFadd{As noted in the Figure~\ref{fig:mad-accuracy-vs-set}, the average accuracy of each observer group increases for every 100 images completed in the case of D-MAD (Figure~\ref{fig:d-mad-accuracy-vs-set}). Although this is only a preliminary finding, it is clear that by viewing a large number of photographs, observers improve their ability to recognize altered images. A positive interpretation of this observation is that a dedicated training program in examination-based MAD would generally efficiently increase the competence in detecting morphs.
	}\DIFaddend 

	\DIFdelbegin \DIFdel{\Rev{We further validate the observation by conducting positive trend analysis using Cox and Stuart Test \cite{conover1999practical}. Specifically, by measuring the accuracy of observers for every 100 pairs of images, assumed to be independent observations, we are interested to know if there is, in fact, a time-dependent trend (i.e., the observations are, in fact, not independent and observers become better at the task). As noted from the Table~\ref{tab:cox-stuart-test}, we clearly see a significance level of the test on a threshold of probability value=0.05 (i.e. $\alpha=0.05$). As the obtained p-value of the tests for each set of 100 pairs is lesser than $\alpha=0.05$, we conclude with sufficient evidence that seeing a larger amount of D-MAD pairs improves the accuracy of observer. This observation further supports the need for training where the observers can see a number of new examples to improve their detection skills. }
	}\DIFdelend \DIFaddbegin \DIFadd{We further validate the observation by conducting a positive trend analysis using Cox and Stuart Test \mbox{
\cite{conover1999practical}}\hspace{0pt}
. We are specifically interested in determining whether there is, in fact, a time-dependent trend by assessing the observers’ accuracy of observers for every 100 pairs of photos, which are presumed to be independent observations (i.e., the observations are, in fact, not independent and observers become better at the task). As noted in Table~\ref{tab:cox-stuart-test}, we see a significant level of the test (\textit{p}<.001). With obtained p-value of the tests for each set of 100 pairs, we conclude with sufficient evidence that seeing a larger amount of D-MAD pairs improves the accuracy of the observer. This discovery strengthens the argument that observers should receive training where they can see a variety of new cases to hone their detection abilities.
	}\DIFaddend 

	\begin{table}[htp]
		\begin{tabular}{l|ccc}
			\hline
			\hline
			Increasing Trend & False & True  &  True \\
			p-value & 0.642 & 	<0.001	& <0.001\\
			\hline
			\hline
		\end{tabular}
		\caption{Cox and Stuart Trend Test for measuring change in accuracy of observers against the number of D-MAD pairs seen.}
		\label{tab:cox-stuart-test}
	\end{table}

	\DIFdelbegin \DIFdel{\Rev{We further complement our test with the hypothesis that the detection accuracy increases by seeing number of examples. For this aspect, we draw observers randomly from total observer populations and conduct a One-way analysis of variance \cite{anovatest}.  We note average accuracy across different group with 57.54\%, 62.95\%, 66.44\% and 69.47\% with \Rev{\textit{p}<.0001} supporting our observation that the accuracy increases with examples seen.}
	}\DIFdelend \DIFaddbegin \DIFadd{We also add the hypothesis that seeing more examples will improve the detection accuracy of our test. For this aspect, we draw observers randomly from total observer populations and conduct a One-way analysis of variance }[\DIFadd{44}]\DIFadd{. We note the average accuracy for each category with 57.54\%, 62.95\%, 66.44\% and 69.47\% with \textit{p} < .0001 supporting our observation that the accuracy increases with examples seen.
	}\DIFaddend 

	\subsubsection{\Rev{Results for Experiment-II - S-MAD}}
\begin{figure*}[h]
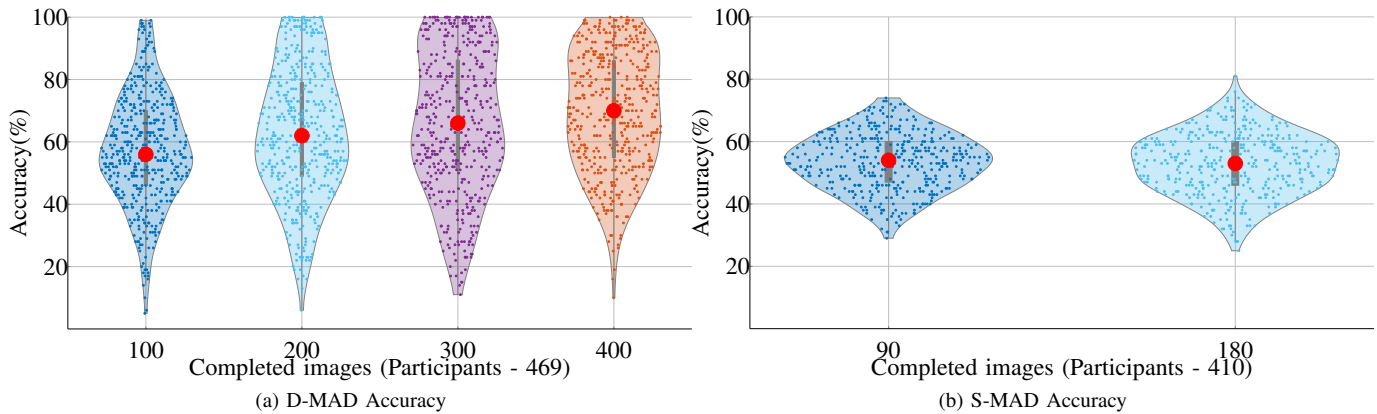

		\centering 
		\subfloat[D-MAD Accuracy]{%
			\includegraphics[trim=0 0 0 0, width=.5\linewidth]{figures-210915/d_mad_100by100_violinplot.tex}%
			\label{fig:d-mad-accuracy-vs-set}%
		}
		\subfloat[S-MAD Accuracy]{%
			\includegraphics[trim=0 0 0 0, width=.5\linewidth]{figures-210915/s_mad_90_by_90_violinplot.tex}%
			\label{fig:s-mad-accuracy-vs-set}%
		}
		\caption{Improvement in MAD accuracy over the total number of images seen.}
		\label{fig:mad-accuracy-vs-set}%
	\end{figure*}

	\DIFdelbegin \DIFdel{\Rev{While a similar observation cannot be drawn from S-MAD settings as shown in Figure~\ref{fig:s-mad-accuracy-vs-set}, we can only hypothesize the complexity of the problem and devise better training strategies in detecting morphs based on single images. We further assert that better competence through training programmes on an understanding of image quality help in future works.}
	}\DIFdelend \DIFaddbegin \DIFadd{While a similar conclusion cannot be derived from S-MAD settings depicted in Figure~\ref{fig:s-mad-accuracy-vs-set}, we can only speculate on the difficulty of the issue and come up with better training methods for morph detection based on single photos. We further assert that better competence through training programs on an understanding of image quality help in future works.
	}\DIFaddend 

	\subsection{Bona fide versus morphed image detection in ABC Gate Settings in D-MAD}
\begin{figure*}[h]
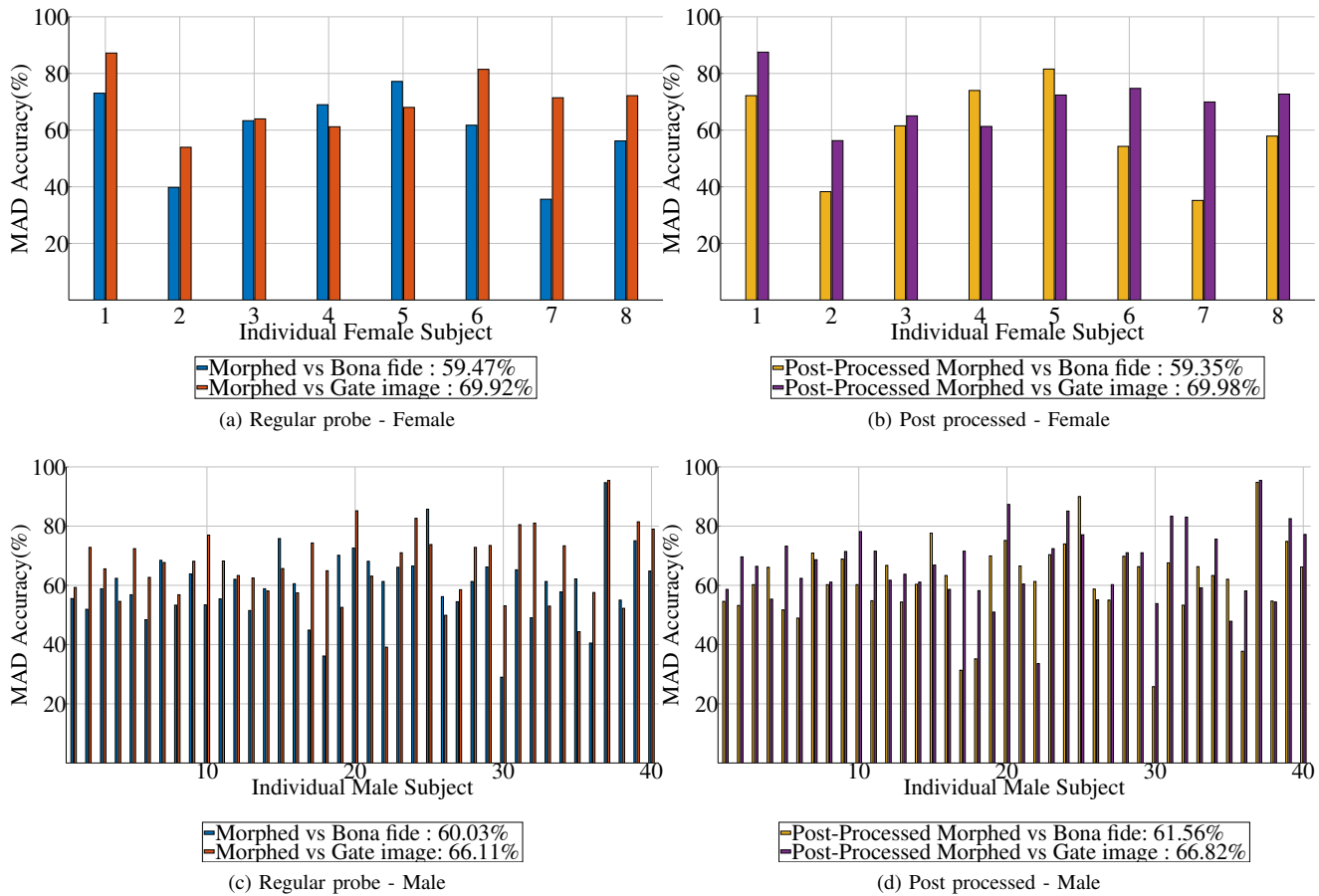

		\centering 
		\subfloat[Regular probe - Female]{%
			\includegraphics[trim=0 0 0 0, width=.48\linewidth]{figures-210915/d_mad_b_m_p_morphed_Female_subject_acc-digital.tex}%
			\label{fig:female-d-mad-probe-digital}%
		}
		\subfloat[Post processed - Female]{%
			\includegraphics[trim=0 0 0 0, width=.48\linewidth]{figures-210915/d_mad_b_m_p_morphed_Female_subject_acc-postprocessed.tex}%
			\label{fig:female-d-mad-probe-postprocessed}%
		}\\
		\subfloat[Regular probe - Male]{%
			\includegraphics[trim=0 0 0 0, width=.48\linewidth]{figures-210915/d_mad_b_m_p_morphed_Male_subject_acc-digital.tex}%
			\label{fig:male-d-mad-probe-digital}%
		}
		\subfloat[Post processed - Male]{%
			\includegraphics[trim=0 0 0 0, width=.48\linewidth]{figures-210915/d_mad_b_m_p_morphed_Male_subject_acc-postprocessed.tex}%
			\label{fig:male-d-mad-probe-postprocessed}%
		}\\
		\caption{MAD accuracy in D-MAD setting with regular probe images v/s ABC Gate probe images for morphed images of male and female subjects}
		\label{fig:male-female-d-mad-digital-vs-postprocessed}%
	\end{figure*}

	In this \DIFdelbegin \DIFdel{section, we analyse the role }\DIFdelend \DIFaddbegin \DIFadd{part, we examine the function }\DIFaddend of ABC gate images\DIFdelbegin \DIFdel{which are usually available }\DIFdelend \DIFaddbegin \DIFadd{ and morphed images shown alongside as in a }\DIFaddend border-crossing scenario to detect morphs\DIFdelbegin \DIFdel{, i. e., morphed images shown together with ABC gates and bona fide images shown together with ABC gate images. As noted from }\DIFdelend \DIFaddbegin \DIFadd{. As noted in }\DIFaddend Figure~\ref{fig:male-female-d-mad-digital-vs-postprocessed}, the accuracy of the \DIFdelbegin \DIFdel{morphing attack detection }\DIFdelend \DIFaddbegin \DIFadd{MAD }\DIFaddend increases when the ABC gate images are provided as a reference image. We specifically note an increase of 10\% in detection accuracy for female subjects while the accuracy improvement for morphed images of male faces results in around 5\% average accuracy. 
\DIFdelbegin 
\DIFdel{To further establish if the ABC gates help in detecting morphed images to a better extent, we conduct Kruskal-Wallis test across two groups of detection , i.e., }\DIFdelend \DIFaddbegin \DIFadd{We conduct a Kruskal–Wallis test between two detection groups }\DIFaddend when the ABC gate images are available versus when not-available \DIFaddbegin \DIFadd{to see whether the ABC gates help in detecting morphed images to a greater extent }\DIFaddend (n = 80, 80 correspondingly). We note a \DIFdelbegin \DIFdel{\Rev{\textit{H}(1)=7.55, \textit{p}=.006} }\DIFdelend \DIFaddbegin \DIFadd{\textit{H}(1)=7.55, \textit{p}=.006 }\DIFaddend for male face images and (\textit{H}(1)=2.81, \textit{p}=.093) for female face images. We conclude that there is no real advantage for spotting morphed images when ABC gate images are provided. Our hypothesis here is that the face images of female subjects vary significantly due to varying hairstyles and everyday make-up making the observers not detect all the morphed images. However\DIFdelbegin \DIFdel{this needs further investigations by extending }\DIFdelend \DIFaddbegin \DIFadd{, this requires additional research, which will need enlarging }\DIFaddend the dataset with and without make-up\DIFdelbegin \DIFdel{for female subjects. While }\DIFdelend \DIFaddbegin \DIFadd{. While it is noted that }\DIFaddend the average p-value \DIFdelbegin \DIFdel{across }\DIFdelend \DIFaddbegin \DIFadd{for }\DIFaddend male and female face \DIFdelbegin \DIFdel{images together is observed to be H(1)=6.36, p=.011 suggesting }\DIFdelend \DIFaddbegin \DIFadd{images is \textit{H}(1)=6.36, \textit{p}=.011 suggests }\DIFaddend the need for \DIFdelbegin \DIFdel{better trusted }\DIFdelend \DIFaddbegin \DIFadd{more reliable }\DIFaddend live capture to \DIFdelbegin \DIFdel{make the detection better}\DIFdelend \DIFaddbegin \DIFadd{improve the detection}\DIFaddend . We further hypothesize that \DIFdelbegin \DIFdel{high quality }\DIFdelend \DIFaddbegin \DIFadd{high-quality }\DIFaddend live images from ABC gates could generally provide more details for border guards, examiners, etc., making comparisons and MAD easier. 

	\section{MAD Algorithms v/s Observer Accuracy}
	\label{sec:mad-algorithms-vs-observers}
	We further analyze the performance of human observers against two algorithms in both categories of D-MAD and S-MAD as explained in the section below.

	\subsubsection{\Rev{Results for Experiment-I - D-MAD}}
	\DIFdelbegin \DIFdel{To compare the }\DIFdelend \DIFaddbegin \DIFadd{We select DeepFeature Difference (DFD)  \cite{scherhag2020deep} and Local Binary Pattern - Support Vector Machine (LBP-SVM) \cite{scherhag2019detection} to compare the automated D-MAD algorithms against average human observer accuracy as both the algorithms were evaluated under NIST FRVT MORPH challenge \cite{NistFrvtMorph}. }\DIFaddend  While DFD \cite{scherhag2020deep} obtained best performance in detecting digital images, LBP-SVM \cite{scherhag2019detection} obtained reasonable accuracy in NIST FRVT MORPH challenge \cite{NistFrvtMorph}. As noted from the results in Table~\ref{tab:d-mad-accuracy-versus-observers-s-mad}, the DFD \cite{scherhag2020deep} obtains very high accuracy,  LBP-SVM \cite{scherhag2019detection} obtains random accuracy of 50\%. The human observer accuracy in best case is  64.61\%, \DIFdelbegin \DIFdel{lower than DFD \mbox{
\cite{scherhag2020deep} }\hspace{0pt}
but }\DIFdelend \DIFaddbegin \DIFadd{which is }\DIFaddend better than the weakest algorithm \DIFaddbegin \DIFadd{but less accurate than DFD \mbox{
\cite{scherhag2020deep}}\hspace{0pt}
}\DIFaddend . The high accuracy of DFD can be attributed to extracting \DIFdelbegin \DIFdel{identity specific }\DIFdelend \DIFaddbegin \DIFadd{identity-specific }\DIFaddend features for detecting morphs. \DIFdelbegin \DIFdel{\Rev{We also provide the violin plots for APCER and BPCER in Figure~\ref{fig:apcer-bpcer-d-mad} to illustrate the wrong classifications in bona fide and morph attacks.}
	}\DIFdelend \DIFaddbegin \DIFadd{To demonstrate incorrect classifications in bona fide and morph attacks, we also include the violin plots for APCER and BPCER in  Figure~\ref{fig:apcer-bpcer-d-mad}.
	}\DIFaddend

\begin{figure*}[h]
	\centering 
	\subfloat[APCER for D-MAD experiments with regular probe images]{%
		\includegraphics[trim=0 0 0 0, width=.48\linewidth]{figures-210915/revision-figures/d_mad_apcer_revision.tex}%
		\label{fig:d-mad-apcer-graphs}%
	}
	\subfloat[APCER for D-MAD experiments with probe images from ABC Gates]{%
		\includegraphics[trim=0 0 0 0, width=.48\linewidth]{figures-210915/revision-figures/d_mad_apcer_abcgate_revision.tex}%
		\label{fig:d-mad-abcgates-apcer-graphs}%
	}\\
	\subfloat[BPCER for D-MAD experiments]{%
		\includegraphics[trim=0 0 0 0, width=.48\linewidth]{figures-210915/revision-figures/d_mad_bpcer_revision.tex}%
		\label{fig:d-mad-bpcer-graphs}%
	}
	\caption{APCER and BPCER for D-MAD}
	\label{fig:apcer-bpcer-d-mad}%
\end{figure*}

	\begin{table*}[htp]
	\color{black}
	\centering
	\resizebox{0.65\textwidth}{!}{
		\begin{tabular}{llccc}
			\hline
			Data & Approach &  Accuracy  & APCER & BPCER \\
			\hline
			\hline
			\multirow{3}{*}{Digital} & DeepFeature Difference \cite{scherhag2020deep} & 97.91  & 0 & 4.16\\
			& LBP-SVM \cite{scherhag2019detection} &  50.00  & 0 & 100\\
			& Average Observer Accuracy & 63.80  & 39.08 & 36.55 \\
			\hline
			\multirow{3}{*}{Print-Scan} & DeepFeature Difference \cite{scherhag2020deep} & 97.91  & 0 & 4.16\\
			& LBP-SVM \cite{scherhag2019detection} &  50.00 & 0 & 100\\
			& Average Observer Accuracy &  62.76 & 37.11 & - \\
			\hline
			\hline
			\multirow{3}{*}{Post-Processed} & DeepFeature Difference \cite{scherhag2020deep} & 97.91 & 0 & 4.16\\
			& LBP-SVM \cite{scherhag2019detection} &  50.00  & 0 & 100\\
			- Digital & Average Observer Accuracy &  64.61  & 35.35 & - \\
			\hline
			\hline
			\multirow{3}{*}{Post-Processed} & DeepFeature Difference \cite{scherhag2020deep} & 96.87  & 0 & 6.25 \\
			& LBP-SVM \cite{scherhag2019detection} &  50.00  & 0 & 100\\
			- Print-scan & Average Observer Accuracy &  63.90 & 35.37 & - \\
			\hline
		\end{tabular}
	}
	\caption{MAD accuracy obtained for two state-of-art algorithms compared with human observer average accuracy in D-MAD settings. *Note - LBP-SVM \cite{scherhag2019detection} provides random performance of 50\% due to nature of algorithm not able to scale up to the images in newly constructed HOMID dataset. BPCER is not reported in cases where there were not enough samples used for human observer analysis.}
	\label{tab:d-mad-accuracy-versus-observers-s-mad}
\end{table*}


		\subsubsection{\Rev{Results for Experiment-II - S-MAD}}
	We \DIFdelbegin \DIFdel{benchmark the average }\DIFdelend \DIFaddbegin \DIFadd{compare the typical }\DIFaddend human observer accuracy \DIFdelbegin \DIFdel{against two state-of-the-art approaches for }\DIFdelend \DIFaddbegin \DIFadd{to two cutting-edge }\DIFaddend S-MAD \DIFaddbegin \DIFadd{methods}\DIFaddend . We choose Hybrid features \cite{ramachandra2019towards} and Ensemble features \cite{venkatesh2020single} for detecting morphing attacks based on the performance obtained in \DIFaddbegin \DIFadd{the }\DIFaddend NIST FRVT MORPH challenge \cite{NistFrvtMorph} with the best performance in detecting \DIFdelbegin \DIFdel{printed and scanned }\DIFdelend \DIFaddbegin \DIFadd{printed-and-scanned }\DIFaddend morph images. \DIFdelbegin \DIFdel{While }\DIFdelend \DIFaddbegin \DIFadd{The Ensemble features \mbox{
\cite{venkatesh2020single} }\hspace{0pt}
combine textural features with a collection of classifiers, in contrast to }\DIFaddend the Hybrid features \cite{ramachandra2019towards} \DIFdelbegin \DIFdel{uses both scale spaceand color spacecombined with multiple classifiers, the Ensemble features \mbox{
\cite{venkatesh2020single} }\hspace{0pt}
uses textural features in conjunction with a set of }\DIFdelend \DIFaddbegin \DIFadd{which mix scale space, color space, and numerous }\DIFaddend classifiers. We present the accuracy in Table~\ref{tab:s-mad-accuracy-versus-observers-s-mad}\DIFaddbegin \DIFadd{. }\DIFaddend for two chosen algorithms and average human observer accuracy. We note that both algorithms perform better than the human observers in S-MAD for both chosen \DIFdelbegin \DIFdel{algorithms }\DIFdelend \DIFaddbegin \DIFadd{algorithms }\DIFaddend indicating the challenging nature of detecting morphs when no reference image is available. \DIFdelbegin \DIFdel{\Rev{Further, violin plots for APCER and BPCER are provided in Figure~\ref{fig:apcer-bpcer-s-mad} to illustrate the wrong classifications in bona fide and morph attacks in S-MAD setting.}
	}\DIFdelend \DIFaddbegin \DIFadd{To further demonstrate incorrect classifications in genuine and morph attacks in an S-MAD environment, violin plots for APCER and BPCER are in Figure~\ref{fig:apcer-bpcer-s-mad}.
	}\DIFaddend

	\begin{table*}[htp]
	\centering
	\color{black}
	\resizebox{0.65\textwidth}{!}{
		\begin{tabular}{llccc}
			\hline
			Data & Approach &  Accuracy & APCER & BPCER \\
			\hline
			\hline
			\multirow{3}{*}{Digital} & Ensemble Features \cite{venkatesh2020single} & 70.00  & 30.00 & 30.00 \\
			& Hybrid Features \cite{ramachandra2019towards} & 73.34  & 3.33 & 76.66 \\
			& Average Observer Accuracy & 58.17   & 38.40 & 45.31\\
			\hline
			\multirow{3}{*}{Print-Scan} & Ensemble Features \cite{venkatesh2020single} & 63.34  & 36.66 & 36.66 \\
			& Hybrid Features \cite{ramachandra2019towards} & 56.67  & 3.33 & 80.00 \\
			& Average Observer Accuracy &  53.88  & 32.23  & 60.02\\
			\hline
			\hline
			\multirow{3}{*}{Post-Processed} & Ensemble Features \cite{venkatesh2020single} & 70.00  & 26.66 & 33.33 \\
			& Hybrid Features \cite{ramachandra2019towards} & 66.67  & 6.66 & 76.66 \\
			- Digital & Average Observer Accuracy &  57.96 & 38.78 & - \\
			\hline
			\hline
			\multirow{3}{*}{Post-Processed} & Ensemble Features \cite{venkatesh2020single} & 60.00 & 30.00 & 40.00 \\
			& Hybrid Features \cite{ramachandra2019towards} & 56.67  & 3.33 & 76.66 \\ 
			- Print-scan  & Average Observer Accuracy &  54.23 & 31.51 & - \\
			\hline
		\end{tabular}
	}
	\caption{MAD accuracy obtained for two state-of-art algorithms compared with human observer average accuracy in S-MAD settings. *Note - BPCER is not reported in cases where there were not enough samples used for human observer analysis.}
	\label{tab:s-mad-accuracy-versus-observers-s-mad}
\end{table*}

	\begin{figure*}[h]
		\centering 
		\subfloat[APCER for S-MAD experiments]{%
				\includegraphics[trim=0 0 0 0, width=.48\linewidth]{figures-210915/revision-figures/s_mad_apcer_revision-2.tex}%
				\label{fig:s-mad-apcer-graphs}
			}
		\subfloat[BPCER for S-MAD experiments]{%
				\includegraphics[trim=0 0 0 0, width=.48\linewidth]{figures-210915/revision-figures/s_mad_bpcer_revision.tex}%
				\label{fig:s-mad-bpcer-graphs}
			}
		\caption{APCER and BPCER for S-MAD}
		\label{fig:apcer-bpcer-s-mad}%
	\end{figure*}


	\section{Summary of key findings and their implications}
	Based on the analysis presented in previous sections, we note a few major observations in this section and discuss their implications:
	\begin{itemize}
		\item \textbf{Availability of reference image:} \DIFdelbegin \DIFdel{On a general note, the detection accuracy in }\DIFdelend \DIFaddbegin \DIFadd{Generally speaking, }\DIFaddend S-MAD settings \DIFdelbegin \DIFdel{is lower than the accuracy in }\DIFdelend \DIFaddbegin \DIFadd{have a lower detection accuracy than }\DIFaddend D-MAD settings. It can be deduced that the observers make a better decision when a reference image is available.
		\item \textbf{Impact of prior training:} \DIFdelbegin \DIFdel{Observers }\DIFdelend \DIFaddbegin \DIFadd{When examining the average accuracy, both in S-MAD and D-MAD scenarios, observers }\DIFaddend with prior training in face comparison tend to \DIFdelbegin \DIFdel{perform }\DIFdelend \DIFaddbegin \DIFadd{do }\DIFaddend better than observers \DIFdelbegin \DIFdel{trained }\DIFdelend \DIFaddbegin \DIFadd{with prior experience }\DIFaddend in document examination \DIFdelbegin \DIFdel{when looking at the average accuracy - in both S-MAD and D-MAD settings  (}\DIFdelend  Fig~\ref{fig:mad-vs-training} and Fig~\ref{fig:mad-vs-training-document}). An implication of this analysis directly suggests that certain elements of MAD in facial comparison training may be beneficial to include.
		\item \textbf{Impact of participation:} Through \DIFdelbegin \DIFdel{the }\DIFdelend analysis, we \DIFdelbegin \DIFdel{note that the observers become better at detecting the morphed images just by participating }\DIFdelend \DIFaddbegin \DIFadd{find that just by taking part }\DIFaddend in the experiments, the observers become better at detecting the morphed images. Specifically, we note the improvement in accuracy throughout the experiment and after seeing \DIFdelbegin \DIFdel{a number of }\DIFdelend \DIFaddbegin \DIFadd{several }\DIFaddend examples (Figure~\ref{fig:mad-accuracy-vs-set}). This \DIFdelbegin \DIFdel{validates our previous observation that dedicated }\DIFdelend \DIFaddbegin \DIFadd{supports our earlier finding that specialized }\DIFaddend training programs and \DIFdelbegin \DIFdel{educating the observers with various examples will improve }\DIFdelend \DIFaddbegin \DIFadd{teaching observers with varied instances will increase }\DIFaddend detection accuracy. This observation is also in line with research done on the effectiveness of courses in facial comparisons, where the researchers suggest that examination-based training could be an important part of a training \DIFdelbegin \DIFdel{programme \mbox{
\cite{moreton2021international}}\hspace{0pt}
. However, further research is required to determine if this is also the case for MAD}\DIFdelend \DIFaddbegin \DIFadd{program \mbox{
\cite{moreton2021international}}\hspace{0pt}
. To find out if this also applies to MAD, more investigation is necessary}\DIFaddend .
	\end{itemize}

	We thus answer the initial questions posed in this article:

	\begin{itemize}
		\item How good are ID document examiners at detecting morphing attacks?
		\begin{itemize}
			\item  \DIFdelbegin \DIFdel{\Rev{The identity document examiners are still prone to make errors in detection of morphing attacks and their performance varies based on line of work (refer Figure~\ref{fig:s-mad-line-of-work} and Figure~\ref{fig:d-mad-line-of-work}). }
		}\DIFdelend \DIFaddbegin \DIFadd{The performance of identity document examiners varies depending on their field of employment and they are nonetheless susceptible to errors in the detection of morphing attacks (refer Figure~\ref{fig:s-mad-line-of-work} and Figure~\ref{fig:d-mad-line-of-work}).
		}\DIFaddend \end{itemize}
		\item Are people with certain types of training better than others at detecting morphing attacks?
		\begin{itemize}
			\item  \DIFdelbegin \DIFdel{\Rev{Analyzing the accuracy of various groups of observers indicates that certain types of training having positive impact for MAD than others.}
		}\DIFdelend \DIFaddbegin \DIFadd{The accuracy of various groups of observers indicate that some methods of training were more effective for MAD than others.
		}\DIFaddend \end{itemize}
		\item Does long experience working in a certain field (for instance, facial examination in ID control) positively impact performance?
		\begin{itemize}
			\item \DIFdelbegin \DIFdel{\Rev{We note an indicative correlation between detection accuracy and experience in the line of work. As compared to observers in different line of work, face comparison experts perform better in detecting morphs suggesting certain positive effect of prior experience.}
		}\DIFdelend \DIFaddbegin \DIFadd{We note an indicative correlation between detection accuracy and experience in the line of work. Face comparison experts perform better in identifying morphs that may indicate a certain favorable influence of prior experience than observers in other fields of work.
		}\DIFaddend \end{itemize}
		\item \Rev{Are expert observers with training or experience in checking identity/identity documents better than those without training?}
		\begin{itemize}
			\item \DIFdelbegin \DIFdel{\Rev{The observers with no-training can perform as good as human experts with training. However, it must be noted that the average error rate of the people with training is much lower than the people with no training (either document training or facial examination), indicating that people with no training or experience in checking identity/identity documents perform on an average lower.}
		}
\DIFdelend \DIFaddbegin \DIFadd{The observers with no training can perform as well as human experts with training. However, it should be noted that the average error rate of those who have received training is significantly lower than that of those who have not (for either document training or facial examination), indicating that those who have not received any training or experience in.
		}\DIFaddend 
	\end{itemize}
	\item How do MAD algorithms perform compared to human observers?
		\DIFaddbegin 
		\DIFaddend \begin{itemize}
			\item \DIFdelbegin \DIFdel{\Rev{The algorithms perform better in both S-MAD and D-MAD setting as compared to human observers indicating a gap in ability to detect morphs as efficiently as algorithms (refer Table~\ref{tab:d-mad-accuracy-versus-observers-s-mad} and Table~\ref{tab:s-mad-accuracy-versus-observers-s-mad}).}
		}\DIFdelend \DIFaddbegin \DIFadd{The algorithms perform better in both S-MAD and D-MAD setting as compared to human observers indicating a gap in the ability to detect morphs as efficiently as algorithms  (refer Table~\ref{tab:d-mad-accuracy-versus-observers-s-mad} and Table~\ref{tab:s-mad-accuracy-versus-observers-s-mad}).
		}\DIFaddend \end{itemize}
	\end{itemize}


	\subsection{Future Directions}
	\label{ssec:future-directions}
	\DIFdelbegin \DIFdel{One of the potential directions for the follow-up work is to understand the improvement of detection accuracy if observers with different competence decide on a given image }\DIFdelend \DIFaddbegin \DIFadd{Understanding the improvement in detection accuracy when observers with varying levels of competence select a particular image is one of the prospective avenues for subsequent research}\DIFaddend . For instance, it would be interesting to understand if border guards and expert facial examiners consider different facial properties before deciding on the image as border guards often have some document training in many countries. \DIFdelbegin \DIFdel{A better }\DIFdelend \DIFaddbegin \DIFadd{Designing better algorithms and including the factors that people consider when choosing the best course of action could result in a greater }\DIFaddend understanding of the decision-making process\DIFdelbegin \DIFdel{could lead to designing better algorithms and incorporating the areas humans look at while making the right decision}\DIFdelend . Furthermore, obtaining the confidence of observers in the decision-making process can be used to fully understand what experts observe in an image to deem it as a morphed image.

	A question that needs to be answered in the future is if ID examiners with training in MAD perform better than people with no MAD training. \DIFdelbegin \DIFdel{Further, it is of interest to measure the }\DIFdelend \DIFaddbegin \DIFadd{The }\DIFaddend impact of training \DIFdelbegin \DIFdel{. Another important question to be addressed is to verify if super-recognizers possess an innate ability to detect morphing attacks without training}\DIFdelend \DIFaddbegin \DIFadd{may also be measured, which is interesting. The ability of superrecognizers to recognize morphing attacks intuitively without training is a crucial issue that needs to be investigated}\DIFaddend . In the current study, we have 23 \DIFdelbegin \DIFdel{super-recognizers}\DIFdelend \DIFaddbegin \DIFadd{superrecognizers}\DIFaddend , and a dedicated study on \DIFdelbegin \DIFdel{super-recognizers }\DIFdelend \DIFaddbegin \DIFadd{superrecognizers }\DIFaddend is currently being conducted whose findings will be reported in the follow-up work.


	\subsection{Potential errors in the study}
	\label{ssec:potential-errors}
	We \DIFdelbegin \DIFdel{recognise }\DIFdelend \DIFaddbegin \DIFadd{recognize }\DIFaddend that the study may, in some cases, be prone to error. For \DIFdelbegin \DIFdel{example, participants knowing they were being tested may have put in an extra }\DIFdelend \DIFaddbegin \DIFadd{instance, participants may have exerted more }\DIFaddend effort and spent more time on the tasks in the experiments than they would have in a real-life scenario \DIFaddbegin \DIFadd{if they were aware that they were being tested}\DIFaddend . In contrast, other participants may have spent less time and made less effort, as it was not a real case they were working on.

	We also \DIFdelbegin \DIFdel{recognise that we may have introduced bias by periodically informing the participants of their scores during the experiments. It may have affected the results on whether being exposed to many morphed images actually helps improve performance . Fatigue could also have played a role and negatively affected performance, as }\DIFdelend \DIFaddbegin \DIFadd{acknowledge that by periodically updating the participants on their results throughout the experiment, we may have introduced bias. It might have had an impact on the findings regarding whether exposure to numerous morphed images improves performance. Given that }\DIFaddend the experiments would \DIFdelbegin \DIFdel{likely be }\DIFdelend \DIFaddbegin \DIFadd{have been }\DIFaddend an additional task on top of their \DIFdelbegin \DIFdel{ordinary work}\DIFdelend \DIFaddbegin \DIFadd{regular work, fatigue may have also been a factor and negatively impacted performance}\DIFaddend .

	\DIFdelbegin 

\DIFdelend The time the participants spent on the experiments could also be prone to errors, as participants may not have clicked on the \DIFdelbegin \DIFdel{"Continue later" }\DIFdelend \DIFaddbegin \DIFadd{“Continue later” }\DIFaddend button if they, for example, were interrupted during the experiments. \DIFdelbegin \DIFdel{Finally, knowing }\DIFdelend \DIFaddbegin \DIFadd{Additionally, participants may have experienced stress if they were aware }\DIFaddend that the time spent was \DIFdelbegin \DIFdel{recorded could also have introduced stress to some participants, and these factorsneed further investigation}\DIFdelend \DIFaddbegin \DIFadd{being recorded. More research is needed to fully understand these factors}\DIFaddend .

	\DIFaddbegin

	\DIFaddend \section{Conclusion}
	\label{sec:conclusion}
	\DIFdelbegin \DIFdel{Morphing attack detection }\DIFdelend \DIFaddbegin \DIFadd{MAD }\DIFaddend has been extensively studied using automated algorithms, and a very limited set of works have investigated the detection ability of human observers in detecting morphs. We \DIFdelbegin \DIFdel{conduct a benchmark study of human observers who }\DIFdelend compare face images in \DIFdelbegin \DIFdel{everyday professional life in }\DIFdelend two different settings\DIFdelbegin \DIFdel{of }\DIFdelend - S-MAD and D-MAD\DIFdelbegin \DIFdel{. A new database or morphed images was created using }\DIFdelend \DIFaddbegin \DIFadd{ using human observers in everyday professional life as a benchmark. With the help of }\DIFaddend 48 \DIFdelbegin \DIFdel{unique subjectsleading to $400$ morphed images under }\DIFdelend \DIFaddbegin \DIFadd{different subjects, 400 morphed images were produced using }\DIFaddend the Differential-MAD (D-MAD) setting, and \DIFdelbegin \DIFdel{$400$ }\DIFdelend \DIFaddbegin \DIFadd{400 }\DIFaddend probe images were \DIFdelbegin \DIFdel{captured from ABC gate reflecting }\DIFdelend \DIFaddbegin \DIFadd{taken at the ABC gate to represent a }\DIFaddend border-crossing scenario. In addition, a new database of \DIFdelbegin \DIFdel{$180$ morphed images were }\DIFdelend \DIFaddbegin \DIFadd{180 morphed images was developed }\DIFaddend under the Single Image - MAD (S-MAD) setting. The benchmark \DIFdelbegin \DIFdel{analysed }\DIFdelend \DIFaddbegin \DIFadd{analyzed }\DIFaddend with 469 observers for D-MAD and 410 observers for S-MAD from more than 40 countries indicates the challenging nature of morphed image detection. \DIFdelbegin \DIFdel{The analysis points to expert observers' missing competence leading to not detecting }\DIFdelend \DIFaddbegin \DIFadd{According to the analysis, }\DIFaddend close to 30\% of morphed images \DIFdelbegin \DIFdel{. Several sub-analysis also indicates varying competence of }\DIFdelend \DIFaddbegin \DIFadd{were missed by expert observers due to a lack of competence. Numerous subanalyses also show that }\DIFaddend different observers from \DIFdelbegin \DIFdel{different line-of-work in detecting the }\DIFdelend \DIFaddbegin \DIFadd{various fields of expertise have varying degrees of proficiency in spotting }\DIFaddend morphing attacks. The analysis also \DIFdelbegin \DIFdel{points to initial indications of improved }\DIFdelend \DIFaddbegin \DIFadd{shows early signs of enhanced }\DIFaddend morphing detection if the observers \DIFdelbegin \DIFdel{are trained specifically on morphing detection . 
	}\DIFdelend \DIFaddbegin \DIFadd{receive specialized morphing detection training.
	}\DIFaddend 

	\bibliographystyle{IEEEtran}
	\bibliography{morphing-human-observer-200411}

	\ifCLASSOPTIONcaptionsoff
	\newpage
	\fi

	\clearpage
	\newpage
	\counterwithin{figure}{section}
	\counterwithin{table}{section}
	\setcounter{page}{1}
	\onecolumn
	\begin{appendices}
		\section*{Supplementary Material \\
			Analyzing Human Observer Ability in Morphing Attack Detection - Where Do We Stand?\\
			Sankini Rancha Godage$^\ddag$, Frøy Løvåsdal$^\dag$, Sushma Venkatesh$^\ddag$, \\ Kiran Raja$^\ddag$, Raghavendra Ramachandra$^\ddag$, Christoph Busch$^\ddag$ \\$^\ddag$Norwegian University of Science and Technology (NTNU), Norway \\ $^\dag$The Norwegian Police Directorate, Norway}

		\section{\Rev{Related works on Human observers for general face recognition}}
		\Rev{The vast amount of recent research in the field of face comparison demonstrates that people make mistakes when comparing faces that are unfamiliar to them \cite{bruce2001matching,megreya2006unfamiliar}. Identifying differences in unfamiliar faces is a difficult task for humans \cite{blauch2020computational}.  White et al. \cite{white2014passport} found that the performance of human experts trained in face identification (passport officers, ID card checkers) and human volunteers with no prior experience (student volunteers) in face verification had similar error rates \footnote{referred as matching in the original article}.}

		\Rev{Thirty passport officers took part in the study (21 Females, Mean age = 48.0, SD = 11.7) where an average of 10\% error was reported. Additionally, 14\% of nonmated photos were incorrectly accepted while 6\% of valid photos were incorrectly rejected. The study also discovered no connection between the proficiency of the face recognition test and the experience of the observers. In the second test conducted after two years, the same subject group was requested to scan their current passport picture and a new facial image (taken under a better-illuminated condition and improved camera quality). The authors compared the recently captured image with the subjects’ most recent passport photos using the newly captured image as the target image. Each participant was asked to evaluate the image against every possible pair combination. Eighty-four trials of images were shown to the observers and overall accuracy was shown to be 70.9\% (89.4\% accuracy on nonmated comparison trials). The research found that comparing unfamiliar faces is a human task that is prone to error (whether trained or not). The study also discovered that making decisions are not improved by presenting an outdated reference photo identification document.}

	 	\Rev{Phillips et al. \cite{phillips2018face} examined the performance of the human observers and \Rev{algorithms} in a National Institute of Standards and Technology(NIST) project \cite{o2007face,o2008humans}. These test results showed that both humans and machines could accurately detect faces. To analyze the experimental data Phillips et al.\cite{phillips2018face} proposed the Cross-Modal Performance Analysis framework, which was adapted from neuroscience techniques. They used both static images and video imagery in this experiment. In experimental trials, two static face images or videos were on the digital screen in each trial. They compared two images captured under different conditions like a studio environment, ambient lighting, different hairstyle, and clothes. The observers were given five options “(i) Are you sure they are the same person? (ii) Not sure they are the same. (iii) Do not know (iv) Different person (v) Not sure they are different persons.” To analyze the data, the authors used the Receiver Operating Characteristic Curve (ROC). According to O’Toole et al. \cite{o2007face}’s recommendations, all images and videos were only shown for 2 seconds. Humans outperform algorithms when faced with challenging static faces and video, according to research by O’Toole et al. \cite{o2007face}.}

		\section{Quality of morphed images}
		We investigate whether there are any discernible differences between the reference images and the ABC gate images in two different contexts: (1) digital morphed images without postprocessing and with postprocessing compared against digital bona fide images, (2) postprocessed and printed-scanned morphed images compared against ABC gate images. Through the Kruskal-Wallis test, we note in the former case we obtain a \Rev{\textit{p}=.502}, and in the latter, we note \Rev{\textit{p}=.806} both suggesting that observers do not have any advantage when the images are not postprocessed. This illustrates the difficulty of detecting morphs in the digital or print-scan domain while also demonstrating how well-made the morphs produced in this work are at deceiving a human observer.

		\section{Illustration of one subject morphed against all subjects}
			\begin{figure}[H]
			\centering
			\subfloat{\includegraphics[width = 10ex]{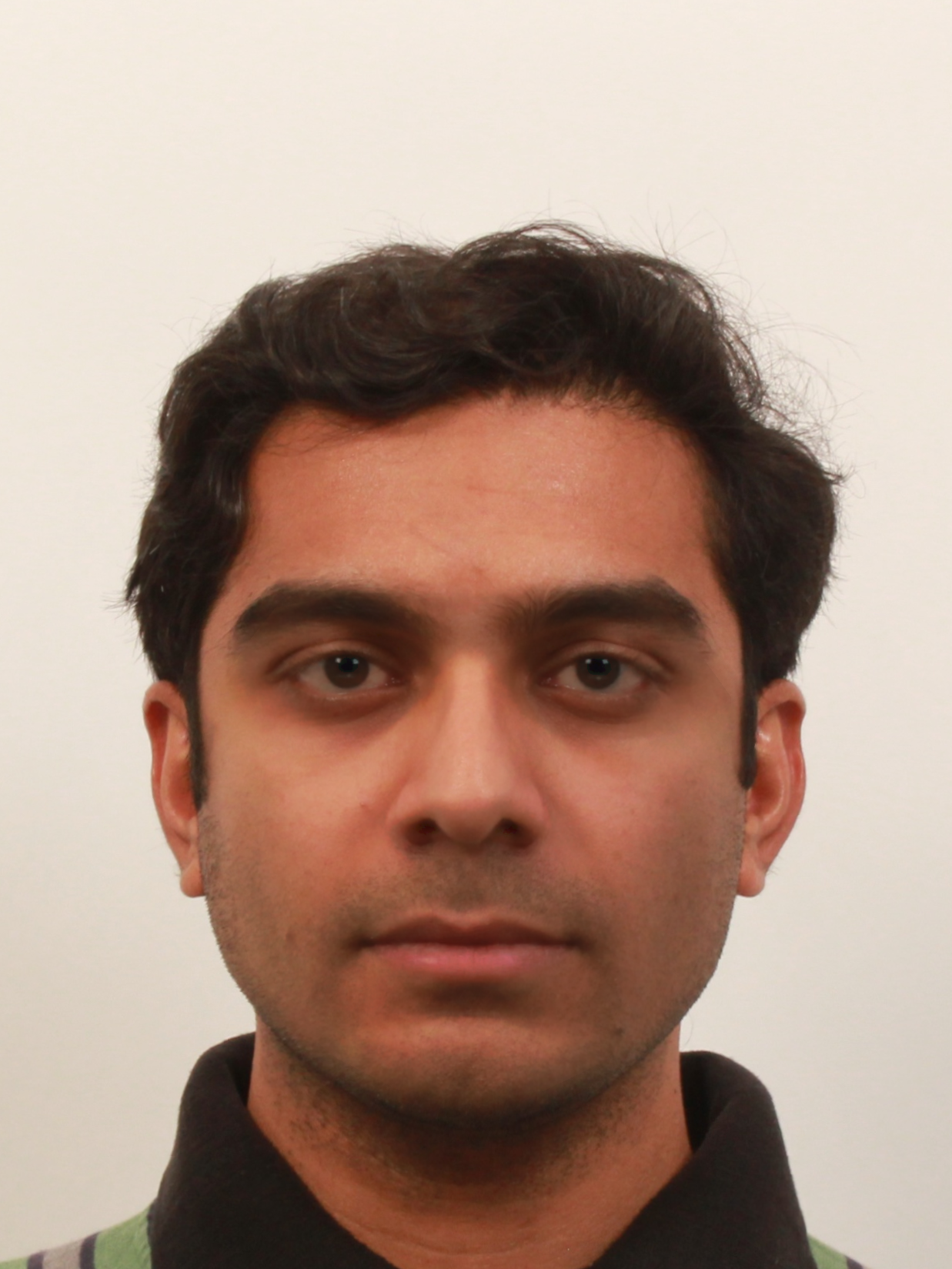}}
			\subfloat{\includegraphics[width = 10ex]{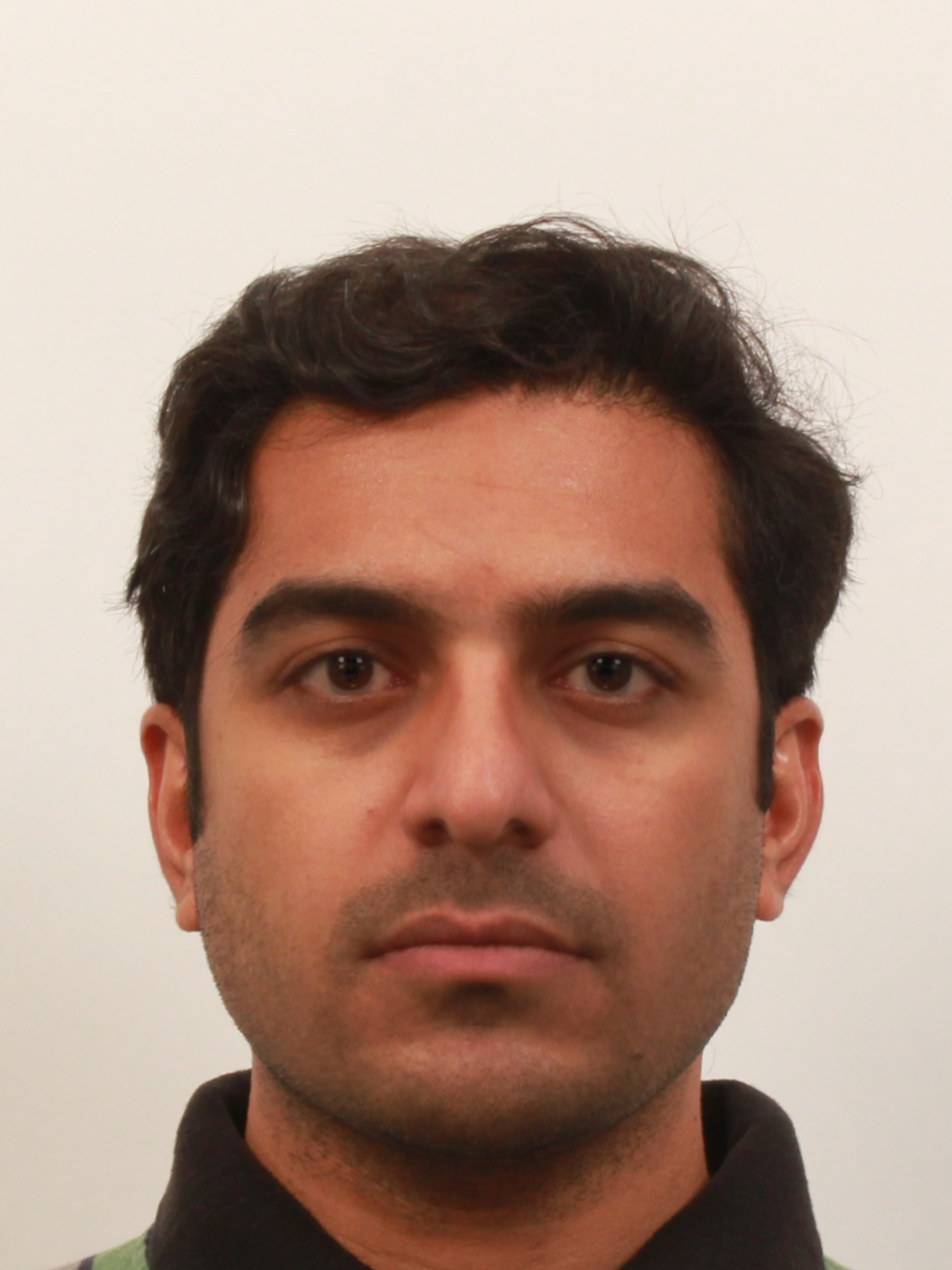}}
			\subfloat{\includegraphics[width = 10ex]{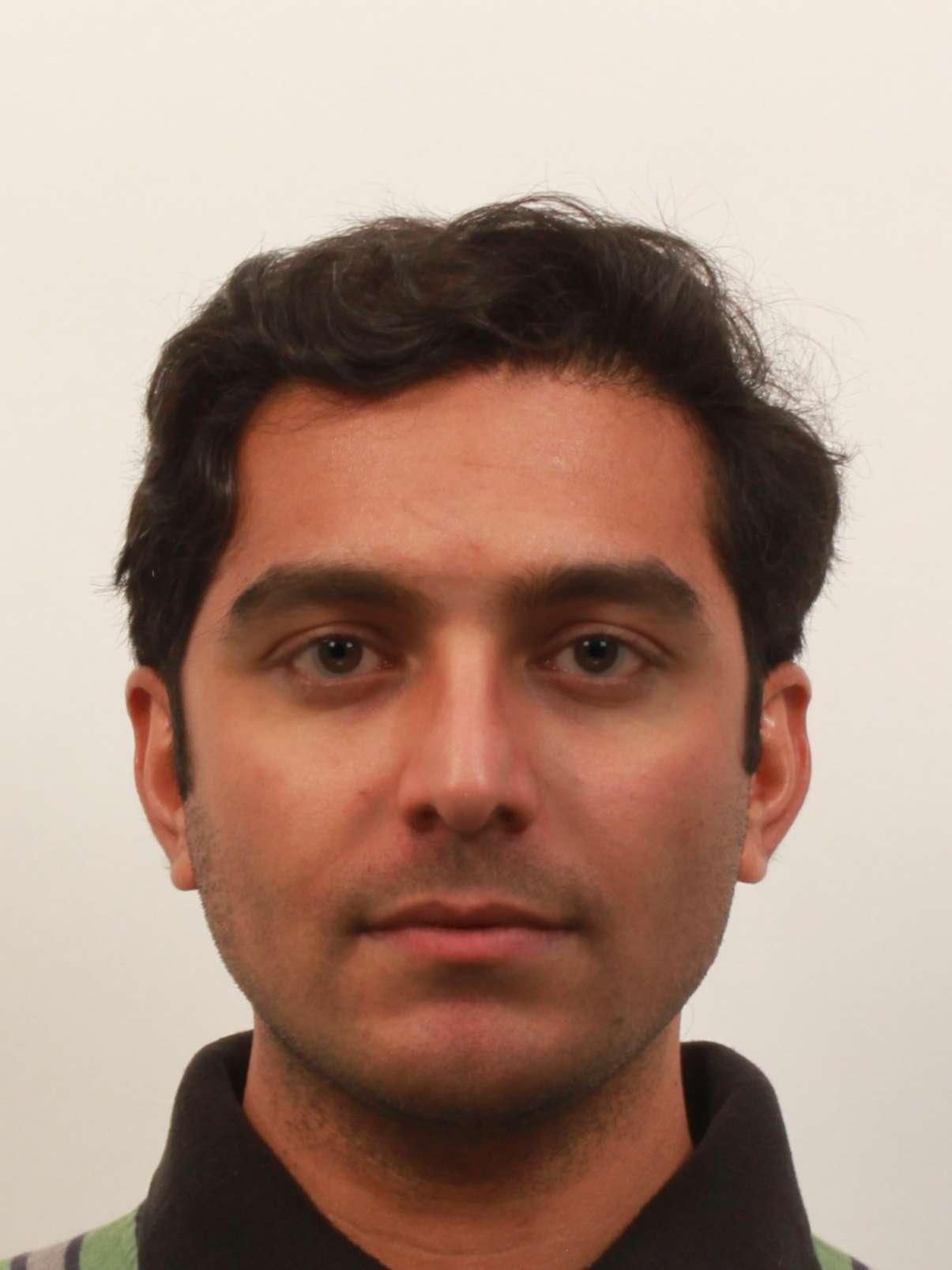}}
			\subfloat{\includegraphics[width = 10ex]{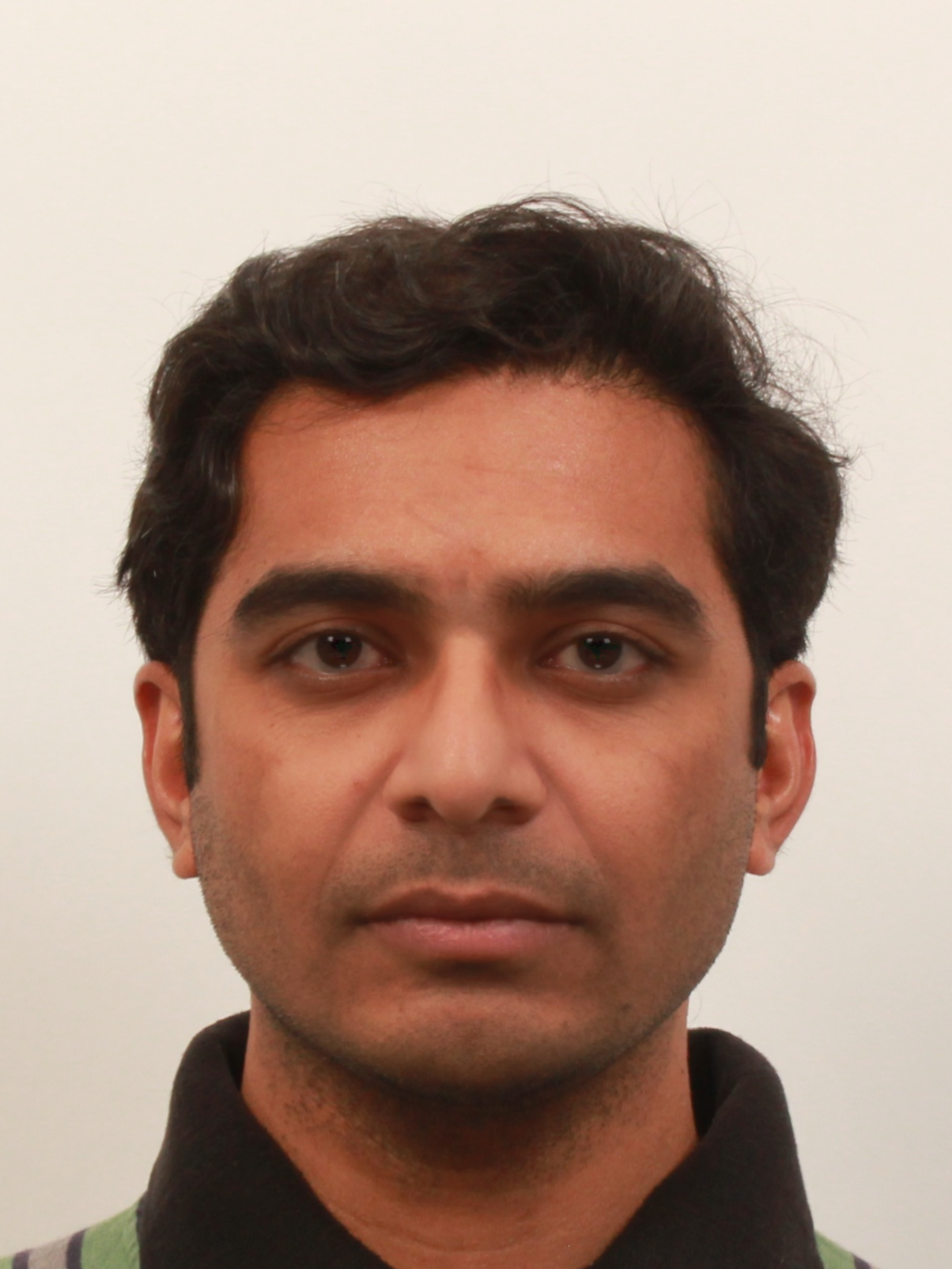}}
			\subfloat{\includegraphics[width = 10ex]{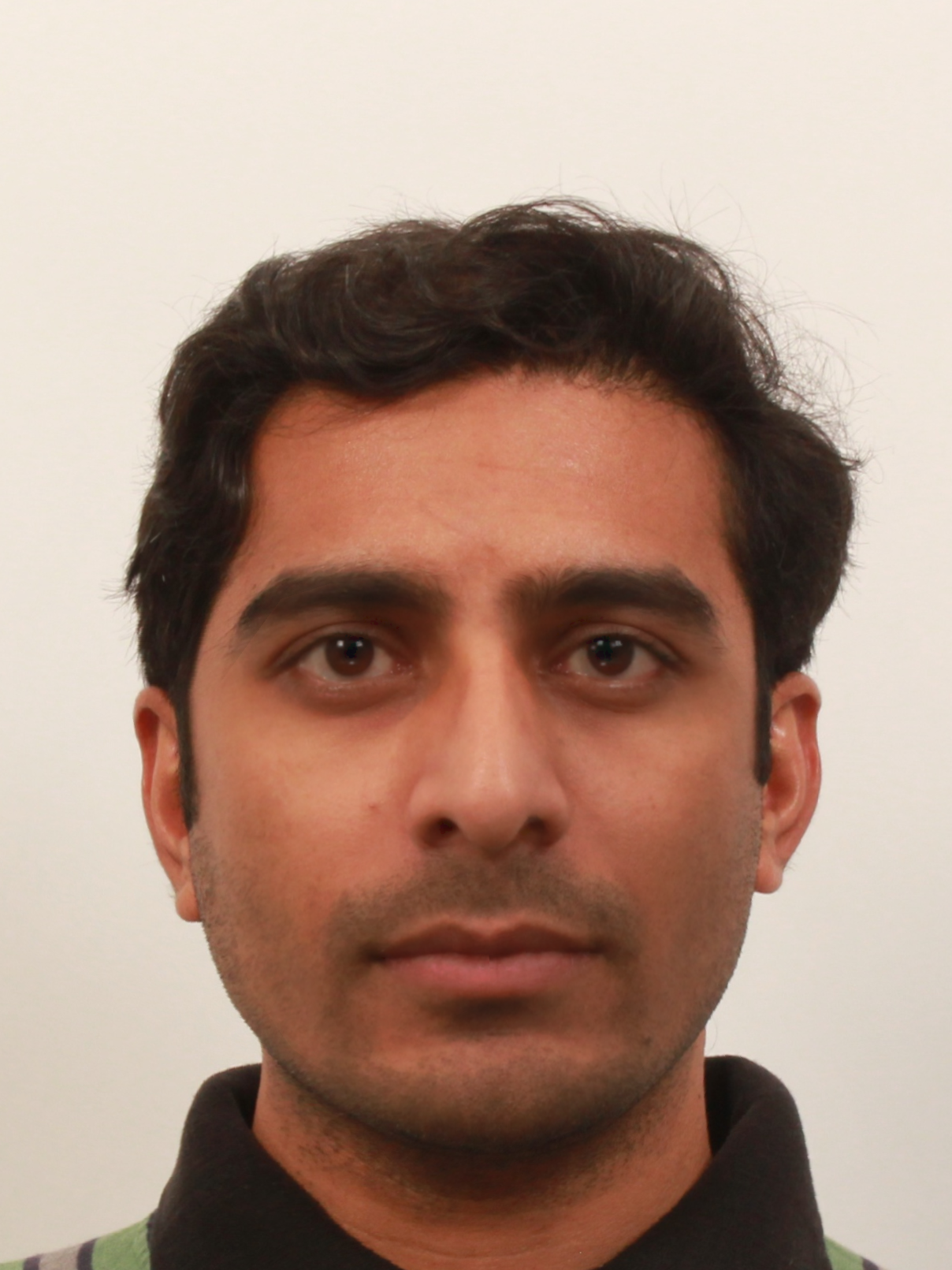}}\\
			\subfloat{\includegraphics[width = 10ex]{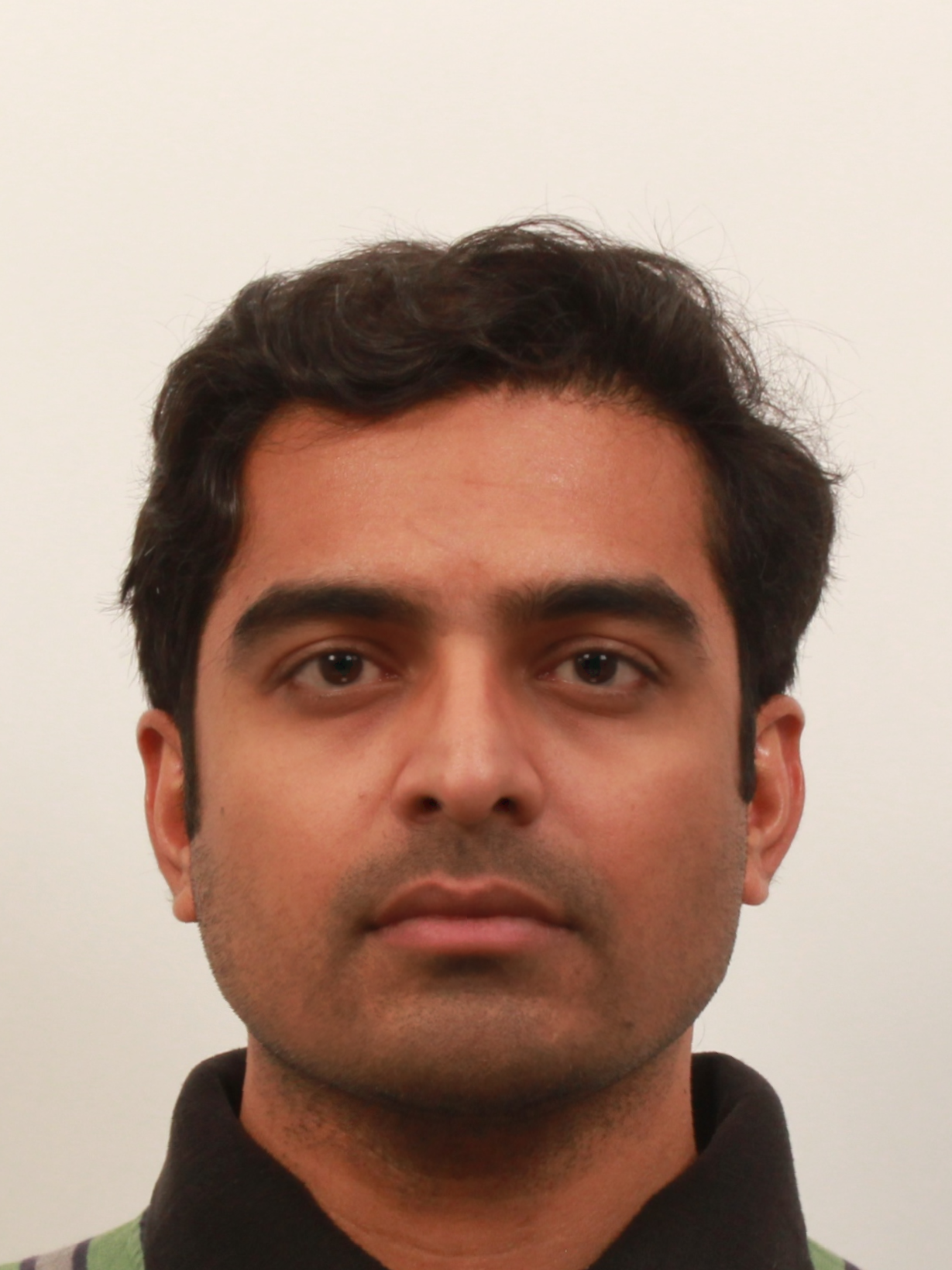}}
			\subfloat{\includegraphics[width = 10ex]{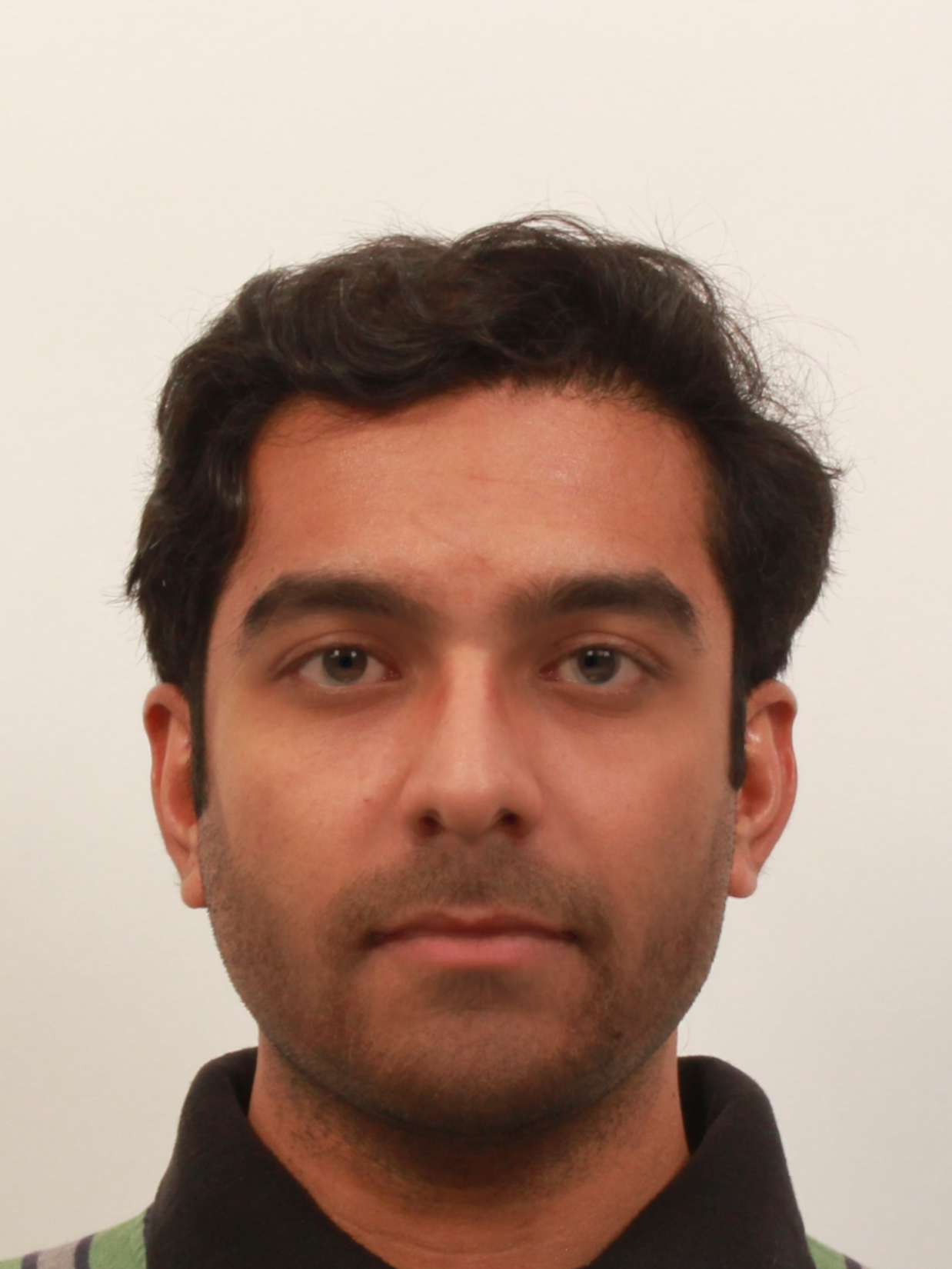}}
			\subfloat{\includegraphics[width = 10ex]{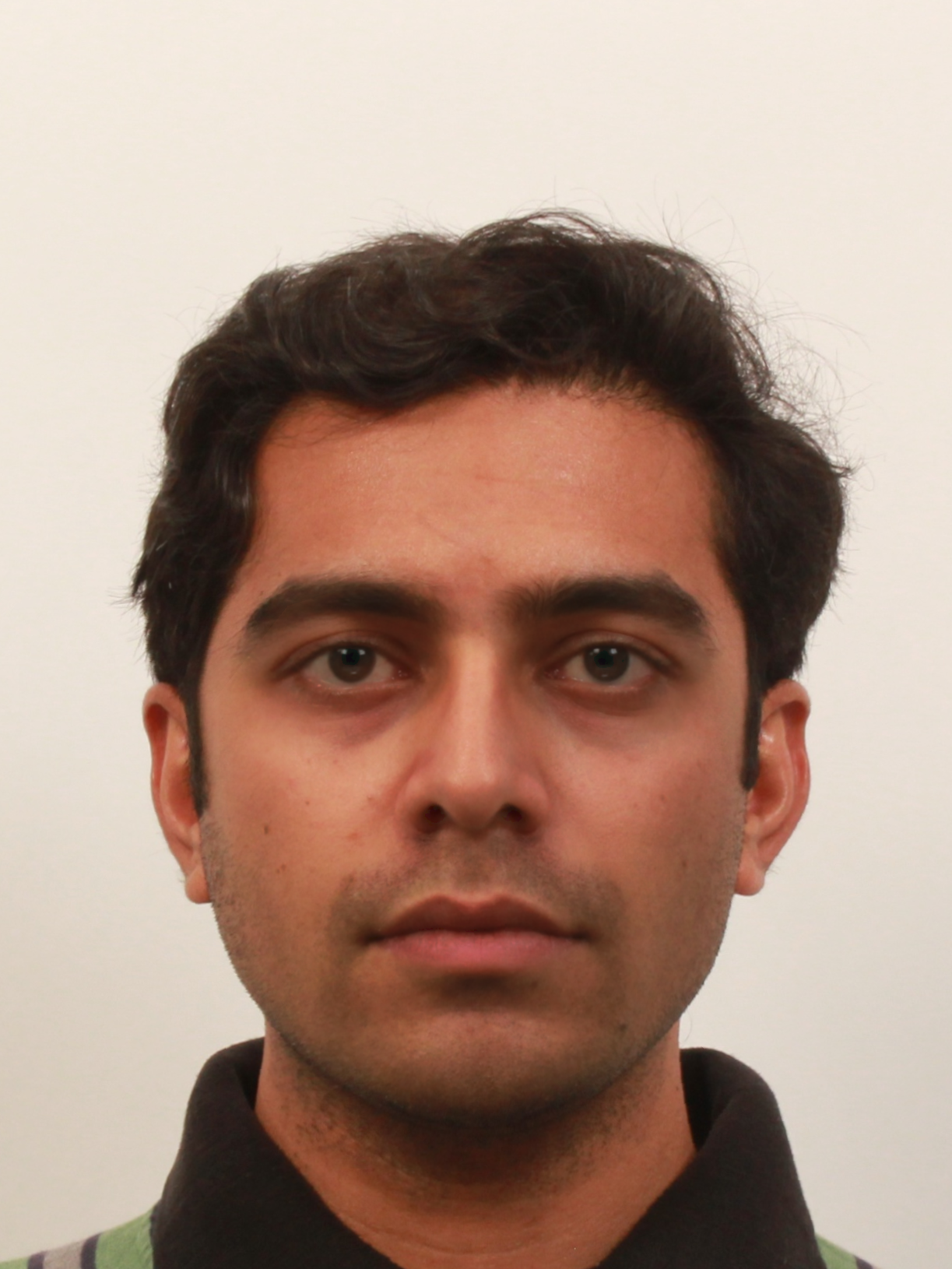}}
			\subfloat{\includegraphics[width = 10ex]{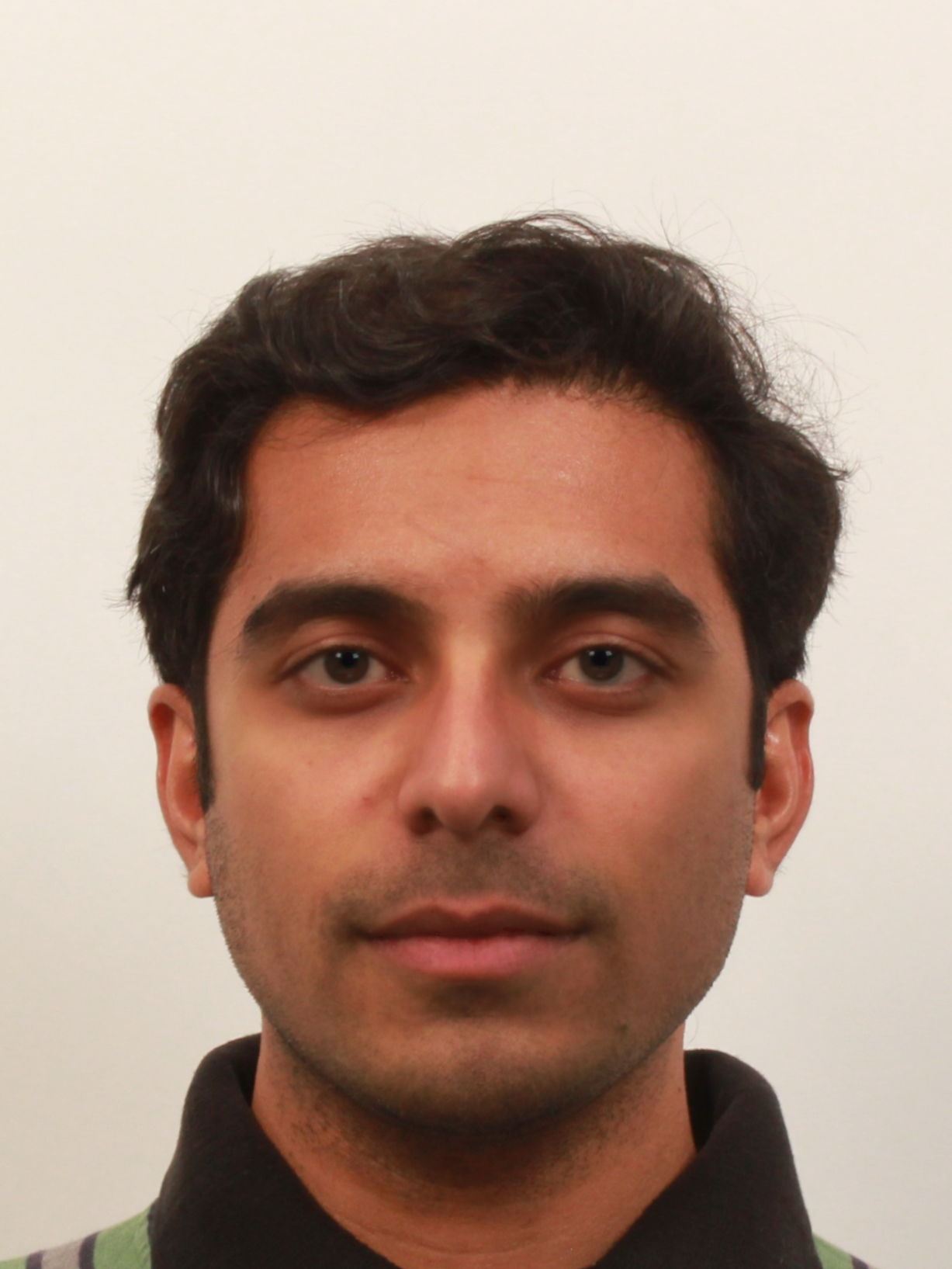}}
			\subfloat{\includegraphics[width = 10ex]{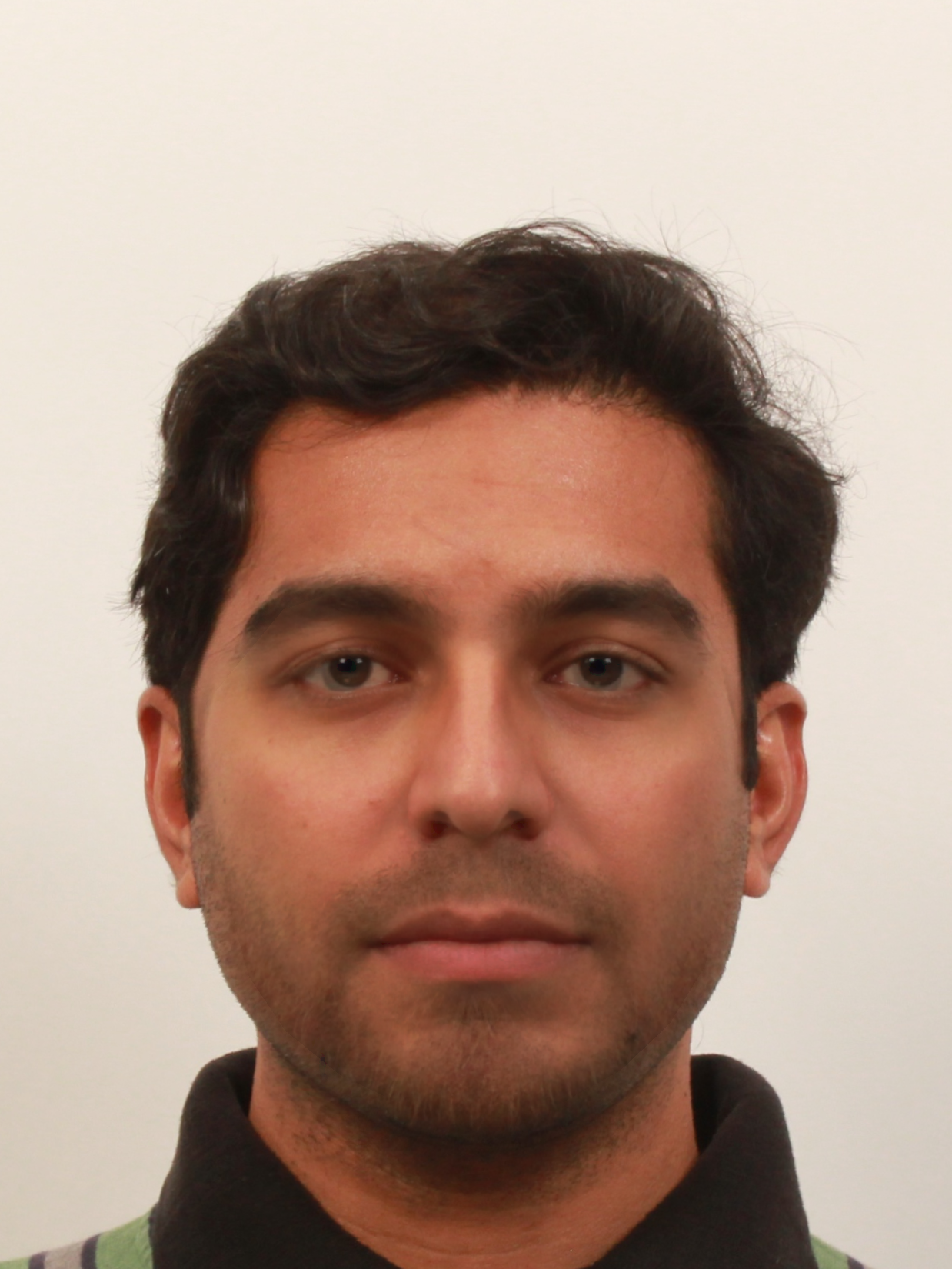}}\\
			\subfloat{\includegraphics[width = 10ex]{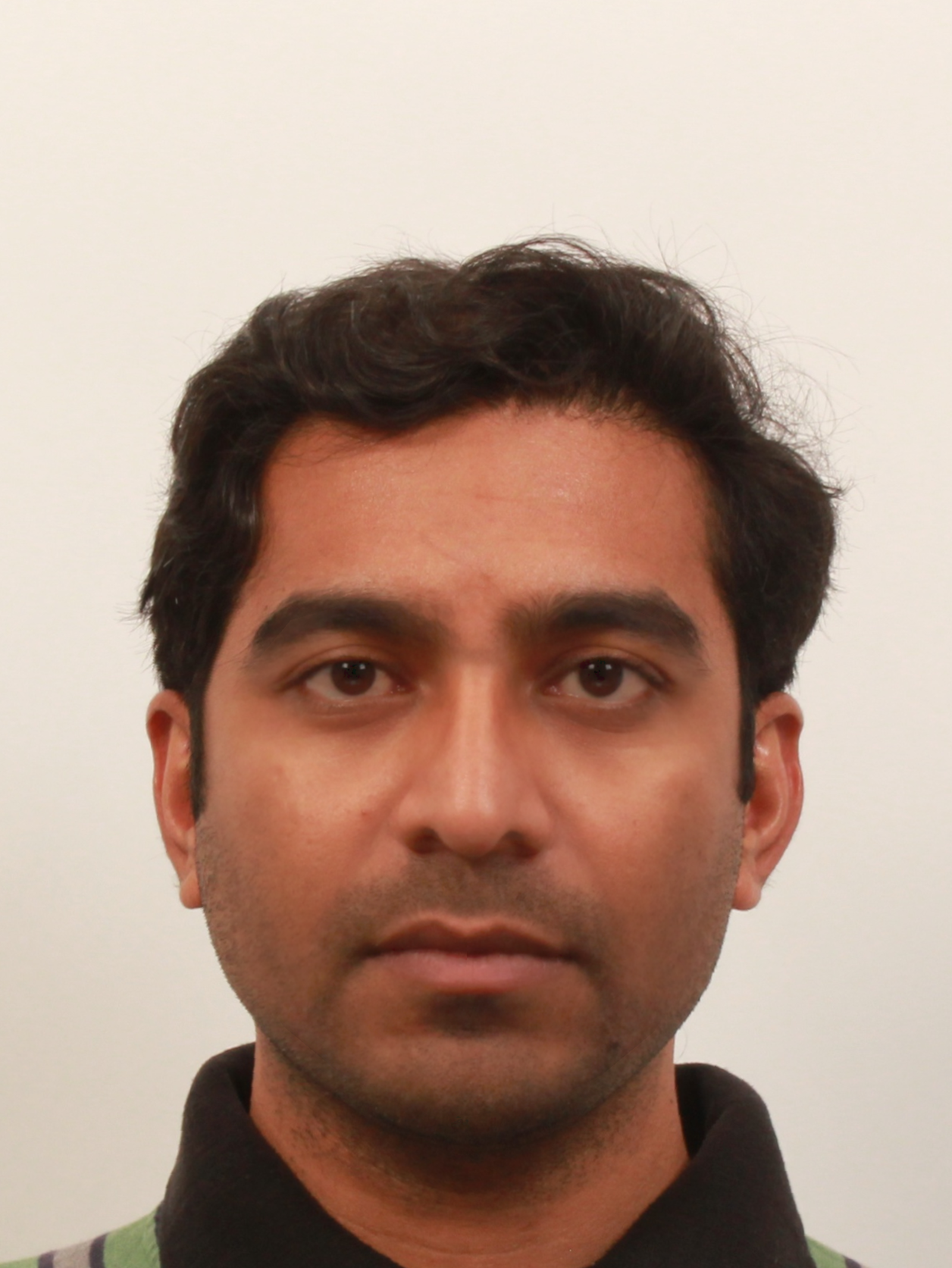}}
			\subfloat{\includegraphics[width = 10ex]{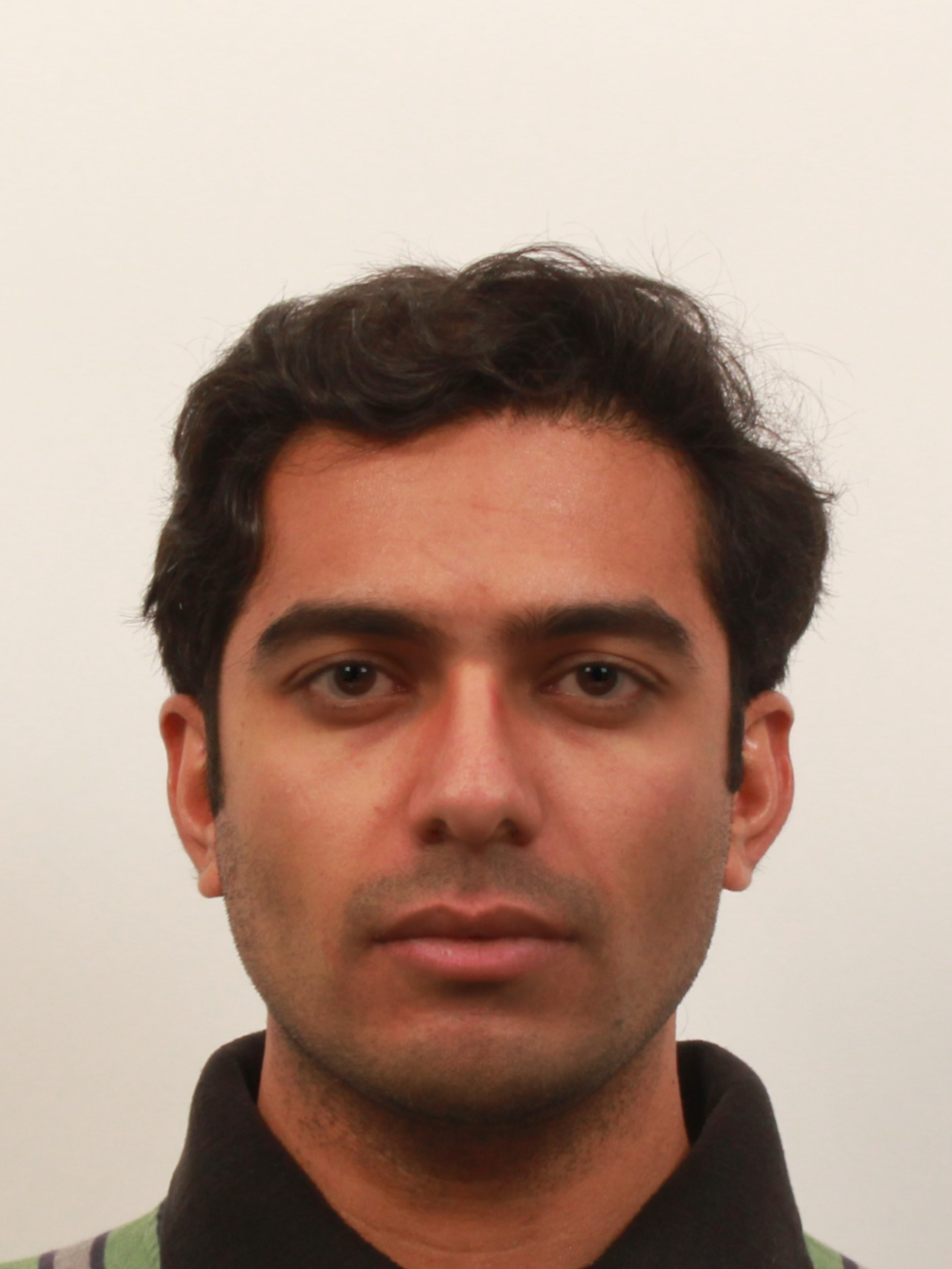}}
			\subfloat{\includegraphics[width = 10ex]{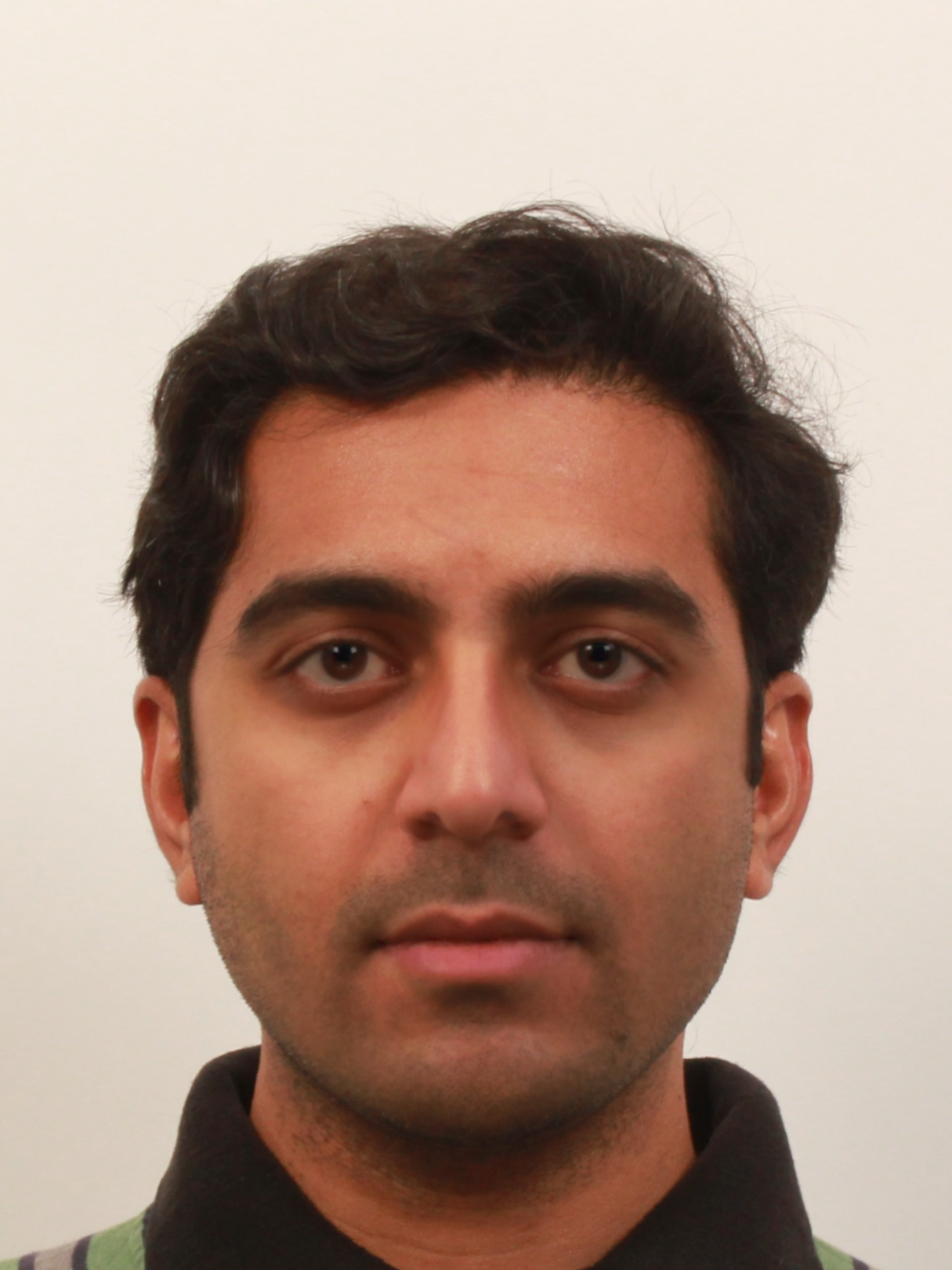}}
			\subfloat{\includegraphics[width = 10ex]{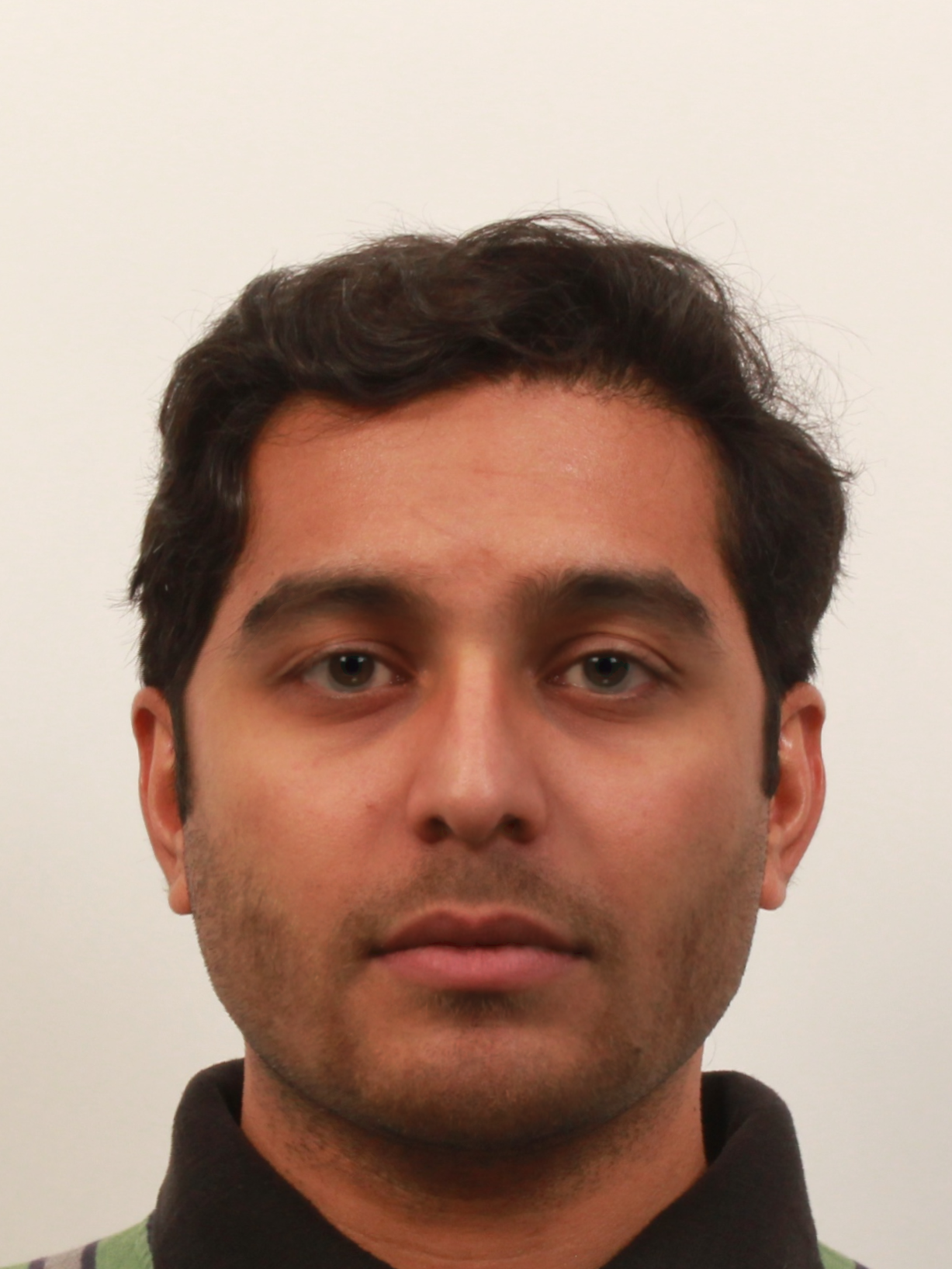}}
			\subfloat{\includegraphics[width = 10ex]{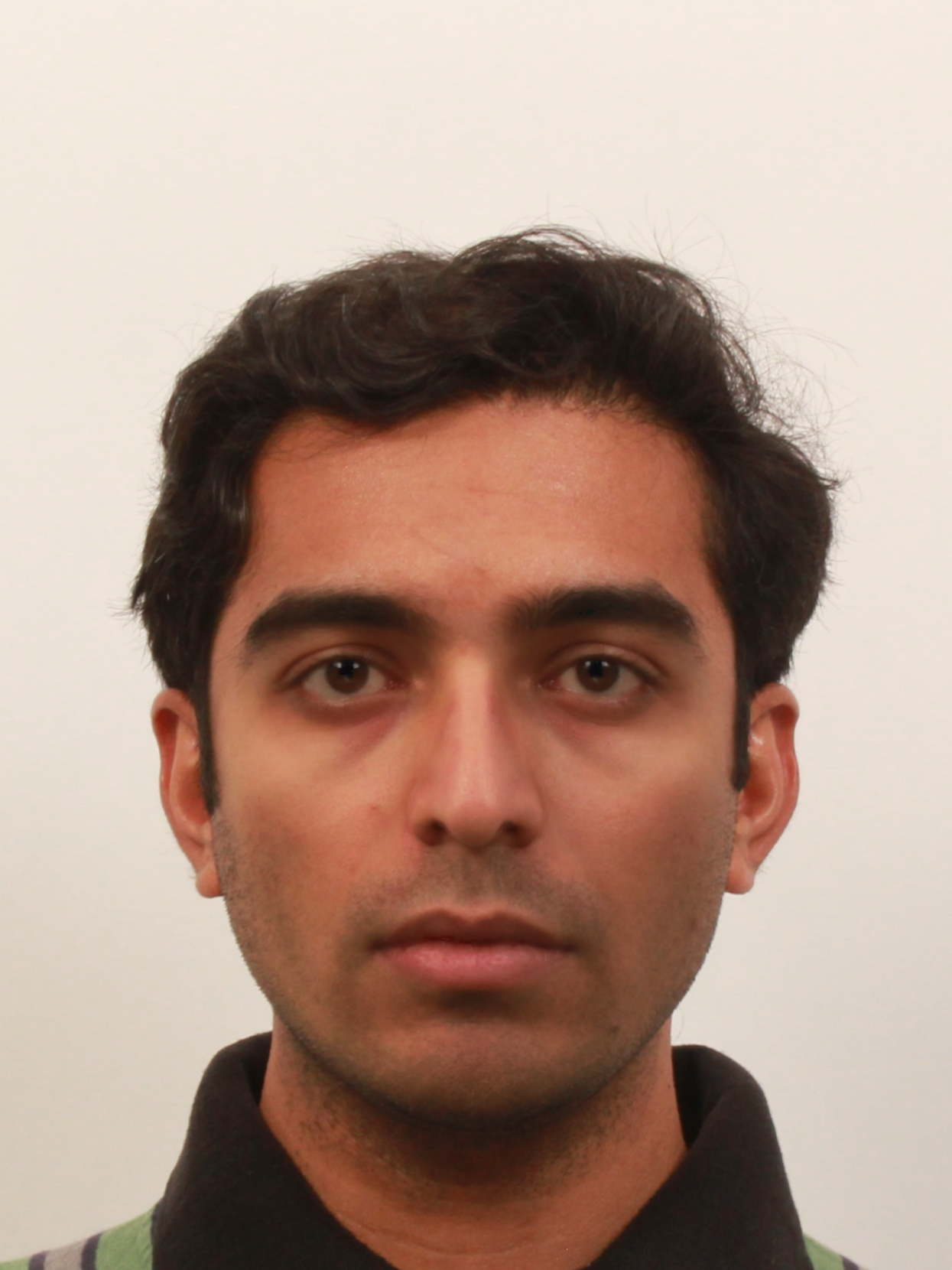}}\\
			\caption{Example of one subject morphed with various subjects in the database with a morphing factor $0.5$. }
			\label{figg18}
		\end{figure}

		\section{Additional Analysis - MAD Accuracy v/s Total Time Spent}
\begin{figure}[h]
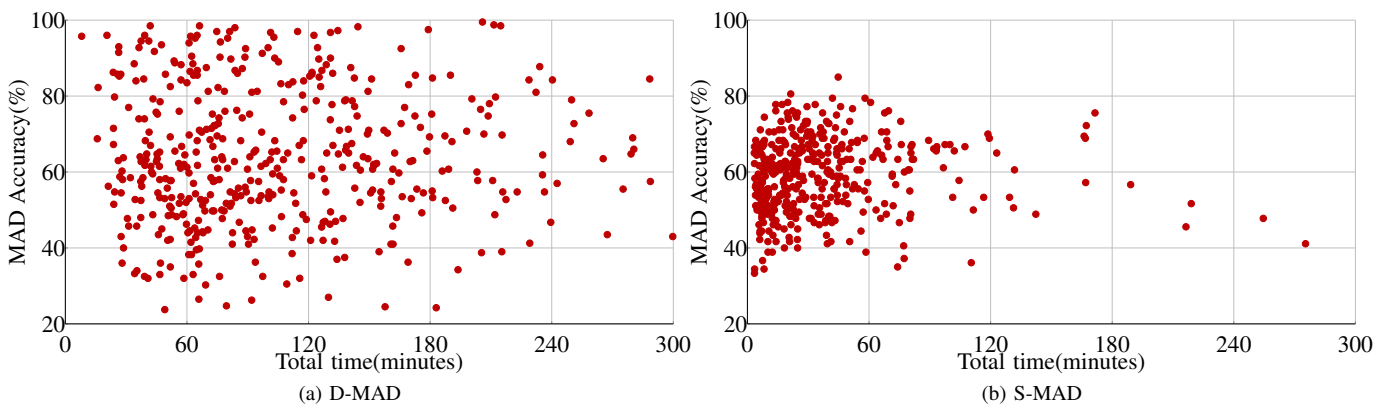

			\centering 
			\subfloat[D-MAD]{%
				\includegraphics[trim=0 0 0 0, width=.5\linewidth]{figures-210915/d_mad_time_vs_acc.tex}%
				\label{fig:d-mad-time-spent}%
			}
			\subfloat[S-MAD]{%
				\includegraphics[trim=0 0 0 0, width=.5\linewidth]{figures-210915/s_mad_time_vs_acc.tex}%
				\label{fig:s-mad-time-spent}%
			}
			\caption{MAD accuracy versus the time spent for completing the experiment. (* Illustrated up to 300 minutes for the sake of simplicity)}
			\label{fig:mad-time-spent}%
		\end{figure}

		We also examine the data to see if there is any correlation between the amount of time taken to make decisions and accuracy because we are intrigued by the variations in accuracy that observers exhibit when spotting the morphs. We present the results of total time spent for D-MAD and S-MAD settings as provided in Figure~\ref{fig:mad-time-spent}. We make the following observations from this set of analyses:
		\begin{itemize}
			\item As it appears, there is no clear correlation between the time spent on experiments versus the accuracy obtained. Several observers who finished the experiments in D-MAD settings in less than 90 minutes attained an accuracy of over 90\%. In contrast, some observers who spent less than 90 minutes on the task had lower accuracy (less than 50\%) for 400 image pairs when using the D-MAD setting.
			\item While the accuracy of S-MAD is generally lower, the observers have completed the experiments in less than 90 minutes and have obtained an average accuracy of 65\%.
			\item A small number of observers took a longer time to complete the experiments because the experiment allowed them to do so over a period of time; these observers are spread out after 180 minutes. 
			\item When closely examining the graphs, it can be concluded that there does not seem to be a direct link between the amount of time spent and the level of accuracy attained.
		\end{itemize} 

		\newpage
		\section{Additional Analysis - MAD accuracy v/s Gender}
		\begin{figure}[h]
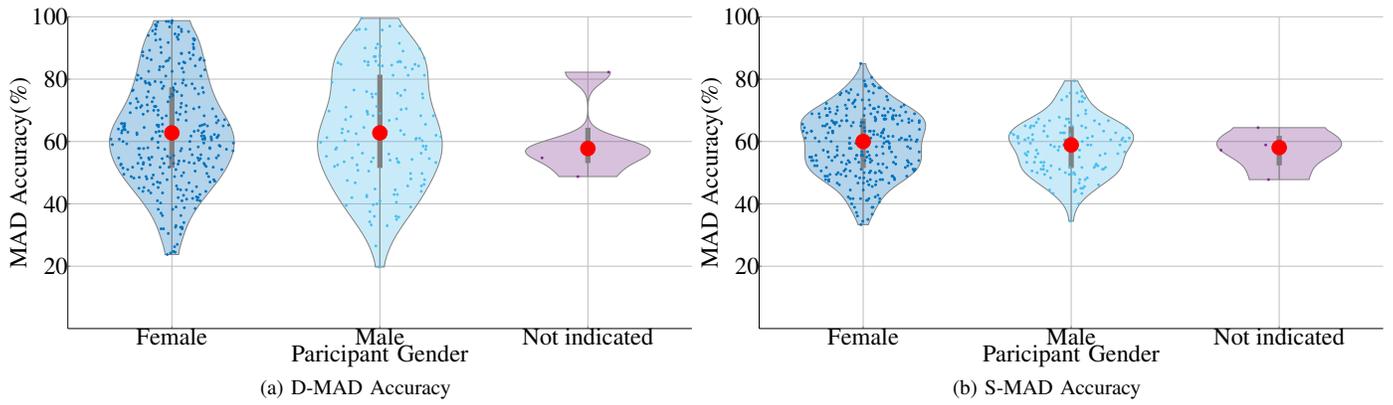

			\centering 
			\subfloat[D-MAD Accuracy]{%
				\includegraphics[width=0.5\columnwidth]{figures-210915/d_mad_genderVSaccuracy_violinplot.tex}%
				\label{fig:d-mad-gender-vs-mad}
			}
			\subfloat[S-MAD Accuracy]{%
				\includegraphics[width=0.5\columnwidth]{figures-210915/s_mad_genderVSaccuracy_violinplot.tex}%
				\label{fig:s-mad-gender-vs-mad}
			}
			\caption{Gender v/s MAD accuracy of human observers}
			\label{fig:gender-vs-mad}
		\end{figure}

		\newpage
		\section{Additional Analysis - MAD accuracy for observers in various organizations}

		\begin{figure}[h]
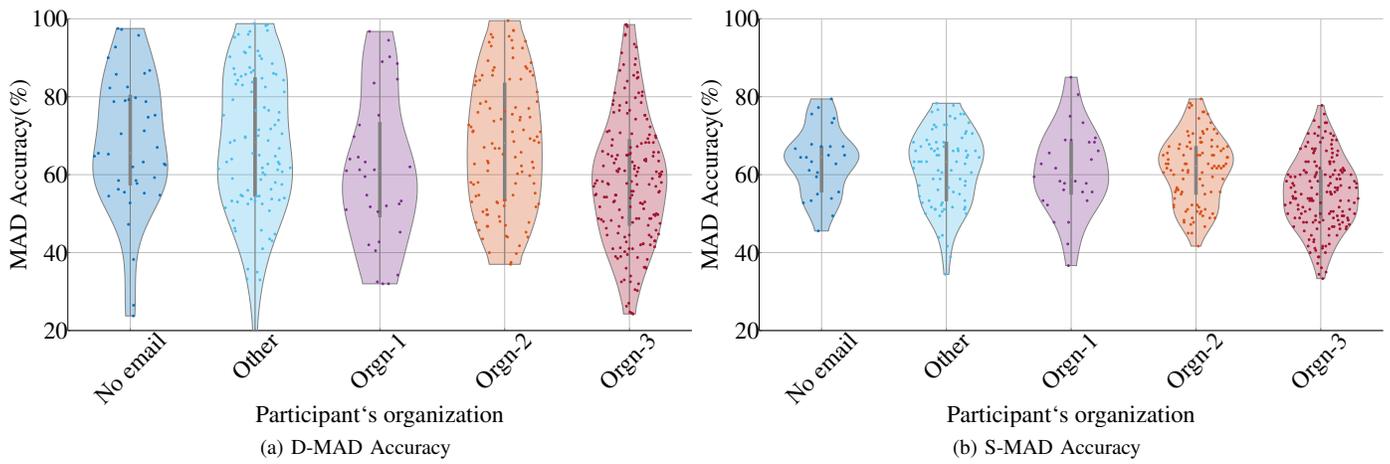

			\centering 
			\subfloat[D-MAD Accuracy]{%
				\includegraphics[width=0.5\columnwidth]{appendix-figures/d_mad_organization_vs_acc_violin.tex}%
				\label{fig:d-mad-organization}
			}
			\subfloat[S-MAD Accuracy]{%
				\includegraphics[width=0.5\columnwidth]{appendix-figures/s_mad_organization_vs_acc_violin.tex}%
				\label{fig:s-mad-organization}
			}
			\caption{MAD accuracy of human observers in various organizations (*Organization names categorized according to nature of work and presented in anonymized form)}
			\label{fig:mad-organization-accuracy}
		\end{figure}
		\newpage
		\section{Additional Analysis - Highest detection of morphs after post-processing in S-MAD setting}
		\begin{figure}[h]
			\centering 
			\includegraphics[width=0.95\columnwidth]{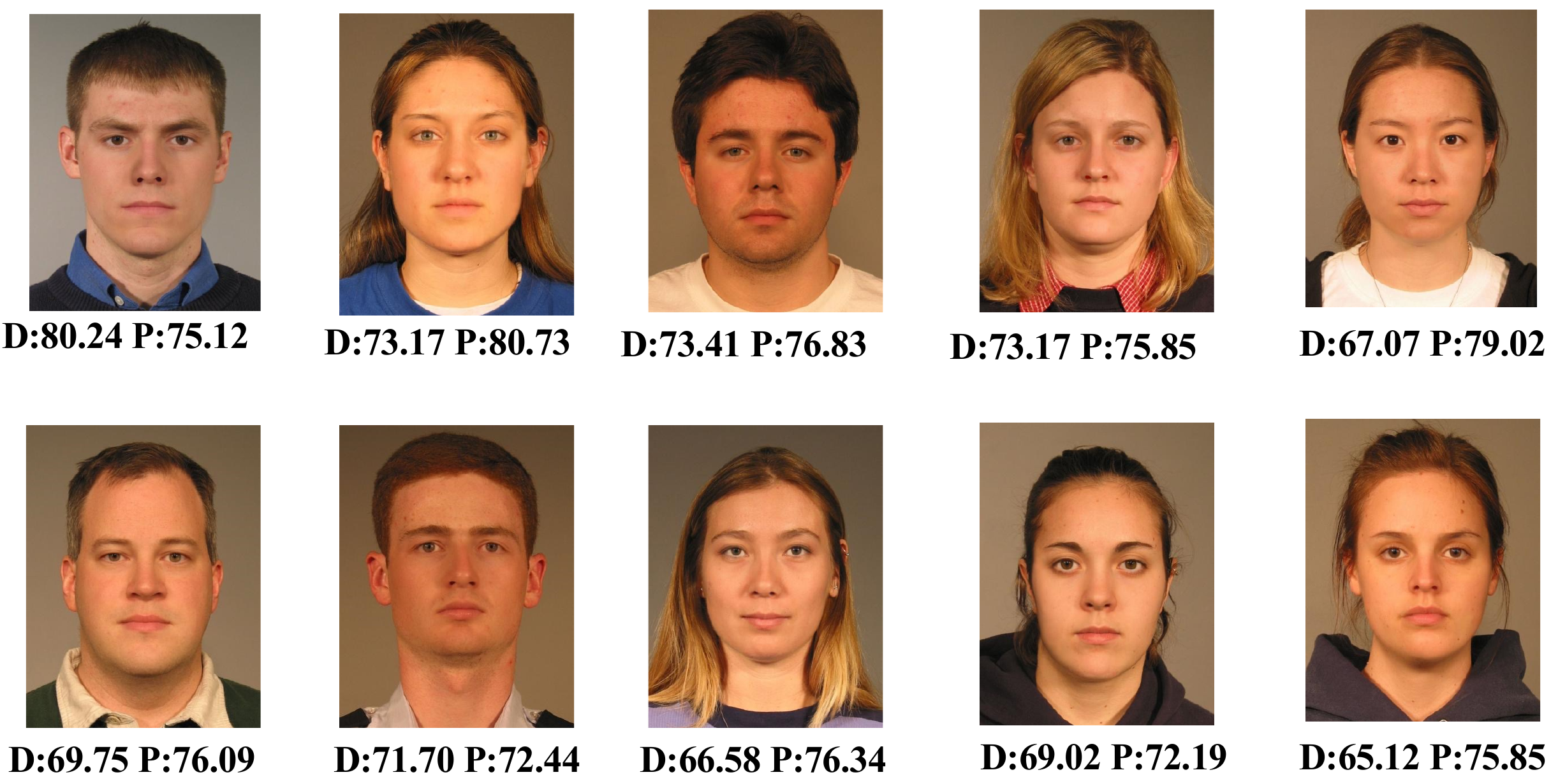}%
			\caption{Illustration of subjects whose post-processed morphed images obtain highest detection accuracy in S-MAD settings. The average detection accuracy is noted below each pair of images (D denotes digital and P denotes printed-scanned).}
			\label{fig:s-mad-lowest-accuracy}%
		\end{figure}

		\newpage
		\section{Additional Analysis - Lowest detection of bona fide in S-MAD setting}
		\label{sec:visual-analysis-s-mad}
\begin{figure}[h]
			\centering 
			\includegraphics[width=0.95\columnwidth]{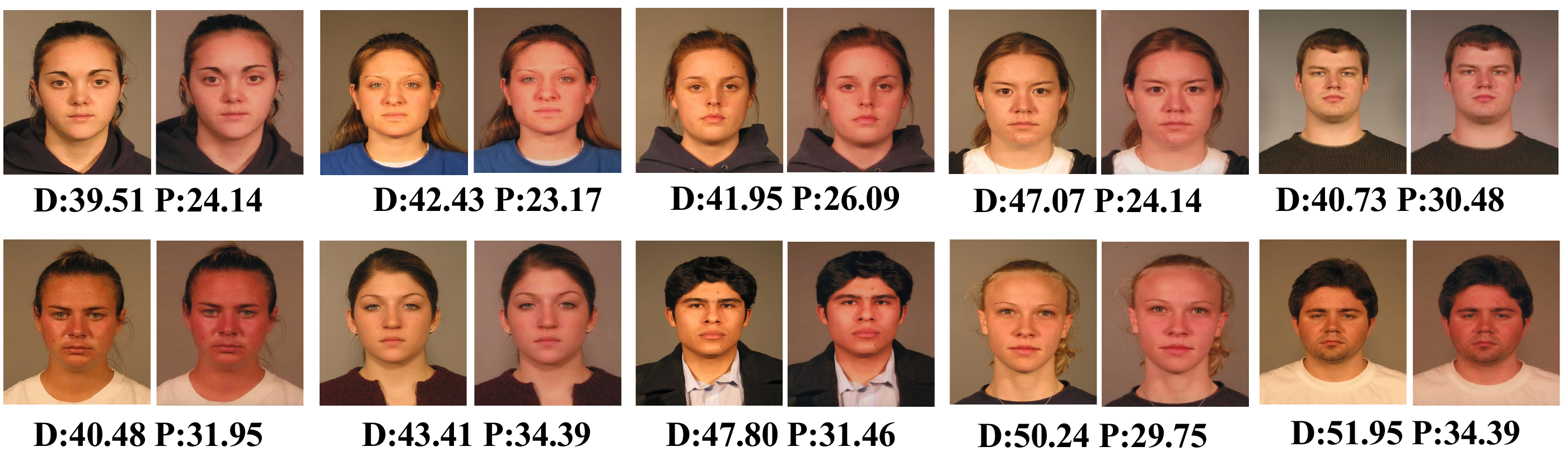}%
			\caption{Illustration of bona fide images obtaining the lowest detection accuracy in S-MAD settings. The average detection accuracy is noted below each pair of images (D denotes digital and P denotes printed-scanned). As noted from the accuracy, the average classification error increases when the images are printed and scanned in S-MAD settings.}
			\label{fig:s-mad-lowest-accuracy-bonafide-print-scan}%
		\end{figure}

		The observers tend to wrongly classify the bona fide images as morph images in S-MAD setting when the images are printed and scanned as noted in Figure~\ref{fig:s-mad-lowest-accuracy-bonafide-print-scan}. As noted from the accuracy, the average classification error increases when the images are printed and scanned in S-MAD settings, with an minimum increase of 10\%. 

		\section{Analysis of morphed images with lowest detection accuracy}
		In this section, we analyze some sample images that obtained the lowest detecting accuracy in both D-MAD and S-MAD settings to complement the quantitative analysis.

		\subsection{Additional Analysis - Analysis of morphed images with lowest detection accuracy in D-MAD}
		\label{sec:visual-analysis-d-mad}
		
\begin{figure}[h]
			\centering 
			\subfloat[Regular probe images. The top row are images in digital domain and the bottom row presents the images after printing and scanning.]{%
				\includegraphics[width=0.95\columnwidth]{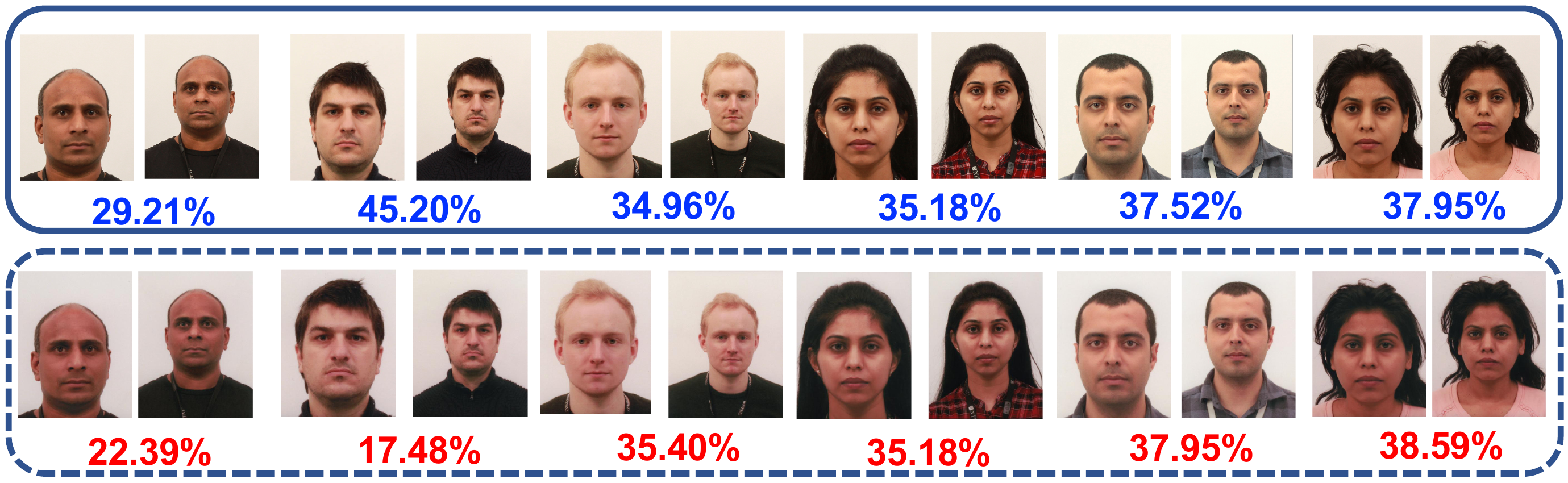}%
				\label{fig:d-mad-probe-images}%
			}\qquad
			\subfloat[Probe images from ABC Gate. The top row are images in the digital domain and the bottom row presents the images after printing and scanning.]{%
				\includegraphics[width=0.95\columnwidth]{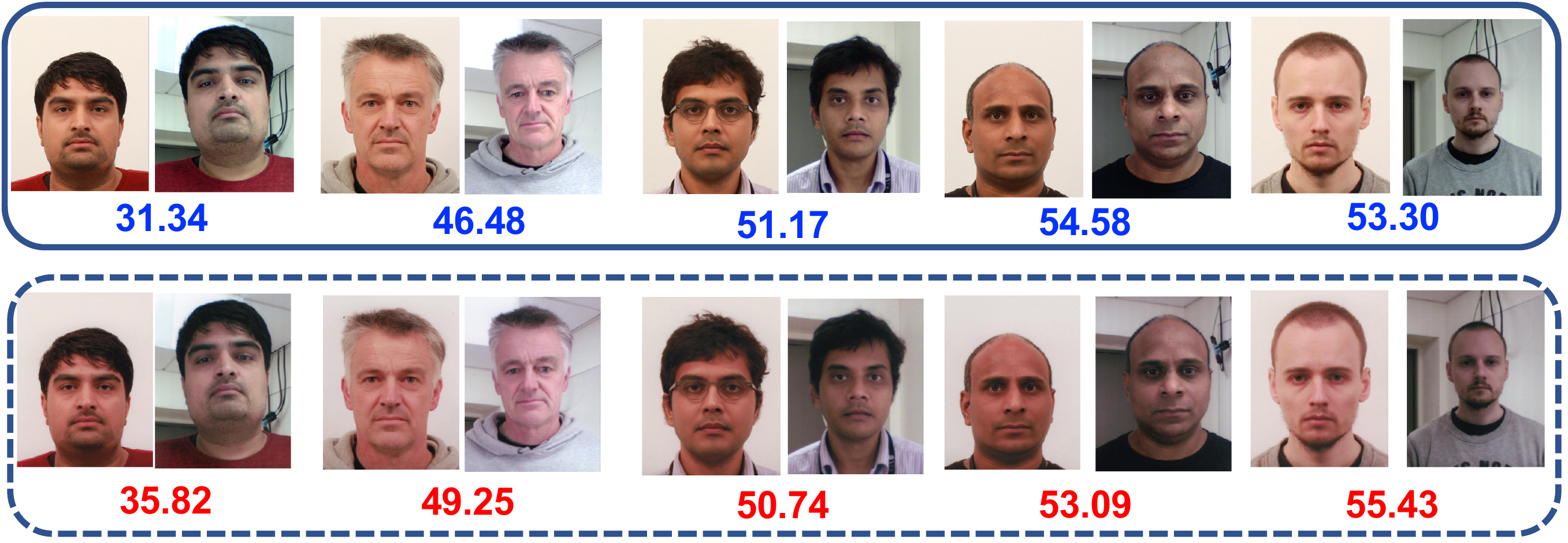}%
				\label{fig:d-mad-probe-images-gate}%
			}
			\caption{Illustration of post-processed morphed images obtaining lowest detection accuracy when compared against probe images and probe images from an ABC Gate. The average detection accuracy is noted below each pair of images.}
			\label{fig:d-mad-probe-and-gate-images}%
		\end{figure}

		Figure~\ref{fig:d-mad-probe-images} displays the detection accuracy for probe images obtained from standard cameras, while Figure ~\ref{fig:d-mad-probe-images-gate} displays the detection accuracy for probe images obtained from an ABC gate. In both cases, digital images are displayed in the top row, while printed-scanned images are displayed in the bottom row. We note a small variation in average accuracy, but the images indicate the detection accuracy does not drastically vary across observers for some sets of images. The detection accuracy for probe images obtained from conventional cameras is shown in Figure~\ref{fig:d-mad-probe-images}, whereas the detection accuracy for probe images obtained from an ABC gate is shown in Figure ~\ref{fig:d-mad-probe-images-gate}. In both instances, printed-scanned images are shown in the bottom row, while digital images are shown in the top row. While it is difficult to conclude from this observation, we would like to draw the readers’ attention to the fact that the large observer pool was made up of Caucasians, which may have been influenced by the “other-race effect” \cite{yaros2019memory}, in which observers had lower accuracy on morphed images of Asian ethnicity. However, an indicative conclusion highlights the need for additional research on examining the relationship between ethnicity and MAD, which is outside the purview of this article.

		\newpage
		\subsection{Additional Analysis - Analysis of morphed images with lowest detection accuracy in S-MAD}
		\label{sec:visual-analysis-s-mad-post-processed}
\begin{figure}[h]
			\centering 
			\includegraphics[width=0.95\columnwidth]{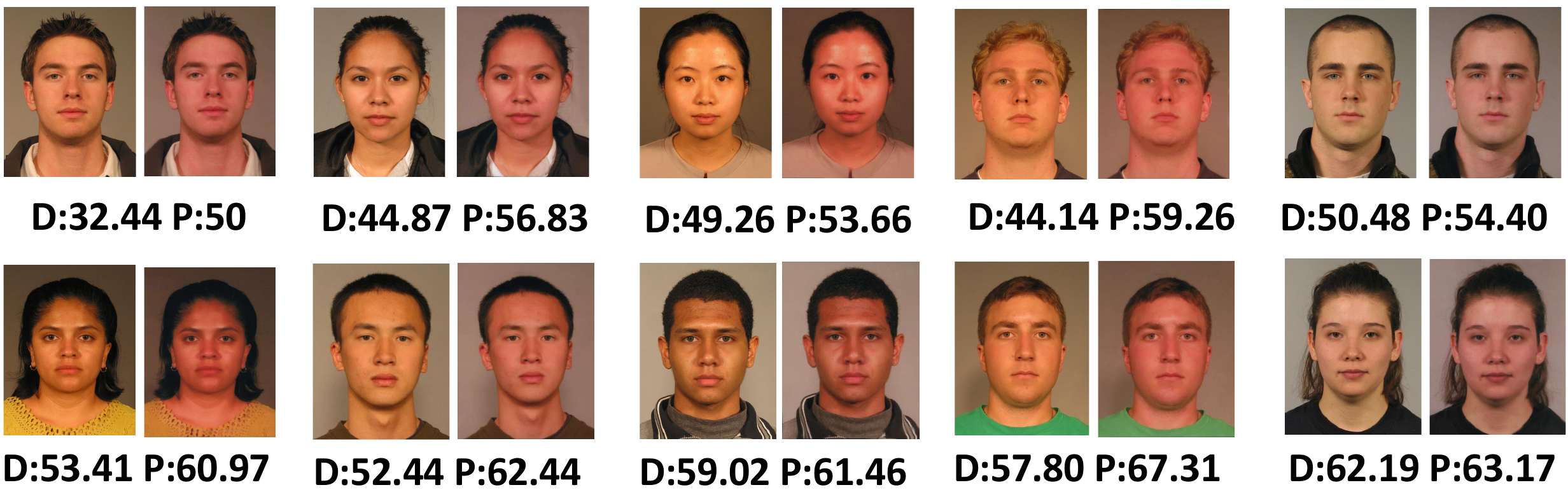}%
			\caption{Illustration of post-processed morphed images obtaining lowest detection accuracy in S-MAD settings. The average detection accuracy is noted below each pair of images (D denotes digital and P denotes printed-scanned).}
			\label{fig:s-mad-lowest-accuracy-post-processed}%
		\end{figure}

		We further analyze the images that obtained the lowest detection accuracy and present the same in Figure~\ref{fig:s-mad-lowest-accuracy-post-processed}. We observe that a sizable portion of the images with the lowest accuracy of detection are not Caucasian. The low accuracy could be potentially due to the “other-race effect” as a significant amount of observers belong to European countries. Although this is one possibility, we can also interpret the same fact as observers who are accustomed to working with a particular population daily. Further research into this aspect is required.
		As a second observation, we note that detection of morphed images in the digital domain is much lower than morphed images when printed and scanned. While algorithms are effective at detecting images in the digital domain \cite{venkatesh2020detecting,venkatesh2021morphsurvey, raja2020morphing}, humans perform significantly better in an S-MAD setting with printed-scanned morphed images. Although the detection accuracy difference between digital and printed-scanned images ranges from 10\% to 1\%, it clearly shows the importance of understanding the difference between humans and algorithms.

%
%
%
%

	\end{appendices}

\end{document}